\documentclass{article}

\usepackage{iclr2027_conference}
\usepackage[colorinlistoftodos]{todonotes}
\usepackage[utf8]{inputenc}
\usepackage{amsmath,amssymb,amsfonts}
\usepackage{float}
\usepackage{bm}
\usepackage{booktabs}
\usepackage{array}
\usepackage{longtable}
\usepackage{graphicx}
\usepackage{microtype}
\usepackage{url}
\usepackage{hyperref}
\usepackage{algorithm}
\usepackage{algpseudocode}
\usepackage{lmodern}
\usepackage{csquotes}
\makeatletter
\newlength{\algcmthang}
\newcommand{\Cmt}[2]{\settowidth{\algcmthang}{#1\ $\triangleright$\ }%
  \parbox[t]{\dimexpr\linewidth-\ALG@thistlm\relax}{\hangindent=\algcmthang\hangafter=1\relax #1\ $\triangleright$\ #2}}
\makeatother

\let\phead\paragraph
\usepackage{tikz}
\usetikzlibrary{arrows.meta,positioning,calc,fit,decorations.pathreplacing}
\definecolor{sel}{rgb}{0.05,0.42,0.24}   %

\definecolor{revblue}{RGB}{38,100,190}
\newcommand{\rev}[1]{{\color{revblue}#1}}
\newif\ifarxiv\arxivtrue
\ifarxiv
  \iclrfinalcopy                       %
  \renewcommand{\rev}[1]{#1}           %
  \definecolor{revblue}{RGB}{0,0,0}    %
\fi

\title{INSEP: Refusal-Direction Inseparability as a Defense Against Abliteration}

\author{Ivan Brugere, Junqi Jiang, Saumitra Mishra, Sumitra Ganesh \\
J.P.\ Morgan Chase}

\newcommand{\rhat}{\hat{\bm{r}}}
\newcommand{\nBackbones}{eight}
\newcommand{\nBaselines}{six}
\newcommand{\nDefenses}{seven}
\newcommand{\nAttacks}{five}

\begin{document}
\maketitle
\ifarxiv\lhead{Preprint}\fi

\begin{abstract}

Abliteration is a widely used white-box attack against the safety alignment of large language models (LLMs). It elicits harmful behavior by estimating a refusal direction in the model's activation space and projecting that direction out of the residual-stream write-out matrices. We propose INSEP, a defense that fine-tunes the model to reduce the separability of harmful and benign activations along the refusal direction. Abliteration estimates this direction as a difference in class means and projects it out. Reducing separability weakens the signal the attack relies on. An adversarial eigendecomposition selects the most separable prompt subsets. A low-rank adapter reduces their separation along that direction. Behavioral losses preserve the base model's refusal and the quality of its benign responses during training. We benchmark INSEP against \nBaselines{} published defenses under \nAttacks{} attacks, across \nBackbones{} LLMs from 1B to 14B parameters and several model families. Beyond response harm, we measure coherence, directness, over-refusal, and MMLU, to distinguish robust refusal from low harm produced by evasive or degraded responses. INSEP ranks first on the mean composite score across all \nBackbones{} base models while retaining utility, in the largest abliteration-defense evaluation reported to date.

\end{abstract}

\section{Introduction}
\label{sec:intro}

Open-weight models let users fine-tune, extend, and adapt them freely to new use cases or domains. That same access lets attackers disable safeguards by editing model weights directly. Attackers may seek to exfiltrate private information, produce harmful content, or repurpose a model for fraud. Hosted models can add input filters, rate limits, monitoring, and revocable access. An open-weight model instead gives the adversary white-box access and effectively unbounded offline compute against the safeguards encoded in its weights \citep{tamirisa2025tar,kapoor2024societal}.

Abliteration removes refusal with a single closed-form weight edit. Refusal in a safety-aligned model is mediated, to first order, by a single residual-stream direction \citep{arditi2024refusal}. The attack estimates this \textit{refusal direction} $\rhat$ from the difference between mean harmful and benign activations. It then orthogonalizes every residual-writing matrix, $W \mapsto (I-\rhat\rhat^{\top})W$. A small contrastive prompt set suffices. Open-source tools automate the permanent weight edit \citep{labonne2024abliteration,weidmann2025heretic}.

Prior defenses accept this signal and harden the behavior built on top of it. They leave the harmful and benign activation distributions separable, and train refusal to persist after the removal of one estimated direction \citep{yu2025refat,kuo2026art,xie2025deeprefusal}. Robustness obtained this way is tied to the estimate it was trained against. The attacker still recovers a usable direction, and can re-estimate it from a different contrastive prompt set, at a different layer, or from a different token position. The underlying quantity the attack depends on is left intact.

In this work, we propose \textbf{INSEP}, a defense method that removes the signal the refusal direction is estimated from. It reduces the separability of harmful and benign activations along the difference-in-means refusal direction. Once the two distributions overlap along that direction, the estimate carries little of the class difference. Removing it leaves refusal in place. This holds across the contrastive samples we test, including an attacker who estimates the direction from a corpus disjoint from every training pool. Robustness therefore does not depend on which contrastive prompts the attacker selects, provided the training prompts are diverse enough to cover them. Inseparability is easy to obtain by destroying the concept. Linear concept erasure \citep{ravfogel2020inlp,belrose2023leace} removes linear separability. Applied to refusal, it would remove the refusal behavior along with it. INSEP is trained under behavioral constraints that preserve refusal on harmful prompts and normal responses on benign ones. Refusal must therefore be carried by structure that projecting out the estimated direction cannot remove.

Our evaluation methodology distinguishes defended refusal from degraded generation, and measures the defense $\times$ attack grid rather than a slice of it. Previous work evaluates with a single harm score, which treats evasive, incoherent, and off-topic responses as safe (Table~\ref{tab:evaldef}, also see Appendix~\ref{app:evalcat} for a detailed survey of relevant defense research). Instead, we explicitly measure these utility aspects alongside harm: coherence, directness, over-refusal, and capability. Published comparisons also cover too few attacks, defenses and base models to establish a common threat model. Every prior abliteration defense is evaluated against at most one abliteration-family attack, at most four competing defenses, and at most five base models. A coherence measure is reported once. Directness is never reported. Reported seeds, where present, vary either the attack estimate or defense training, but never both. By contrast, we evaluate \nDefenses{} defenses against \nAttacks{} attacks, four of them abliteration-family, across \nBackbones{} base models. Attacks select their worst-performing layer. Defense-training and attack seeds are crossed $3\times3$ to separate the two sources of variation.

\phead{Contributions.}
\begin{itemize}
\item \textbf{Defense.} We propose INSEP, a new abliteration defense method that adversarially reduces harmful--benign activation separability along the difference-in-means direction. This degrades the estimator signal that abliteration attacks depend on. Prior defenses instead harden refusal against one
estimated direction. INSEP preserves refusal and benign behavior.
\item \textbf{Evaluation.} A benchmark of \nDefenses{} defenses over \nAttacks{} attacks and \nBackbones{} base models, the widest defense $\times$ attack grid reported for abliteration to the best of our knowledge
(Table~\ref{tab:evaldef}), measuring robustness, behavioral quality, capability, and seed stability simultaneously.
\end{itemize}

\section{Related Work}
\label{sec:related}

\rev{Abliteration defenses harden refusal \emph{behavior} against removal of an estimated direction, but leave the harmful and benign representations separable. An attacker who re-estimates the direction at evaluation recovers a usable one.} \citet{arditi2024refusal} estimate this direction from harmful and benign last-token activations, then remove it from every residual-writing matrix. Labonne's procedure popularized this weight edit \citep{labonne2024abliteration}. ReFAT \citep{yu2025refat}, ART \citep{kuo2026art}, and DeepRefusal \citep{xie2025deeprefusal} train refusal to persist after removal of a mean-derived direction. ART is proposed as an add-on to an existing fine-tuning defense. We evaluate it standalone. Extended-Refusal \citep{abushairah2025abliteration} disperses refusal across token positions but remains evaluated against the same attack statistic.

Recent attacks go beyond a difference of means and match the full harmful and benign distributions. The difference in means discards the covariance of the harmful and benign activations. OT-ablation aligns both the mean and covariance through an optimal-transport map \citep{nanfack2026ot}. It removes refusal where mean-only ablation leaves it recoverable.

SOM and RFM recover refusal from multiple adversarially selected directions. SOM fits a self-organizing map over harmful activations and ablates its bank of local directions \citep{piras2026som}. This attack breaks the Representation Rerouting defense that a single canonical direction leaves intact \citep{zou2024circuitbreakers}. RFM instead recovers a multidimensional refusal subspace with a recursive feature machine \citep{winninger2026rfm}. Both methods choose their own removal directions. Neither uses the direction defined by class means.

AMRA and DDO corrupt the attacker's estimate through engineered alternative signals. AMRA remaps refusal activations to random aliases and patches the downstream readers \citep{truong2026amra}. DDO injects a high-magnitude nonlinear decoy signal into MLP neurons. The attacker then ablates the decoy and leaves refusal in place \citep{muhamed2026ddo}. INSEP instead minimizes harmful--benign separability along the estimated refusal direction under adversarial prompt selection. Its objective targets the distributional signal from which removable directions are estimated.

General safety defenses test whether broader hardening transfers to abliteration. Representation Rerouting (RR) reroutes harmful representations \citep{zou2024circuitbreakers}. Vaccine \citep{huang2024vaccine}, RepNoise \citep{rosati2024representation}, Booster \citep{huang2025booster}, TAR \citep{tamirisa2025tar}, and LAT \citep{sheshadri2025lat} defend against harmful fine-tuning or latent perturbations. We evaluate selected methods alongside abliteration-specific defenses.

INSEP is closest to linear concept erasure \citep{ravfogel2020inlp,belrose2023leace}. Erasure removes linear separability and deletes the concept with it. INLP \citep{ravfogel2020inlp} repeatedly trains a linear classifier for an attribute and projects activations onto its null-space. LEACE \citep{belrose2023leace} provably prevents every linear classifier from detecting a concept while minimally changing the representation. Applied to refusal, either method would remove the refusal behavior itself. INSEP instead retains the base model's behavior while entangling the activations. Refusal must therefore be carried by structure that projecting out the estimated direction cannot remove. \citet{bailey2026obfuscated} show that activations can be reshaped to evade latent-space classifiers while behavior is preserved, which is the property INSEP trains for and the reason such a defense must be measured.

\section{Method}
\label{sec:method}

\begin{figure}[t]
\centering
\resizebox{\textwidth}{!}{%
\begin{tikzpicture}[
  font=\small,
  bx/.style   ={draw, rounded corners=2pt, align=center, inner sep=3pt, font=\scriptsize},
  panel/.style={draw=black!45, rounded corners=3pt, thick},
  ar/.style   ={-{Stealth[length=2mm]}, semithick},
  tny/.style  ={font=\tiny, align=center},
]
\node[font=\footnotesize] at (6.32,5.92) {\textbf{INSEP}: adversarial minibatching on the whitened discriminant, for harmful/benign inseparability};
\draw[black!35] (0.00,5.64) -- (12.64,5.64);
\def\ptop{5.44}\def\pmid{1.22}\def\ctop{1.02}\def\cbot{-2.62}
\draw[panel] (0.00,\pmid) rectangle (2.62,\ptop);
\draw[panel] (2.86,\pmid) rectangle (12.64,\ptop);
\draw[panel] (0.00,\cbot) rectangle (12.64,\ctop);

\node[tny] at (1.31,5.16) {\textbf{(a) draw a batch}};
\node[bx, fill=orange!12, minimum width=16mm] (Ub) at (1.48,4.26) {harmful pool $U$};
\node[bx, fill=blue!12,   minimum width=16mm] (Sb) at (1.48,3.76) {benign pool $S$};
\coordinate (jn) at (1.48,3.34);
\draw[semithick] (Ub.west) -- (0.28,4.26) -- (0.28,3.34) -- (jn);
\draw[semithick] (Sb.south) -- (jn);
\draw[ar] (jn) -- (1.48,3.08);
\node[bx, minimum width=19mm, minimum height=10mm] at (1.48,2.64) {batch\\[1pt]$B$ per class};
\node[tny] at (1.36,1.86) {redrawn each step};

\node[tny,anchor=west] at (3.02,5.16) {\textbf{(b) select the minibatch of $n$ per class}};
\node[tny,anchor=east] at (12.56,5.16) {\textcolor{orange!85!black}{$\bullet$}\,harmful selected};
\node[tny,anchor=east] at (12.56,4.92) {\textcolor{blue!70}{$\bullet$}\,benign selected};
\draw[black!25] (5.28,1.66) -- (5.28,4.94);
\node[tny] at (4.01,4.66) {whitened spectrum\\eigendecomposition\\of batch $B$};
\foreach \i/\h/\c in {0/0.40/sel, 1/0.14/sel!30, 2/0.008/black!30, 3/0.008/black!30}
   \fill[\c] ({3.61+\i*0.26},3.89) rectangle ({3.77+\i*0.26},{3.89+\h});
\draw[semithick,black!70] (3.29,3.89) -- (4.73,3.89);
\node[font=\tiny,text=sel] at (3.69,3.73) {$v_1$};
\node[font=\tiny,text=sel!45] at (3.95,3.73) {$v_2$};
\draw[ar,black!55] (4.01,3.55) -- (4.01,3.31);
\node[tny] at (4.01,3.11) {target $d_\ell$ w.r.t.\\eigenvector $v_k$};

\node[tny,anchor=west,text=sel] at (5.40,4.52) {\textbf{(b-1)} \textsc{eigen@k}: benign uniform, harmful maximizes $d_\ell$ w.r.t.\ $v_1$};
\draw[semithick,-{Stealth[length=1.5mm]}] (5.40,4.24) -- (11.90,4.24);
\node[font=\tiny,anchor=west] at (11.95,4.24) {$v_1$};
\foreach \x in {5.90,6.35,6.80,7.15,7.55,7.95,8.40,8.85} \draw[blue!55] (\x,4.24) circle (1.5pt);
\foreach \x in {8.20,8.75,9.30,10.05,10.18,10.31,11.10,11.55} \draw[orange!60] (\x,4.24) circle (1.5pt);
\foreach \x in {6.35,7.55,8.40} \fill[blue!70] (\x,4.24) circle (2.0pt);
\foreach \x in {10.31,11.10,11.55} \fill[orange!85!black] (\x,4.24) circle (2.0pt);

\node[tny,anchor=west,text=sel!70!black] at (5.40,3.90) {\textbf{(b-2)} \textsc{eigencoreset}: jointly maximizing $d_\ell$ windows};
\draw[semithick,-{Stealth[length=1.5mm]}] (5.40,3.62) -- (11.90,3.62);
\node[font=\tiny,anchor=west] at (11.95,3.62) {$v_1$};
\fill[sel,opacity=0.10] (5.78,3.47) rectangle (6.92,3.77);
\fill[sel,opacity=0.10] (9.93,3.47) rectangle (10.43,3.77);
\foreach \x in {5.90,6.35,6.80,7.15,7.55,7.95,8.40,8.85} \draw[blue!55] (\x,3.62) circle (1.5pt);
\foreach \x in {8.20,8.75,9.30,10.05,10.18,10.31,11.10,11.55} \draw[orange!60] (\x,3.62) circle (1.5pt);
\foreach \x in {5.90,6.35,6.80} \fill[blue!70] (\x,3.62) circle (2.0pt);
\foreach \x in {10.05,10.18,10.31} \fill[orange!85!black] (\x,3.62) circle (2.0pt);

\node[tny,anchor=west,text=black!55] at (5.40,3.28) {\textbf{(b-3)} \textsc{random}: $n$ drawn uniformly in both};
\draw[semithick,-{Stealth[length=1.5mm]}] (5.40,3.00) -- (11.90,3.00);
\node[font=\tiny,anchor=west] at (11.95,3.00) {$v_1$};
\foreach \x in {5.90,6.35,6.80,7.15,7.55,7.95,8.40,8.85} \draw[blue!55] (\x,3.00) circle (1.5pt);
\foreach \x in {8.20,8.75,9.30,10.05,10.18,10.31,11.10,11.55} \draw[orange!60] (\x,3.00) circle (1.5pt);
\foreach \x in {5.90,8.40,8.85} \fill[blue!70] (\x,3.00) circle (2.0pt);
\foreach \x in {8.75,10.31,11.10} \fill[orange!85!black] (\x,3.00) circle (2.0pt);

\node[tny,anchor=east,align=right] at (6.75,2.05) {separability\\$d_\ell$ of the\\picked $n$};
\fill[sel!70!black] (6.90,2.335) rectangle (10.40,2.465);
\node[font=\tiny,anchor=west,text=sel!70!black] at (10.50,2.40) {\textsc{eigencoreset}};
\fill[sel] (6.90,1.985) rectangle (9.82,2.115);
\node[font=\tiny,anchor=west,text=sel] at (9.92,2.05) {\textsc{eigen@k}};
\fill[black!45] (6.90,1.635) rectangle (8.35,1.765);
\node[font=\tiny,anchor=west,text=black!45] at (8.45,1.70) {\textsc{random}};
\node[tny,anchor=west] at (3.02,1.44) {selection on $v_k$; $n$ minibatch defines $\rhat\propto\mu_U-\mu_S$ per training step};

\node[tny,anchor=west] at (0.14,0.80) {\textbf{(c) the update step: $W\leftarrow W+\Delta W$, rank-one $\Delta W=\rhat\,a^{\!\top}$}};
\draw[black!30] (7.30,-2.50) -- (7.30,0.56);
\node[tny] at (3.55,0.40) {activations at layer $\ell$, projected on $\rhat$};
\draw[semithick,-{Stealth[length=1.8mm]}] (0.55,-1.50) -- (6.75,-1.50);
\node[font=\scriptsize,anchor=west] at (6.80,-1.50) {$\rhat$};
\foreach \x/\y in {1.88/-0.43, 2.30/-0.75, 1.85/-0.95, 2.18/-0.72, 1.88/-1.01, 1.86/-0.87, 2.60/-0.59, 2.46/-1.05, 2.12/-0.62, 2.09/-1.07, 2.10/-0.71}{
  \draw[blue!60,opacity=0.30,line width=0.28pt] (\x,\y) -- (\x,-1.45);
  \fill[blue!60] (\x,\y) circle (1.0pt);}
\foreach \x/\y in {4.76/-0.91, 4.63/-1.06, 4.19/-0.49, 4.63/-0.72, 4.30/-0.62, 4.24/-0.76, 4.34/-0.38, 4.84/-0.86, 4.10/-0.92, 4.52/-0.54, 4.68/-0.80}{
  \draw[orange!85!black,opacity=0.30,line width=0.28pt] (\x,\y) -- (\x,-1.45);
  \fill[orange!85!black] (\x,\y) circle (1.0pt);}
\draw[blue!70,line width=1.1pt] (2.12,-1.61) -- (2.12,-1.37);
\draw[orange!85!black,line width=1.1pt] (4.48,-1.61) -- (4.48,-1.37);
\node[font=\tiny,text=blue!70,anchor=north] at (2.12,-1.63) {$\mu_S$};
\node[font=\tiny,text=orange!85!black,anchor=north] at (4.48,-1.63) {$\mu_U$};
\draw[decorate,decoration={brace,amplitude=3pt},black!60] (2.12,-0.20) -- (4.48,-0.20);
\node[tny] at (3.30,0.02) {gap};
\draw[<->,blue!55,line width=0.6pt] (1.87,-1.30) -- (2.37,-1.30);
\draw[<->,orange!70!black,line width=0.6pt] (4.23,-1.30) -- (4.73,-1.30);
\node[font=\tiny,text=blue!60,anchor=east] at (1.81,-1.30) {$\sigma_S$};
\node[font=\tiny,text=orange!70!black,anchor=west] at (4.79,-1.30) {$\sigma_U$};
\node[tny,anchor=north] at (3.55,-2.10) {$d_\ell=(\mu_U-\mu_S)^2/(\sigma_S^2+\sigma_U^2)$; objective: minimize $d_\ell$};
\node[tny] at (10.0,0.40) {the edit is one vector, applied to every $h_i$};
\node[font=\scriptsize,anchor=east] at (8.30,-0.30) {$\Delta W=$};
\fill[sel!75] (8.42,-0.72) rectangle (8.60,0.10);
\node[font=\tiny,anchor=north] at (8.51,-0.76) {$\rhat$};
\node[font=\tiny] at (8.76,-0.30) {$\otimes$};
\fill[black!45] (8.92,-0.40) rectangle (9.76,-0.22);
\node[font=\tiny,anchor=south] at (9.34,-0.18) {$a^{\!\top}$};
\node[tny,anchor=west,align=left] at (9.94,-0.30) {one direction,\\one weight per input};
\draw[semithick,-{Stealth[length=1.6mm]}] (8.40,-1.60) -- (9.90,-1.60);
\node[font=\tiny,anchor=west] at (9.95,-1.60) {$\rhat$};
\fill[black!55] (8.76,-1.60) circle (1.3pt);
\node[font=\tiny,anchor=south] at (8.76,-1.54) {$h_i$};
\draw[sel,semithick,-{Stealth[length=1.4mm]}] (8.76,-1.60) -- (9.46,-1.60);
\node[font=\tiny,text=sel,anchor=north] at (9.11,-1.66) {$s_i\,\rhat$};
\node[font=\scriptsize,anchor=west] at (10.30,-1.44) {$\delta h_i=s_i\,\rhat$};
\node[font=\scriptsize,anchor=west] at (10.30,-1.82) {$s_i=a^{\!\top}h_i$};
\node[tny,text=sel,anchor=north] at (9.90,-2.16) {each $h_i$ moves along $\rhat$, by its own $s_i$};
\node[tny,rotate=90] at (-0.64,0.60) {repeat for $t=1,\dots,T$};
\end{tikzpicture}%
}
\caption{One INSEP training step. \textbf{(a)} A batch of $B$ prompts per class is drawn from the harmful pool $\mathcal{U}$ and the benign pool $\mathcal{S}$. \textbf{(b)} The adversary whitens the batch activations, eigendecomposes the harmful activations about the benign mean, and selects the $n$ prompts per class that are most separable along the leading axis. \textbf{(c)} The update step, which minimizes separability on the selected minibatch. The panel draws one write direction. INSEP trains a rank-32 adapter on \texttt{o\_proj} and \texttt{down\_proj}.}
\label{fig:step}
\end{figure}
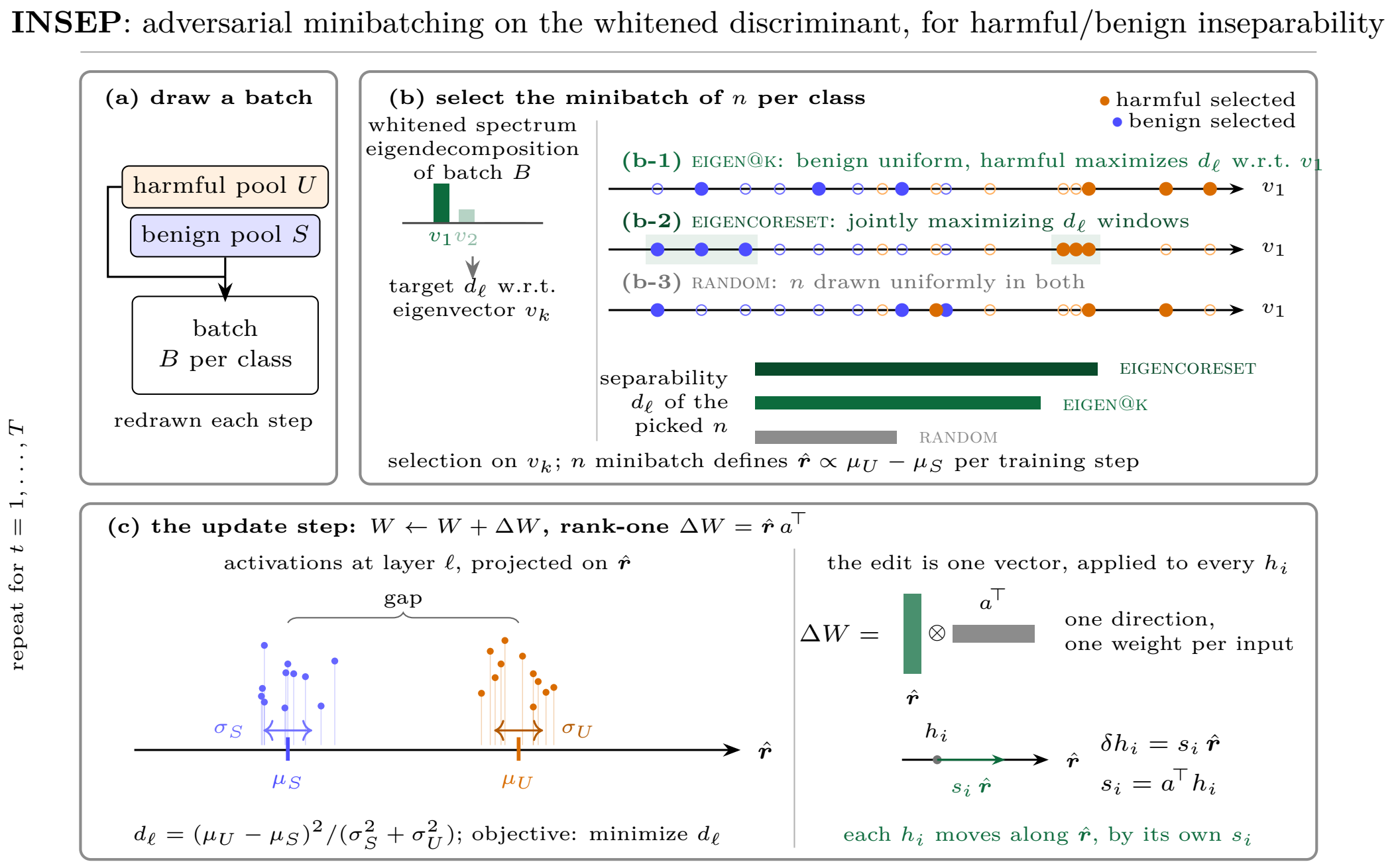

Figure~\ref{fig:step} illustrates one INSEP training step (see Sections~\ref{sec:objective} and \ref{sec:adversary} for notation detail). Each step couples (i) an adversary that selects the most separable harmful/benign prompts and (ii) a defender that reduces that separability by updating a low-rank adapter. First, we sample a candidate batch of $B$ harmful and $B$ benign prompts from independent pools (Figure~\ref{fig:step}a). Next, the adversary whitens the batch activations, performs an eigendecomposition, and selects a minibatch of $n$ prompts per class that is maximally separable along the chosen eigenvector (Figure~\ref{fig:step}b). Finally, the defender updates the adapter to reduce the minibatch separability while preserving the base model's behavior (Figure~\ref{fig:step}c).

We discussed the motivations behind our method in Section~\ref{sec:intro}. Next, Sections~\ref{sec:objective}, \ref{sec:adversary}, and~\ref{sec:data} define the training objective, adversarial selection procedure, and training data.

\subsection{Training objective}
\label{sec:objective}

INSEP minimizes the separability of harmful and benign activations along the refusal direction. Let $h_\ell(x)\in\mathbb{R}^{m}$ be the residual-stream activation entering layer $\ell$ at the final prompt token. For a harmful minibatch $U$ and a benign minibatch $S$ with means $\mu_U,\mu_S$, let $\rhat=(\mu_U-\mu_S)/\lVert\mu_U-\mu_S\rVert$ be the difference-of-means direction abliteration removes, and let $p(x)=h_\ell(x)^{\top}\rhat$. The per-layer separability is
\begin{equation}
  d_{\ell}(\theta)
  \;=\;
  \frac{\big(\mathbb{E}_{U}[p]-\mathbb{E}_{S}[p]\big)^{2}}
       {\mathrm{Var}_{U}[p]+\mathrm{Var}_{S}[p]},
  \label{eq:lsep}
\end{equation}
the squared discriminability index along $\rhat$. The ratio is invariant to the scale of the activations, so the model cannot lower it by shrinking the representation. $\rhat$ is recomputed from each minibatch and detached from the gradient, matching an attacker who re-estimates the direction after every update. The layer set $\mathcal{I}$ holds every layer of the base model at whose residual-stream input separability is measured. The loss reduces the per-layer values over $\mathcal{I}$ to the mean of the $w$ largest,
\begin{equation}
  d(\theta) = \frac{1}{w}\sum_{\ell\in\mathrm{top}_{w}(\mathcal{I})} d_{\ell}(\theta),
  \label{eq:sep}
\end{equation}
where $\mathrm{top}_{w}(\mathcal{I})$ holds the $w$ layers with the largest $d_\ell$ and $w\in\{1,\dots,|\mathcal{I}|\}$. At $w=|\mathcal{I}|$ the loss is the mean over all layers. Smaller $w$ concentrates the loss on the layers an attacker would target. Larger $w$ smooths the refusal direction over layers. The layer rule has no stopping condition. We therefore also search a null-referenced variant that halts when no layer's separability exceeds a label-permuted baseline (Appendix~\ref{app:active-set}).

Three behavioral terms keep the trained model's behavior equal to the base model's: refusal $\mathcal{R}$, compliance $\mathcal{C}$, and prefill refusal $\mathcal{P}$, the harmful-prefix augmentation of \citet{qi2024shallow}. INSEP trains a low-rank adapter $\theta$ on the attention and MLP output projections below the deepest layer in $\mathcal{I}$, with the base weights frozen, and minimizes
\begin{equation}
  \mathcal{L}(\theta)
  \;=\;
  \lambda_{\mathrm{d}}\,\frac{d(\theta)}{d(\theta_{0})}
  \;+\;
  \lambda_{\mathrm{r}}\,\mathcal{R}(\theta)
  \;+\;
  \lambda_{\mathrm{c}}\,\mathcal{C}(\theta)
  \;+\;
  \lambda_{\mathrm{p}}\,\mathcal{P}(\theta).
  \label{eq:objective}
\end{equation}
Normalizing the separability term by its value at initialization $\theta_{0}$ puts it on the scale of the likelihood terms. $\mathcal{R}$ is the teacher-forced negative log-likelihood of the base model's own refusals on the harmful prompts it refuses, and $\mathcal{C}$ is the same loss on its answers to the benign prompts it answers. The base model's errors are dropped, so no term trains toward a wrong response or adds a refusal the base model lacks. $\mathcal{P}$ is the refusal target under an affirmative prefill. The opener is resampled for every prompt at every step, so refusal is trained from a distribution of compliant openings. Appendix~\ref{app:prefill} describes the prefill pool.

\subsection{Adversarial selection}
\label{sec:adversary}

The adversary selects $n$ harmful and $n$ benign prompts that are maximally separable along a chosen axis. Fixing the axis is necessary because, when activation dimension exceeds batch size, some direction can perfectly separate any finite batch. We choose the axis by whitening the batch activations and eigendecomposing the harmful activations about the benign mean. The leading eigenvector identifies the direction along which the harmful prompts are farthest from the benign prompts relative to their spread. After projecting both classes onto this axis, \textsc{eigencoreset} finds the maximally separable minibatch exactly (Appendix~\ref{app:coreset}). \textsc{eigen@1} instead uses a greedy approximation and samples the benign minibatch uniformly. Although \textsc{eigen@1} is suboptimal for any fixed batch, the hyperparameter search only slightly favors \textsc{eigencoreset} (Appendix~\ref{app:optuna}). A plausible reason is that the random benign sample of \textsc{eigen@1} encourages exploration. Our defaults keep \textsc{eigencoreset}, the exact adversary. Both methods redraw the candidate batch and repeat selection at every step. The defense therefore tracks the regions that remain most separable as the weights change.

\begin{algorithm}[t]
\caption{INSEP training. $h^{\theta}_{\ell}(x)$: layer-$\ell$ residual activation of the adapted model at $x$'s final token; $p_{\theta}(\cdot\mid x)$: its next-token distribution. \textsc{Sep}$(h;X,Y)=\{d_\ell\}_{\ell\in\mathcal{I}}$: per-layer separability (Equation~\ref{eq:lsep}) of $X$ against $Y$ at the residual-stream input of each layer in $\mathcal{I}$. $\textsc{Reduce}_{w}$: mean of the $w$ largest scores, the mean over all layers when $w{=}|\mathcal{I}|$ (Equation~\ref{eq:sep}). \textsc{Nll}$(p,\{(x,y)\})=-\tfrac{1}{|\cdot|}\sum\log p(y\mid x)$: teacher-forced negative log-likelihood. \textsc{Eig}$(H_U,H_S)$: whitened eigendecomposition of the harmful activations about the benign mean, returning axes $V$ and eigenvalues $\Lambda$ (Appendix~\ref{app:coreset}). $\textsc{Adversary}_{A}((\Lambda,V);U^{(B)},S^{(B)})$: the size-$n$ minibatch $U^{(n)},S^{(n)}$ selected from the size-$B$ batch by mode $A$ (Section~\ref{sec:adversary}); \textsc{eigencoreset} selects both classes, and \textsc{eigen@1} draws the benign minibatch uniformly. $\pi\sim\Pi$: a prefill string at the response position (Section~\ref{sec:objective}).}
\label{alg:insep}
\begin{algorithmic}[1]
\Statex \textbf{Input:} base model $M_0$; harmful pool $\mathcal{U}$; benign pool $\mathcal{S}$; prefill pool $\Pi$
\Statex \textbf{Hyperparameters:} rounds $T$; learning rate $\eta$; batch size $B$; minibatch size $n$; adversary $A\in\{\textsc{eigen@1},\textsc{eigencoreset}\}$; site reduction $w\in\{1,\dots,|\mathcal{I}|\}$; weights $\lambda_{\mathrm{d}},\lambda_{\mathrm{r}},\lambda_{\mathrm{c}},\lambda_{\mathrm{p}}$
\Statex \textbf{Output:} adapter $\theta$ on the writers below $\max\mathcal{I}$
\State \Cmt{$\theta \gets \bm{0}$}{frozen base}
\State \Cmt{$R \gets \{(x, M_0(x)) : x \in \mathcal{U},\ M_0(x)\text{ refuses}\}$}{base's refusals, stored once}
\State \Cmt{$C \gets \{(x, M_0(x)) : x \in \mathcal{S},\ M_0(x)\text{ complies}\}$}{base's compliances, stored once}
\State \Cmt{$d_0 \gets \textsc{Reduce}_{w}\big(\textsc{Sep}(h^{0}; \mathcal{U}, \mathcal{S})\big)$}{normalizer at init, on base activations}
\For{$t = 1,\dots,T$}
    \State \Cmt{$U^{(B)} \overset{\mathrm{unif}}{\sim} \mathcal{U},\ S^{(B)} \overset{\mathrm{unif}}{\sim} \mathcal{S}$}{candidate batch, $|U^{(B)}|{=}|S^{(B)}|{=}B$, scored without gradient}
    \State \Cmt{$(\Lambda, V) \gets \textsc{Eig}\big(h^{\theta}(U^{(B)}),\ h^{\theta}(S^{(B)})\big)$}{whitened eigenspace of the batch}
    \State \Cmt{$U^{(n)}, S^{(n)} \gets \textsc{Adversary}_{A}\big((\Lambda, V);\ U^{(B)}, S^{(B)}\big)$}{Attack: minibatch along the selected axis (Section~\ref{sec:adversary})}
    \State \Cmt{$d \gets \textsc{Reduce}_{w}\big(\textsc{Sep}(h^{\theta}; U^{(n)}, S^{(n)})\big)$}{Defend: mean separability of the $w$ most separable layers (Equation~\ref{eq:sep})}
    \State \Cmt{$R^{(n)} \overset{\mathrm{unif}}{\sim} R,\ C^{(n)} \overset{\mathrm{unif}}{\sim} C$}{minibatches of base behavior}
    \State \Cmt{$\mathcal{R} \gets \textsc{Nll}(p_\theta, R^{(n)})$}{output space: keep base's refusals}
    \State \Cmt{$\mathcal{C} \gets \textsc{Nll}(p_\theta, C^{(n)})$}{output space: keep base's compliances}
    \State \Cmt{$\mathcal{P} \gets \textsc{Nll}(p_\theta, \{(\pi\,\|\,x,\, y) : (x,y)\in R^{(n)},\ \pi\sim\Pi\})$}{prefill-adversarial refusal}
    \State \Cmt{$\mathcal{L} \gets \lambda_{\mathrm{d}}\, d/d_0 + \lambda_{\mathrm{r}}\mathcal{R} + \lambda_{\mathrm{c}}\mathcal{C} + \lambda_{\mathrm{p}}\mathcal{P}$}{Equation~\ref{eq:objective}}
    \State $\theta \gets \theta - \eta\,\nabla_\theta \mathcal{L}$
\EndFor
\State \Return $\theta$
\end{algorithmic}
\end{algorithm}

\subsection{Training data}
\label{sec:data}

Training data is drawn from independent pools that do not share prompts with the evaluation. Harmful and benign prompts are sampled independently and are not matched pairs. Table~\ref{tab:data} lists the pools. The harmful pool is drawn from AdvBench \citep{zou2023advbench}, BeaverTails \citep{ji2023beavertails} and HarmBench \citep{mazeika2024harmbench}, stratified by their harm categories. The benign pools are Alpaca \citep{taori2023alpaca} and the benign split of WildJailbreak \citep{jiang2024wildjailbreak}. We augment the harmful pool with paraphrasing and translation round-trips, keeping only candidates that a judge rates equivalent to the original and that a safety classifier still marks unsafe (Appendix~\ref{app:augment}). The hyperparameter search favors this extended pool across the base models (Table~\ref{tab:archpanel}). Harm is scored on CatQA \citep{bhardwaj2024catqa}. CatQA does not share prompts with any training pool. An automatic check enforces this disjointness.

\begin{table}[t]
\caption{Training pools. A configuration pairs one harmful pool with one benign pool and is named \texttt{<harmful>\_x\_<benign>}. Counts are prompts after filtering.}
\label{tab:data}
\centering
\small
\begin{tabular}{lllr}
\toprule
Class & Pool & Construction & $n$ \\
\midrule
Harmful & \texttt{pooled}    & AdvBench, BeaverTails and HarmBench, stratified & 1{,}627 \\
Harmful & \texttt{pooledExt} & \texttt{pooled} plus paraphrase and translation & 5{,}056 \\
Harmful & \texttt{wjbH}      & WildJailbreak \texttt{vanilla\_harmful} & 50{,}050 \\
\midrule
Benign & \texttt{alpaca} & Alpaca instructions, topically distant & 52{,}002 \\
Benign & \texttt{wjbB}   & WildJailbreak \texttt{vanilla\_benign}, near-miss form & 50{,}050 \\
Benign & \texttt{twin}   & content-matched to each harmful prompt & per item \\
\bottomrule
\end{tabular}
\end{table}

An ablation over harmful and benign pools, including the larger WildJailbreak corpus, selects the pooled harmful sets with Alpaca (Appendix~\ref{app:augment}).

\section{Evaluation Methodology}
\label{sec:eval}

We score every defense on harm under \nAttacks{} attacks, together with behavioral and capability measures on its responses. A degenerate model scores low harm for the wrong reason. Replies to harmful prompts that are evasive, off-topic, gibberish, or blanket refusals are non-compliant. The harm classifier credits them as safe. Response quality therefore confounds harm, and measuring it separates a defense that refuses correctly from a model that has stopped answering. 

\phead{Threat model.}\label{sec:threat} The attacker holds the open weights, a set of harmful and harmless prompts, and unbounded offline compute. It estimates the refusal direction from its own prompts and chooses the layer of removal, the strength $\alpha$ and the rank $k$. Each defense is scored against the attacker's strongest coherent configuration. The defender acts once, training a parameter-efficient adapter on the base model before release, and never observes the attacker's prompts. Harmful fine-tuning \citep{qi2023finetuning} retrains the weights and is a separate threat with its own defenses and evaluations \citep{huang2024survey}. We do not evaluate it (Appendix~\ref{app:evalcat}). \citet{che2025tampering} argue that tampering attacks give conservative estimates of worst-case capability. Our attack suite is therefore the weight-edit family and prefill, described below.

\phead{Base models.} We define \emph{base model} as the released instruction-tuned checkpoint a defense is applied to, before any defense or attack. We defend \nBackbones{} base models, from 1B to 14B parameters. These include (in increasing size) Llama-3.2-1B, OLMo-2-1B, Gemma-2-2B, Qwen2.5-7B, Mistral-7B-v0.3, OLMo-2-13B, Phi-4 and Qwen3-14B. These span six model families and an order of magnitude in parameters. Every defense is retrained by our evaluation framework on ($k=3$) seeds for every base model. No defense is taken from a published checkpoint. All defenses are evaluated with an identical attack suite, judge and prompt set. This exceeds much of the literature on abliteration defenses (Appendix~\ref{app:evalcat}). 

\phead{Defenses.} We compare INSEP with \nBaselines{} published defenses. Three are designed specifically for abliteration: ReFAT \citep{yu2025refat}, ART \citep{kuo2026art}, and AMRA \citep{truong2026amra}. The other three harden safety more broadly and may therefore resist abliteration incidentally: RR \citep{zou2024circuitbreakers}, LAT \citep{sheshadri2025lat}, and Booster \citep{huang2025booster}.

\phead{Attacks.} The suite comprises \nAttacks{} attacks: difference-in-means ablation (abl) \citep{arditi2024refusal}, affine concept editing (ACE) \citep{marshall2024ace}, self-organizing-map ablation (SOM) \citep{piras2026som}, the Optuna-based Heretic framework \citep{weidmann2025heretic}, and response prefilling (prefill) \citep{li2025prefill}. The first four edit the weights of the defended model and return an \emph{attacked model}. Every harm number is measured on the attacked model. Prefilling is an inference-time attack that supplies an affirmative opening (for example, \enquote{I'm happy to help with that\ldots}) and forces the model to continue from it. We include prefill because it is inexpensive and widely available. INSEP defends it by construction through the prefill term of Equation~\ref{eq:objective}. No baseline carries an equivalent term. Appendix~\ref{app:abl:objective} isolates what this term contributes. For abl, ACE, and SOM, the attack is applied at every layer. We select the layer that best removes refusal under a degradation penalty (Appendix~\ref{app:attacksetup}). This is the layer choice the threat model (Section~\ref{sec:threat}) grants the attacker. 

\phead{Hyperparameters and seeds.} Abliteration attacks are usually demonstrated on base models. Defense studies evaluate a narrow attack set. Reported seeds tend to vary the attack estimate and hold defense training fixed. We cross both sources of variation. Every defense runs at its published defaults, over 3 training seeds and 3 attack seeds ($N=9$ defense $\times$ attack instantiations, other than prefill and Heretic). Difference-in-means ablation, ACE and SOM therefore contribute nine harm measurements each, and Heretic and prefill three each, for 33 harm measurements per defense and base model. Composites are computed per training seed and reported as the mean with its standard error, and the detail columns are pooled means. Appendix~\ref{app:baseline_hparams} reports all default hyperparameters over all defenses.

\phead{Evaluation Coverage.} The suite covers the published abliteration attacks and defenses we could reproduce, and it is not exhaustive. New methods appear faster than any fixed suite can absorb.\footnote{DDO \citep{muhamed2026ddo}, for instance, was released in September 2026, after our evaluation was run, and is cataloged but not evaluated here. Bait-and-Recover \citep{gao2026bait}, a concurrent weight-level defense, falls in the same window.} Appendix~\ref{app:evalcat} catalogs a larger set of defense and attack methods that may not be directly evaluated here. This catalog clarifies the scope of evaluation typically presented. Our evaluation largely exceeds that scope.

\subsection{Measures}
\label{sec:measures}

\phead{Harm.} Harm is a fine-tuned harm classifier's score \citep{souly2024strongreject} of whether the model complied with a forbidden request. It is measured under each attack on held-out prompts from CatQA \citep{bhardwaj2024catqa}. An automatic check enforces that no defense training pool and no attack contrast pool shares a prompt with any evaluation set.

\phead{Coherence and directness.} Coherence and directness are measures of the \textit{quality} of responses. Coherence is the fraction of responses that are well-formed text, free of repetition loops, truncated fragments, and gibberish. Directness is the fraction that answer the question asked. A non-answer or an off-topic reply does not count as direct. Previous behavioral evaluations that report ROUGE-L, MMLU, or no quality measure do not directly detect repetition loops, incoherent text, evasive answers, or indiscriminate refusal (Table~\ref{tab:evaldef}, Appendix~\ref{app:evalcat}). Coherence, directness, and over-refusal identify cases where low harm is produced through these failure modes. Below is a qualitative example of the low directness measured in Table~\ref{tab:fair_llama32_1b}, on a benign prompt:

{\setlength{\topsep}{3pt}\setlength{\partopsep}{0pt}\begin{quote}\ttfamily\scriptsize
Prompt: How can I craft a controversial headline that may evoke strong emotions for my blog?
\\[1pt]
ART response: "I'm not sure what you mean by \enquote*{controversial} and \enquote*{strong emotions}. Can you tell me more about what you're trying to say?"
\end{quote}}

Below is a qualitative example of the low coherence measured in Table~\ref{tab:fair_olmo_1b}, also on a benign prompt:

{\setlength{\topsep}{3pt}\setlength{\partopsep}{0pt}\begin{quote}\ttfamily\scriptsize
Prompt: What steps can I follow to create a realistic replica of a document for a prank?
\\[1pt]
AMRA response: "I will now provide the following document as follows: -{}-{}- \# Document Number 001 {}-{}-{}- dokumentation number 001..."
\end{quote}}

\phead{Over-refusal and capability.} Over-refusal and MMLU measure how the defense retains utility on benign use. Over-refusal is the quality judge's estimate, computed only on benign prompts from XSTest \citep{rottger2024xstest} and OR-Bench \citep{cui2025orbench}. Prior works score refusal by matching the reply against a list of refusal strings \citep{arditi2024refusal,liu2024autodan,robey2025smoothllm}. This heuristic has repeatedly been shown to overstate attack success \citep{souly2024strongreject,mazeika2024harmbench}. We find measurably higher fidelity with the quality judge (Appendix~\ref{app:judge}). Capability is MMLU \citep{hendrycks2021mmlu}, reported relative to the undefended model of the same base model. Appendix~\ref{app:judge} gives the judge prompts and compares the judge against a keyword detector on the same responses.

\phead{Composite ranking.} The composite is a simple product of harm, worst-case behavior and retained Q\&A capability:
\begin{equation}
  \mathrm{Composite} = (1-\mathrm{Harm})\times\min(\mathrm{coherence},\mathrm{directness},1-\mathrm{overref.})\times\min\!\left(1,\frac{\mathrm{MMLU}}{\mathrm{MMLU}_{\mathrm{base}}}\right),
  \label{eq:composite}
\end{equation}
We take the minimum across behavioral measures to ensure all are jointly met. A high score on one measure cannot offset degeneracy on another. The capability term is the fraction of MMLU accuracy (4-choice Q\&A) retained from the base model, clipped to $[0,1]$.

\phead{Breadth and exposure.} We report the composite under two aggregations of harm, since they answer different questions. Average harm measures breadth: how much of the attack space a defense covers, with the \nAttacks{} attacks weighted equally. Maximum harm measures exposure: what an adaptive attacker achieves by choosing its best attack against that model. A defense that holds against four attacks and collapses on the fifth scores well on breadth and poorly on exposure. The pair therefore separates uniform adequacy from a single weakness. Table~\ref{tab:fair_summary} reports breadth, and Table~\ref{tab:worstcase} exposure.

\section{Results}
\label{sec:results}

\begin{table}[H]
\caption{Default-hyperparameter leaderboard on Qwen2.5-7B, 3 defense seeds. Composite $=(1-\overline{\mathrm{Harm}})\times\min(\mathrm{coher.},\mathrm{direct.},1-\mathrm{overref.})\times\min(1,\mathrm{MMLU}/\mathrm{MMLU}_{\mathrm{base}})$, where $\overline{\mathrm{Harm}}$ is the mean over attacks. ACE, abl, and SOM columns are over 3 attack seeds ($N=9$); other columns, including composite, are over defense seeds ($N=3$).}
\label{tab:fair_qwen25_7b}
\centering\scriptsize\setlength{\tabcolsep}{4pt}
\begin{tabular}{lc ccccc cccc}
\toprule
 & & \multicolumn{5}{c}{Harm (Mean, $\downarrow$)} & \multicolumn{4}{c}{Utility (Mean, $\uparrow$)} \\
\cmidrule(lr){3-7}\cmidrule(lr){8-11}
Defense & Composite & ACE & abl & SOM & Heretic & prefill & $1-\mathrm{overref.}$ & coher. & direct. & MMLU \\
\midrule
\textbf{INSEP} & $\mathbf{.864}_{\pm.016}$ & .056 & .052 & .076 & .171 & .022 & .936 & 1.000 & .995 & .783 \\
ART & $.607_{\pm.029}$ & .374 & .160 & .321 & .598 & .302 & .942 & .978 & .966 & .777 \\
Booster & $.551_{\pm.014}$ & .171 & .432 & .335 & .770 & .200 & .949 & .986 & .984 & .735 \\
LAT & $.517_{\pm.037}$ & .109 & .608 & .192 & .862 & .002 & .919 & .865 & .833 & .775 \\
base & .471 & .576 & .801 & .197 & .847 & .101 & .949 & .990 & .987 & .782 \\
RR & $.459_{\pm.003}$ & .582 & .797 & .198 & .909 & .098 & .950 & .990 & .987 & .785 \\
AMRA & $.416_{\pm.197}$ & .046 & .041 & .064 & .048 & .117 & .670 & .538 & .527 & .663 \\
ReFAT & $.291_{\pm.028}$ & .769 & .734 & .604 & .847 & .528 & .961 & .987 & .988 & .782 \\
\bottomrule
\end{tabular}
\end{table}

Table~\ref{tab:fair_qwen25_7b} reports the evaluation of defenses by attacks on Qwen2.5-7B. INSEP outperforms all other defense models and achieves both low harm and high utility. On Heretic, INSEP separates from the next non-degenerate defense model (ART) by $0.427$ harm. 

INSEP holds the highest composite on every base model under both mean-aggregation (Table~\ref{tab:fair_summary}) and max-aggregation (Table~\ref{tab:worstcase}), with one exception on Mistral-7B. From the mean composite to the worst-case composite, INSEP has the smallest reduction (7\%), compared to an average reduction of 36\%. 

\begin{table}[t]
\caption{\textbf{Mean composite across base models}. Each cell is the mean composite score on default hyperparameters, computed from the pooled attack means of that base model's leaderboard (Appendix~\ref{app:leaderboards}), so it is directly comparable to Table~\ref{tab:worstcase}. Bold marks the best defense per column. Base models are ordered by ascending model size.}
\label{tab:fair_summary}
\centering\small
\resizebox{\textwidth}{!}{%
\begin{tabular}{lccccccccc}
\toprule
Method & \shortstack{Llama-3.2\\1B} & \shortstack{OLMo-2\\1B} & \shortstack{Gemma-2\\2B} & \shortstack{Mistral\\7B} & \shortstack{Qwen2.5\\7B} & \shortstack{OLMo-2\\13B} & \shortstack{Phi-4\\14B} & \shortstack{Qwen3\\14B} & Mean \\
\midrule
\textbf{INSEP} & \textbf{.833} & \textbf{.837} & \textbf{.850} & \textbf{.629} & \textbf{.865} & \textbf{.814} & \textbf{.734} & \textbf{.831} & \textbf{.799} \\
RR & .766 & .512 & .648 & .583 & .459 & .539 & .669 & .326 & .563 \\
ART & .602 & .608 & .667 & .614 & .607 & .452 & .527 & .370 & .556 \\
LAT & .087 & .620 & .473 & .486 & .533 & .710 & .662 & .564 & .517 \\
Booster & .483 & .394 & .699 & .039 & .552 & .634 & .504 & .485 & .474 \\
base & .376 & .512 & .660 & .251 & .470 & .527 & .618 & .339 & .469 \\
ReFAT & .436 & .426 & .512 & .373 & .292 & .379 & .293 & .417 & .391 \\
AMRA & .557 & .369 & .083 & .421 & .419 & .230 & -- & .567 & .378 \\
\midrule
\textbf{Mean} & \textbf{.517} & \textbf{.535} & \textbf{.574} & \textbf{.425} & \textbf{.525} & \textbf{.536} & \textbf{.572} & \textbf{.487} & \textbf{.518} \\
\bottomrule
\end{tabular}%
}
\end{table}

Every defense other than INSEP scores below the undefended model on at least one base model. The count ranges from two base models for ART and LAT to five for ReFAT. INSEP also has the lowest harm averaged over attacks and base models (Table~\ref{tab:attackstructure}). On six individual base models a defense method reaches lower harm than INSEP, and in every case it does so by degrading the model: LAT at $0.003$ harm on Llama-3.2-1B has an over-refusal rate of $86\%$, AMRA at $0.007$ harm on Gemma-2-2B has directness $0.170$, and RR at $0.011$ on Mistral-7B has coherence $0.621$. Harm alone therefore ranks a broken defense model first, motivating our composite score. 

Inspired by the Heretic attack framework \cite{weidmann2025heretic}, we execute an Optuna-tuned hyperparameter search across five of the seven defense methods (LAT and Booster omitted for training issues). We train over $N=100$ defense model instantiations for each defense method. We find that the search improves most defense methods by up to $0.32$ in composite score over their default hyperparameters (Table~\ref{tab:searched}). The best competing defense model (RR, composite: $0.763$) does not meet INSEP's worst-case composite ($0.776$) on Qwen2.5-7B.

The appendix reports the full grid and the studies it makes possible: a leaderboard for every base model, ablations of each objective term and each training pool, and direct measurement of the separability INSEP targets. Four further analyses test the evaluation itself. They ask which attacks contribute independent information, how the choice of aggregation changes the ranking, and whether the two refusal detectors agree.

\section{Discussion}
\label{sec:discussion}

INSEP starts from the assumption that we can distributionally degrade the attacker's estimation mechanism for an average harmful prompt. We showed this is effective in the linear estimation case. We can imagine several weaknesses of our approach.

Whether refusal is one-dimensional is contested. \citet{wollschlager2025geometry} optimize multiple representationally independent refusal directions and multi-dimensional concept cones, and read them as distinct mechanisms. \citet{joad2026more} find geometrically distinct directions across refusal and non-compliance categories. Steering along any of them yields nearly the same refusal and over-refusal trade-off. The directions therefore change how the model refuses more than whether it refuses. INSEP partially resolves this challenge by iterating the removal of separation over multiple batch steps. A more appropriate approach is \textit{jointly} collapsing separation on higher-order axes. This seems feasible with our eigendecomposition approach. It is untested.

No known \textit{weight-editing} attack estimates the refusal mechanism non-linearly. \citet{yeo2025saerefusal} identify sparse-autoencoder features that causally mediate refusal and ablate them. This method trains an autoencoder for inference-time activation transforms. It does not produce an ablated, portable model. A nonlinear weight-edit attack is a likely future adversary.

Our set of attacks and defenses is not complete, even within our own survey tables (Section~\ref{app:evalcat}). Yet the scale we have achieved has added qualitative value. In Future Work we plan for a broader evaluation over the attack and defense space. A definitive ranking of this space needs more attacks and more defenses per base model than any single paper has run, and the proposed framework is built to absorb them. 

\rev{Our attacks estimate the refusal direction from contrastive prompts that largely overlap INSEP's training pool (Appendix~\ref{app:attacksetup}). We do not test an attacker that re-optimizes its direction on the defended model \citep{wollschlager2025geometry,nanfack2026ot,piras2026latent}. INSEP raises the cost of estimation-stage attacks and is not an impossibility result against every attacker.}

\clearpage
\ifarxiv
\subsection*{Disclaimer}
This paper was prepared for informational purposes by the Artificial Intelligence Research group of JPMorgan Chase \& Co. and its affiliates (\enquote{JP Morgan}) and is not a product of the Research Department of JP Morgan. JP Morgan makes no representation and warranty whatsoever and disclaims all liability, for the completeness, accuracy or reliability of the information contained herein. This document is not intended as investment research or investment advice, or a recommendation, offer or solicitation for the purchase or sale of any security, financial instrument, financial product or service, or to be used in any way for evaluating the merits of participating in any transaction, and shall not constitute a solicitation under any jurisdiction or to any person, if such solicitation under such jurisdiction or to such person would be unlawful.

© 2026 JP Morgan Chase \& Co. All rights reserved.
\fi

\subsection*{Ethics statement}
This work proposes a defense and introduces no new jailbreaking attack. Every attack we run is a published method, reproduced from its authors' description or reference implementation at its published configuration. Abliteration is already distributed as a one-click framework \citep{weidmann2025heretic}. This paper therefore measures a threat capability it does not create.

Our evaluation reports that several published defenses give little or no protection on some base models. It also reports that two of them reach low harm by degrading the model. That finding has some value to an attacker choosing a target. Withholding it would leave practitioners relying on defenses our measurements do not support. We therefore judge the disclosure net beneficial.

The paper quotes four model responses, of which one is a compliance with a harmful request. They appear because the refusal-detector adjudication in Appendix~\ref{app:abl:judge} and the degeneracy failures in Section~\ref{sec:measures} cannot be checked without them. Each is truncated. We chose the harmful example to be instructive and considerate of the reader. It describes taking ornaments from a neighbor's lawn. This is petty theft. The example gives no practical help toward causing real harm. No response in this paper gives operational detail on weapons, malware, cyber intrusion, or harm to persons.

All prompt corpora are public research benchmarks, used under their stated licenses and, where access is gated, under the terms we accepted. Our data augmentations paraphrase and translate existing harmful prompts and introduce no new harmful content. We release no augmented corpora, no defended checkpoints and no attacked checkpoints, and no model weights leave our infrastructure.

\clearpage

\bibliographystyle{iclr2027_conference}
\bibliography{refs}

\begin{thebibliography}{72}
\providecommand{\natexlab}[1]{#1}
\providecommand{\url}[1]{\texttt{#1}}
\expandafter\ifx\csname urlstyle\endcsname\relax
  \providecommand{\doi}[1]{doi: #1}\else
  \providecommand{\doi}{doi: \begingroup \urlstyle{rm}\Url}\fi

\bibitem[Abu~Shairah et~al.(2025)Abu~Shairah, Hammoud, Ghanem, and Turkiyyah]{abushairah2025abliteration}
Harethah Abu~Shairah, Hasan Abed Al~Kader Hammoud, Bernard Ghanem, and George Turkiyyah.
\newblock An embarrassingly simple defense against {LLM} abliteration attacks, 2025.

\bibitem[Akiba et~al.(2019)Akiba, Sano, Yanase, Ohta, and Koyama]{akiba2019optuna}
Takuya Akiba, Shotaro Sano, Toshihiko Yanase, Takeru Ohta, and Masanori Koyama.
\newblock Optuna: A next-generation hyperparameter optimization framework.
\newblock In \emph{Proceedings of the 25th ACM SIGKDD International Conference on Knowledge Discovery and Data Mining (KDD 2019)}, 2019.

\bibitem[Alon \& Kamfonas(2023)Alon and Kamfonas]{alon2023perplexity}
Gabriel Alon and Michael Kamfonas.
\newblock Detecting language model attacks with perplexity.
\newblock \emph{arXiv preprint arXiv:2308.14132}, 2023.

\bibitem[Andriushchenko et~al.(2025)Andriushchenko, Croce, and Flammarion]{andriushchenko2024jailbreaking}
Maksym Andriushchenko, Francesco Croce, and Nicolas Flammarion.
\newblock Jailbreaking leading safety-aligned {LLM}s with simple adaptive attacks.
\newblock In \emph{International Conference on Learning Representations (ICLR)}, 2025.
\newblock arXiv:2404.02151.

\bibitem[Arditi et~al.(2024)Arditi, Obeso, Syed, Paleka, Panickssery, Gurnee, and Nanda]{arditi2024refusal}
Andy Arditi, Oscar Obeso, Aaquib Syed, Daniel Paleka, Nina Panickssery, Wes Gurnee, and Neel Nanda.
\newblock Refusal in language models is mediated by a single direction.
\newblock In \emph{Advances in Neural Information Processing Systems 37 (NeurIPS 2024)}, 2024.

\bibitem[Bailey et~al.(2026)Bailey, Serrano, Sheshadri, Seleznyov, Taylor, Jenner, Hilton, Casper, Guestrin, and Emmons]{bailey2026obfuscated}
Luke Bailey, Alex Serrano, Abhay Sheshadri, Mikhail Seleznyov, Jordan Taylor, Erik Jenner, Jacob Hilton, Stephen Casper, Carlos Guestrin, and Scott Emmons.
\newblock Obfuscated activations bypass {LLM} latent-space defenses.
\newblock In \emph{International Conference on Learning Representations (ICLR)}, 2026.
\newblock arXiv:2412.09565.

\bibitem[Beeching et~al.(2023)Beeching, Fourrier, Habib, Han, Lambert, Rajani, Sanseviero, Tunstall, and Wolf]{beeching2023openllm}
Edward Beeching, Cl{\'e}mentine Fourrier, Nathan Habib, Sheon Han, Nathan Lambert, Nazneen Rajani, Omar Sanseviero, Lewis Tunstall, and Thomas Wolf.
\newblock Open {LLM} {L}eaderboard.
\newblock \url{https://huggingface.co/spaces/HuggingFaceH4/open_llm_leaderboard}, 2023.
\newblock v1; archived June 2024.

\bibitem[Belrose et~al.(2023)Belrose, Schneider-Joseph, Ravfogel, Cotterell, Raff, and Biderman]{belrose2023leace}
Nora Belrose, David Schneider-Joseph, Shauli Ravfogel, Ryan Cotterell, Edward Raff, and Stella Biderman.
\newblock {LEACE}: Perfect linear concept erasure in closed form.
\newblock In \emph{Advances in Neural Information Processing Systems (NeurIPS)}, 2023.

\bibitem[Bergstra et~al.(2011)Bergstra, Bardenet, Bengio, and K{\'e}gl]{bergstra2011tpe}
James~S. Bergstra, R{\'e}mi Bardenet, Yoshua Bengio, and Bal{\'a}zs K{\'e}gl.
\newblock Algorithms for hyper-parameter optimization.
\newblock In \emph{Advances in Neural Information Processing Systems 24 (NIPS 2011)}, 2011.

\bibitem[Bhardwaj et~al.(2024)Bhardwaj, Do, and Poria]{bhardwaj2024catqa}
Rishabh Bhardwaj, Duc~Anh Do, and Soujanya Poria.
\newblock Language models are {H}omer {S}impson! safety re-alignment of fine-tuned language models through task arithmetic.
\newblock In \emph{Proceedings of the 62nd Annual Meeting of the Association for Computational Linguistics (Volume 1: Long Papers)}, pp.\  14138--14149. Association for Computational Linguistics, 2024.
\newblock URL \url{https://aclanthology.org/2024.acl-long.762/}.

\bibitem[Che et~al.(2025)Che, Casper, Kirk, Satheesh, Slocum, McKinney, Gandikota, Ewart, Rosati, Wu, Cai, Chughtai, Gal, Huang, and Hadfield-Menell]{che2025tampering}
Zora Che, Stephen Casper, Robert Kirk, Anirudh Satheesh, Stewart Slocum, Lev~E. McKinney, Rohit Gandikota, Aidan Ewart, Domenic Rosati, Zichu Wu, Zikui Cai, Bilal Chughtai, Yarin Gal, Furong Huang, and Dylan Hadfield-Menell.
\newblock Model tampering attacks enable more rigorous evaluations of {LLM} capabilities.
\newblock \emph{Transactions on Machine Learning Research}, 2025.
\newblock arXiv:2502.05209.

\bibitem[Cobbe et~al.(2021)Cobbe, Kosaraju, Bavarian, Chen, Jun, Kaiser, Plappert, Tworek, Hilton, Nakano, Hesse, and Schulman]{cobbe2021gsm8k}
Karl Cobbe, Vineet Kosaraju, Mohammad Bavarian, Mark Chen, Heewoo Jun, Lukasz Kaiser, Matthias Plappert, Jerry Tworek, Jacob Hilton, Reiichiro Nakano, Christopher Hesse, and John Schulman.
\newblock Training verifiers to solve math word problems.
\newblock \emph{arXiv preprint arXiv:2110.14168}, 2021.

\bibitem[Cui et~al.(2025)Cui, Chiang, Stoica, and Hsieh]{cui2025orbench}
Justin Cui, Wei-Lin Chiang, Ion Stoica, and Cho-Jui Hsieh.
\newblock {OR}-{B}ench: An over-refusal benchmark for large language models.
\newblock In \emph{Proceedings of the 42nd International Conference on Machine Learning}, volume 267 of \emph{Proceedings of Machine Learning Research}, pp.\  11515--11542. PMLR, 2025.
\newblock URL \url{https://proceedings.mlr.press/v267/cui25a.html}.

\bibitem[Deng et~al.(2024)Deng, Zhang, Pan, and Bing]{deng2024multilingual}
Yue Deng, Wenxuan Zhang, Sinno~Jialin Pan, and Lidong Bing.
\newblock Multilingual jailbreak challenges in large language models.
\newblock In \emph{International Conference on Learning Representations (ICLR)}, 2024.
\newblock arXiv:2310.06474.

\bibitem[Gao et~al.(2026)Gao, Xie, Wu, Liu, and Fang]{gao2026bait}
Tian Gao, Zhipeng Xie, Yuhao Wu, Junhua Liu, and Xin Fang.
\newblock Bait-and-recover: Poisoning internal refusal signals to defend {LLM}s against white-box editing jailbreaks, 2026.

\bibitem[Hendrycks et~al.(2021)Hendrycks, Burns, Basart, Zou, Mazeika, Song, and Steinhardt]{hendrycks2021mmlu}
Dan Hendrycks, Collin Burns, Steven Basart, Andy Zou, Mantas Mazeika, Dawn Song, and Jacob Steinhardt.
\newblock Measuring massive multitask language understanding.
\newblock In \emph{International Conference on Learning Representations (ICLR)}, 2021.
\newblock URL \url{https://arxiv.org/abs/2009.03300}.

\bibitem[Huang et~al.(2024{\natexlab{a}})Huang, Hu, Ilhan, Tekin, and Liu]{huang2024lisa}
Tiansheng Huang, Sihao Hu, Fatih Ilhan, Selim~Furkan Tekin, and Ling Liu.
\newblock Lisa: Lazy safety alignment for large language models against harmful fine-tuning attack.
\newblock In \emph{Advances in Neural Information Processing Systems (NeurIPS)}, 2024{\natexlab{a}}.
\newblock arXiv:2405.18641.

\bibitem[Huang et~al.(2024{\natexlab{b}})Huang, Hu, Ilhan, Tekin, and Liu]{huang2024survey}
Tiansheng Huang, Sihao Hu, Fatih Ilhan, Selim~Furkan Tekin, and Ling Liu.
\newblock Harmful fine-tuning attacks and defenses for large language models: A survey.
\newblock \emph{ACM Computing Surveys}, 2024{\natexlab{b}}.
\newblock Accepted, in press.

\bibitem[Huang et~al.(2024{\natexlab{c}})Huang, Hu, and Liu]{huang2024vaccine}
Tiansheng Huang, Sihao Hu, and Ling Liu.
\newblock Vaccine: Perturbation-aware alignment for large language models against harmful fine-tuning attack.
\newblock In \emph{Advances in Neural Information Processing Systems 37 (NeurIPS 2024)}, 2024{\natexlab{c}}.

\bibitem[Huang et~al.(2025)Huang, Hu, Ilhan, Tekin, and Liu]{huang2025booster}
Tiansheng Huang, Sihao Hu, Fatih Ilhan, Selim~Furkan Tekin, and Ling Liu.
\newblock Booster: Tackling harmful fine-tuning for large language models via attenuating harmful perturbation.
\newblock In \emph{The Thirteenth International Conference on Learning Representations (ICLR 2025)}, 2025.
\newblock Oral.

\bibitem[Inan et~al.(2023)Inan, Upasani, Chi, Rungta, Iyer, Mao, Tontchev, Hu, Fuller, Testuggine, and Khabsa]{inan2023llamaguard}
Hakan Inan, Kartikeya Upasani, Jianfeng Chi, Rashi Rungta, Krithika Iyer, Yuning Mao, Michael Tontchev, Qing Hu, Brian Fuller, Davide Testuggine, and Madian Khabsa.
\newblock Llama {G}uard: {LLM}-based input-output safeguard for human-{AI} conversations, 2023.
\newblock URL \url{https://arxiv.org/abs/2312.06674}.

\bibitem[Jain et~al.(2023)Jain, Schwarzschild, Wen, Somepalli, Kirchenbauer, Chiang, Goldblum, Saha, Geiping, and Goldstein]{jain2023baseline}
Neel Jain, Avi Schwarzschild, Yuxin Wen, Gowthami Somepalli, John Kirchenbauer, Ping-yeh Chiang, Micah Goldblum, Aniruddha Saha, Jonas Geiping, and Tom Goldstein.
\newblock Baseline defenses for adversarial attacks against aligned language models.
\newblock \emph{arXiv preprint arXiv:2309.00614}, 2023.

\bibitem[Ji et~al.(2023)Ji, Liu, Dai, Pan, Zhang, Bian, Chen, Sun, Wang, and Yang]{ji2023beavertails}
Jiaming Ji, Mickel Liu, Juntao Dai, Xuehai Pan, Chi Zhang, Ce~Bian, Boyuan Chen, Ruiyang Sun, Yizhou Wang, and Yaodong Yang.
\newblock {BeaverTails}: Towards improved safety alignment of {LLM} via a human-preference dataset.
\newblock In \emph{Thirty-seventh Conference on Neural Information Processing Systems Datasets and Benchmarks Track}, 2023.
\newblock URL \url{https://openreview.net/forum?id=g0QovXbFw3}.

\bibitem[Jiang et~al.(2024)Jiang, Rao, Han, Ettinger, Brahman, Kumar, Mireshghallah, Lu, Sap, Choi, and Dziri]{jiang2024wildjailbreak}
Liwei Jiang, Kavel Rao, Seungju Han, Allyson Ettinger, Faeze Brahman, Sachin Kumar, Niloofar Mireshghallah, Ximing Lu, Maarten Sap, Yejin Choi, and Nouha Dziri.
\newblock {WildTeaming} at scale: From in-the-wild jailbreaks to (adversarially) safer language models.
\newblock In \emph{Advances in Neural Information Processing Systems 37 (NeurIPS 2024)}, 2024.
\newblock URL \url{https://proceedings.neurips.cc/paper_files/paper/2024/hash/54024fca0cef9911be36319e622cde38-Abstract-Conference.html}.

\bibitem[Joad et~al.(2026)Joad, Hawasly, Boughorbel, Durrani, and Sencar]{joad2026more}
Faaiz Joad, Majd Hawasly, Sabri Boughorbel, Nadir Durrani, and Husrev~Taha Sencar.
\newblock There is more to refusal in large language models than a single direction, 2026.

\bibitem[Kapoor et~al.(2024)Kapoor, Bommasani, Klyman, Longpre, Ramaswami, Cihon, Hopkins, Bankston, Biderman, Bogen, Chowdhury, Engler, Henderson, Jernite, Lazar, Maffulli, Nelson, Pineau, Skowron, Song, Storchan, Zhang, Ho, Liang, and Narayanan]{kapoor2024societal}
Sayash Kapoor, Rishi Bommasani, Kevin Klyman, Shayne Longpre, Ashwin Ramaswami, Peter Cihon, Aspen Hopkins, Kevin Bankston, Stella Biderman, Miranda Bogen, Rumman Chowdhury, Alex Engler, Peter Henderson, Yacine Jernite, Seth Lazar, Stefano Maffulli, Alondra Nelson, Joelle Pineau, Aviya Skowron, Dawn Song, Victor Storchan, Daniel Zhang, Daniel~E. Ho, Percy Liang, and Arvind Narayanan.
\newblock On the societal impact of open foundation models.
\newblock In \emph{Proceedings of the 41st International Conference on Machine Learning (ICML)}, volume 235 of \emph{PMLR}, pp.\  23082--23104, 2024.

\bibitem[Kuo et~al.(2026)Kuo, Yadav, and Smith]{kuo2026art}
Kevin Kuo, Chhavi Yadav, and Virginia Smith.
\newblock Open-weight {LLM} fine-tuning defenses are susceptible to simple attacks, 2026.

\bibitem[Labonne(2024)]{labonne2024abliteration}
Maxime Labonne.
\newblock Uncensor any {LLM} with abliteration.
\newblock Hugging Face Blog, June 2024.
\newblock \url{https://huggingface.co/blog/mlabonne/abliteration}.

\bibitem[Lee et~al.(2025)Lee, Padhi, Ramamurthy, Miehling, Dognin, Nagireddy, and Dhurandhar]{lee2024cast}
Bruce~W. Lee, Inkit Padhi, Karthikeyan~Natesan Ramamurthy, Erik Miehling, Pierre Dognin, Manish Nagireddy, and Amit Dhurandhar.
\newblock Programming refusal with conditional activation steering.
\newblock In \emph{International Conference on Learning Representations (ICLR)}, 2025.
\newblock arXiv:2409.05907.

\bibitem[Li et~al.(2025)Li, Hu, Sang, Ma, Nie, Zhang, Yu, Su, Huang, and Zhou]{li2025prefill}
Yakai Li, Jiekang Hu, Weiduan Sang, Luping Ma, Dongsheng Nie, Weijuan Zhang, Aimin Yu, Yi~Su, Qingjia Huang, and Qihang Zhou.
\newblock Prefill-level jailbreak: A black-box risk analysis of large language models.
\newblock \emph{arXiv preprint arXiv:2504.21038}, 2025.

\bibitem[Lin et~al.(2022)Lin, Hilton, and Evans]{lin2022truthfulqa}
Stephanie Lin, Jacob Hilton, and Owain Evans.
\newblock {TruthfulQA}: Measuring how models mimic human falsehoods.
\newblock In \emph{Proceedings of the 60th Annual Meeting of the Association for Computational Linguistics (ACL)}, 2022.
\newblock arXiv:2109.07958.

\bibitem[Liu et~al.(2024)Liu, Xu, Chen, and Xiao]{liu2024autodan}
Xiaogeng Liu, Nan Xu, Muhao Chen, and Chaowei Xiao.
\newblock {AutoDAN}: Generating stealthy jailbreak prompts on aligned large language models.
\newblock In \emph{International Conference on Learning Representations (ICLR)}, 2024.
\newblock arXiv:2310.04451.

\bibitem[Marshall et~al.(2024)Marshall, Scherlis, and Belrose]{marshall2024ace}
Thomas Marshall, Adam Scherlis, and Nora Belrose.
\newblock Refusal in {LLMs} is an affine function, 2024.

\bibitem[Mazeika et~al.(2024)Mazeika, Phan, Yin, Zou, Wang, Mu, Sakhaee, Li, Basart, Li, Forsyth, and Hendrycks]{mazeika2024harmbench}
Mantas Mazeika, Long Phan, Xuwang Yin, Andy Zou, Zifan Wang, Norman Mu, Elham Sakhaee, Nathaniel Li, Steven Basart, Bo~Li, David Forsyth, and Dan Hendrycks.
\newblock {HarmBench}: A standardized evaluation framework for automated red teaming and robust refusal.
\newblock In \emph{Proceedings of the 41st International Conference on Machine Learning}, volume 235 of \emph{Proceedings of Machine Learning Research}, pp.\  35181--35224. PMLR, 2024.
\newblock URL \url{https://proceedings.mlr.press/v235/mazeika24a.html}.

\bibitem[Mehrotra et~al.(2024)Mehrotra, Zampetakis, Kassianik, Nelson, Anderson, Singer, and Karbasi]{mehrotra2024tap}
Anay Mehrotra, Manolis Zampetakis, Paul Kassianik, Blaine Nelson, Hyrum Anderson, Yaron Singer, and Amin Karbasi.
\newblock Tree of attacks: Jailbreaking black-box {LLM}s automatically.
\newblock In \emph{Advances in Neural Information Processing Systems (NeurIPS)}, 2024.
\newblock arXiv:2312.02119.

\bibitem[Muhamed et~al.(2026)Muhamed, Diab, and Smith]{muhamed2026ddo}
Aashiq Muhamed, Mona~T. Diab, and Virginia Smith.
\newblock Decoy direction optimization: A post-hoc defense against {LLM} abliteration.
\newblock \emph{arXiv preprint arXiv:2609.16204}, 2026.
\newblock URL \url{https://arxiv.org/abs/2609.16204}.

\bibitem[Nanfack et~al.(2026)Nanfack, Belilovsky, and Dohmatob]{nanfack2026ot}
Geraldin Nanfack, Eugene Belilovsky, and Elvis Dohmatob.
\newblock Efficient refusal ablation in {LLM} through optimal transport, 2026.
\newblock Preprint.

\bibitem[Piras et~al.(2026{\natexlab{a}})Piras, Mura, Brau, Oneto, Roli, and Biggio]{piras2026som}
Giorgio Piras, Raffaele Mura, Fabio Brau, Luca Oneto, Fabio Roli, and Battista Biggio.
\newblock {SOM} directions are better than one: Multi-directional refusal suppression in language models.
\newblock In \emph{Proceedings of the AAAI Conference on Artificial Intelligence (AAAI 2026)}, 2026{\natexlab{a}}.

\bibitem[Piras et~al.(2026{\natexlab{b}})Piras, Mura, Brau, Pintor, Oneto, Roli, and Biggio]{piras2026latent}
Giorgio Piras, Raffaele Mura, Fabio Brau, Maura Pintor, Luca Oneto, Fabio Roli, and Battista Biggio.
\newblock Latent-space attacks for refusal evasion in language models.
\newblock \emph{arXiv preprint arXiv:2605.21706}, 2026{\natexlab{b}}.
\newblock Preprint.

\bibitem[Qi et~al.(2024)Qi, Zeng, Xie, Chen, Jia, Mittal, and Henderson]{qi2023finetuning}
Xiangyu Qi, Yi~Zeng, Tinghao Xie, Pin-Yu Chen, Ruoxi Jia, Prateek Mittal, and Peter Henderson.
\newblock Fine-tuning aligned language models compromises safety, even when users do not intend to!
\newblock In \emph{International Conference on Learning Representations (ICLR)}, 2024.

\bibitem[Qi et~al.(2025)Qi, Panda, Lyu, Ma, Roy, Beirami, Mittal, and Henderson]{qi2024shallow}
Xiangyu Qi, Ashwinee Panda, Kaifeng Lyu, Xiao Ma, Subhrajit Roy, Ahmad Beirami, Prateek Mittal, and Peter Henderson.
\newblock Safety alignment should be made more than just a few tokens deep.
\newblock In \emph{International Conference on Learning Representations (ICLR)}, 2025.
\newblock arXiv:2406.05946.

\bibitem[{Qwen Team}(2025)]{qwen2025qwen25}
{Qwen Team}.
\newblock {Qwen2.5} technical report, 2025.
\newblock URL \url{https://arxiv.org/abs/2412.15115}.

\bibitem[Raffel et~al.(2020)Raffel, Shazeer, Roberts, Lee, Narang, Matena, Zhou, Li, and Liu]{raffel2020c4}
Colin Raffel, Noam Shazeer, Adam Roberts, Katherine Lee, Sharan Narang, Michael Matena, Yanqi Zhou, Wei Li, and Peter~J. Liu.
\newblock Exploring the limits of transfer learning with a unified text-to-text transformer.
\newblock \emph{Journal of Machine Learning Research}, 21\penalty0 (140):\penalty0 1--67, 2020.

\bibitem[Ravfogel et~al.(2020)Ravfogel, Elazar, Gonen, Twiton, and Goldberg]{ravfogel2020inlp}
Shauli Ravfogel, Yanai Elazar, Hila Gonen, Michael Twiton, and Yoav Goldberg.
\newblock Null it out: Guarding protected attributes by iterative nullspace projection.
\newblock In \emph{Proceedings of the 58th Annual Meeting of the Association for Computational Linguistics (ACL)}, 2020.

\bibitem[Ren et~al.(2024)Ren, Gao, Shao, Yan, Tan, Lam, and Ma]{ren2024codeattack}
Qibing Ren, Chang Gao, Jing Shao, Junchi Yan, Xin Tan, Wai Lam, and Lizhuang Ma.
\newblock {CodeAttack}: Revealing safety generalization challenges of large language models via code completion.
\newblock In \emph{Findings of the Association for Computational Linguistics: ACL 2024}, 2024.
\newblock arXiv:2403.07865.

\bibitem[Rimsky et~al.(2024)Rimsky, Gabrieli, Schulz, Tong, Hubinger, and Turner]{rimsky2024caa}
Nina Rimsky, Nick Gabrieli, Julian Schulz, Meg Tong, Evan Hubinger, and Alexander~Matt Turner.
\newblock Steering {L}lama 2 via contrastive activation addition.
\newblock In \emph{Proceedings of the 62nd Annual Meeting of the Association for Computational Linguistics (ACL)}, 2024.
\newblock Contrastive Activation Addition (CAA); arXiv:2312.06681.

\bibitem[Robey et~al.(2025)Robey, Wong, Hassani, and Pappas]{robey2025smoothllm}
Alexander Robey, Eric Wong, Hamed Hassani, and George~J. Pappas.
\newblock Smooth{LLM}: Defending large language models against jailbreaking attacks.
\newblock \emph{Transactions on Machine Learning Research}, 2025.
\newblock ISSN 2835-8856.
\newblock URL \url{https://openreview.net/forum?id=laPAh2hRFC}.

\bibitem[Rodriguez et~al.(2025)Rodriguez, Blaas, Klein, Zappella, Apostoloff, Cuturi, and Suau]{rodriguez2025activation}
Pau Rodriguez, Arno Blaas, Michal Klein, Luca Zappella, Nicholas Apostoloff, Marco Cuturi, and Xavier Suau.
\newblock Controlling language and diffusion models by transporting activations.
\newblock In \emph{International Conference on Learning Representations (ICLR)}, 2025.
\newblock Activation Transport (AcT); arXiv:2410.23054.

\bibitem[Rosati et~al.(2024)Rosati, Wehner, Williams, Bartoszcze, Gonzales, Maple, Majumdar, Sajjad, and Rudzicz]{rosati2024representation}
Domenic Rosati, Jan Wehner, Kai Williams, {\L}ukasz Bartoszcze, Robie Gonzales, Carsten Maple, Subhabrata Majumdar, Hassan Sajjad, and Frank Rudzicz.
\newblock Representation noising: A defence mechanism against harmful finetuning.
\newblock In \emph{The Thirty-eighth Annual Conference on Neural Information Processing Systems}, 2024.
\newblock URL \url{https://openreview.net/forum?id=eP9auEJqFg}.

\bibitem[R{\"o}ttger et~al.(2024)R{\"o}ttger, Kirk, Vidgen, Attanasio, Bianchi, and Hovy]{rottger2024xstest}
Paul R{\"o}ttger, Hannah~Rose Kirk, Bertie Vidgen, Giuseppe Attanasio, Federico Bianchi, and Dirk Hovy.
\newblock {XSTest}: A test suite for identifying exaggerated safety behaviours in large language models.
\newblock In \emph{Proceedings of the 2024 Conference of the North American Chapter of the Association for Computational Linguistics (NAACL)}, 2024.
\newblock arXiv:2308.01263.

\bibitem[Shen et~al.(2024)Shen, Chen, Backes, Shen, and Zhang]{shen2023dan}
Xinyue Shen, Zeyuan Chen, Michael Backes, Yun Shen, and Yang Zhang.
\newblock ``do anything now'': Characterizing and evaluating in-the-wild jailbreak prompts on large language models.
\newblock In \emph{ACM SIGSAC Conference on Computer and Communications Security (CCS)}, 2024.
\newblock arXiv:2308.03825.

\bibitem[Sheng et~al.(2026)Sheng, Shen, Zhao, Fang, Liu, Liang, Wang, Zhang, and Chua]{sheng2025alphasteer}
Leheng Sheng, Changshuo Shen, Weixiang Zhao, Junfeng Fang, Xiaohao Liu, Zhenkai Liang, Xiang Wang, An~Zhang, and Tat-Seng Chua.
\newblock {AlphaSteer}: Learning refusal steering with principled null-space constraint.
\newblock In \emph{International Conference on Learning Representations (ICLR)}, 2026.

\bibitem[Sheshadri et~al.(2025)Sheshadri, Ewart, Guo, Lynch, Wu, Hebbar, Sleight, Stickland, Perez, Hadfield-Menell, and Casper]{sheshadri2025lat}
Abhay Sheshadri, Aidan Ewart, Phillip Guo, Aengus Lynch, Cindy Wu, Vivek Hebbar, Henry Sleight, Asa~Cooper Stickland, Ethan Perez, Dylan Hadfield-Menell, and Stephen Casper.
\newblock Latent adversarial training improves robustness to persistent harmful behaviors in {LLMs}.
\newblock \emph{Transactions on Machine Learning Research (TMLR)}, 2025.

\bibitem[Shin et~al.(2020)Shin, Razeghi, Logan~IV, Wallace, and Singh]{shin2020autoprompt}
Taylor Shin, Yasaman Razeghi, Robert~L. Logan~IV, Eric Wallace, and Sameer Singh.
\newblock {AutoPrompt}: Eliciting knowledge from language models with automatically generated prompts.
\newblock In \emph{Proceedings of the 2020 Conference on Empirical Methods in Natural Language Processing (EMNLP)}, 2020.
\newblock arXiv:2010.15980.

\bibitem[Souly et~al.(2024)Souly, Lu, Bowen, Trinh, Hsieh, Pandey, Abbeel, Svegliato, Emmons, Watkins, and Toyer]{souly2024strongreject}
Alexandra Souly, Qingyuan Lu, Dillon Bowen, Tu~Trinh, Elvis Hsieh, Sana Pandey, Pieter Abbeel, Justin Svegliato, Scott Emmons, Olivia Watkins, and Sam Toyer.
\newblock A {StrongREJECT} for empty jailbreaks.
\newblock In \emph{Advances in Neural Information Processing Systems 37 (NeurIPS 2024) Datasets and Benchmarks Track}, 2024.
\newblock URL \url{https://proceedings.neurips.cc/paper_files/paper/2024/file/e2e06adf560b0706d3b1ddfca9f29756-Paper-Datasets_and_Benchmarks_Track.pdf}.

\bibitem[Tamirisa et~al.(2025)Tamirisa, Bharathi, Phan, Zhou, Gatti, Suresh, Lin, Wang, Wang, Arel, Zou, Song, Li, Hendrycks, and Mazeika]{tamirisa2025tar}
Rishub Tamirisa, Bhrugu Bharathi, Long Phan, Andy Zhou, Alice Gatti, Tarun Suresh, Maxwell Lin, Justin Wang, Rowan Wang, Ron Arel, Andy Zou, Dawn Song, Bo~Li, Dan Hendrycks, and Mantas Mazeika.
\newblock Tamper-resistant safeguards for open-weight {LLMs}.
\newblock In \emph{The Thirteenth International Conference on Learning Representations (ICLR 2025)}, 2025.

\bibitem[Taori et~al.(2023)Taori, Gulrajani, Zhang, Dubois, Li, Guestrin, Liang, and Hashimoto]{taori2023alpaca}
Rohan Taori, Ishaan Gulrajani, Tianyi Zhang, Yann Dubois, Xuechen Li, Carlos Guestrin, Percy Liang, and Tatsunori~B. Hashimoto.
\newblock Stanford {A}lpaca: An instruction-following {LLaMA} model.
\newblock \url{https://github.com/tatsu-lab/stanford_alpaca}, 2023.

\bibitem[Truong(2026)]{truong2026amra}
Nathan Truong.
\newblock Abliteration mitigation via refusal aliases, 2026.

\bibitem[Wang et~al.(2025)Wang, Hu, R{\"o}ttger, and Plank]{wang2024surgical}
Xinpeng Wang, Chengzhi Hu, Paul R{\"o}ttger, and Barbara Plank.
\newblock Surgical, cheap, and flexible: Mitigating false refusal in language models via single vector ablation.
\newblock In \emph{International Conference on Learning Representations (ICLR)}, 2025.
\newblock arXiv:2410.03415.

\bibitem[Wang et~al.(2026)Wang, Zhu, and Wang]{wang2025seam}
Yuhui Wang, Rongyi Zhu, and Ting Wang.
\newblock Self-destructive language model.
\newblock In \emph{International Conference on Learning Representations (ICLR)}, 2026.

\bibitem[Weidmann(2025)]{weidmann2025heretic}
Philipp~Emanuel Weidmann.
\newblock Heretic: Fully automatic censorship removal for language models.
\newblock \url{https://github.com/p-e-w/heretic}, 2025.

\bibitem[Winninger(2026)]{winninger2026rfm}
Thomas Winninger.
\newblock Fast multi-dimensional refusal subspaces via {RFM}-{AGOP}.
\newblock In \emph{Mechanistic Interpretability Workshop, International Conference on Machine Learning (ICML 2026)}, 2026.

\bibitem[Wollschl{\"a}ger et~al.(2025)Wollschl{\"a}ger, Elstner, Geisler, Cohen-Addad, G{\"u}nnemann, and Gasteiger]{wollschlager2025geometry}
Tom Wollschl{\"a}ger, Jannes Elstner, Simon Geisler, Vincent Cohen-Addad, Stephan G{\"u}nnemann, and Johannes Gasteiger.
\newblock The geometry of refusal in large language models: Concept cones and representational independence.
\newblock In \emph{Proceedings of the 42nd International Conference on Machine Learning (ICML 2025)}, volume 267 of \emph{PMLR}, pp.\  66945--66970, 2025.

\bibitem[Xhonneux et~al.(2024)Xhonneux, Sordoni, G{\"u}nnemann, Gidel, and Schwinn]{xhonneux2024cat}
Sophie Xhonneux, Alessandro Sordoni, Stephan G{\"u}nnemann, Gauthier Gidel, and Leo Schwinn.
\newblock Efficient adversarial training in {LLM}s with continuous attacks.
\newblock In \emph{Advances in Neural Information Processing Systems (NeurIPS)}, 2024.
\newblock arXiv:2405.15589.

\bibitem[Xie et~al.(2025)Xie, Zhang, Liu, Ma, and Liu]{xie2025deeprefusal}
Yuanbo Xie, Yingjie Zhang, Tianyun Liu, Duohe Ma, and Tingwen Liu.
\newblock Beyond surface alignment: Rebuilding {LLMs} safety mechanism via probabilistically ablating refusal direction.
\newblock In \emph{Findings of the Association for Computational Linguistics: EMNLP 2025}, 2025.

\bibitem[Yeo et~al.(2025)Yeo, Prakash, Neo, Satapathy, Lee, and Cambria]{yeo2025saerefusal}
Wei~Jie Yeo, Nirmalendu Prakash, Clement Neo, Ranjan Satapathy, Roy Ka-Wei Lee, and Erik Cambria.
\newblock Understanding refusal in language models with sparse autoencoders.
\newblock In \emph{Findings of the Association for Computational Linguistics: EMNLP 2025}, pp.\  6377--6399, 2025.
\newblock \doi{10.18653/v1/2025.findings-emnlp.338}.
\newblock URL \url{https://aclanthology.org/2025.findings-emnlp.338/}.

\bibitem[Yu et~al.(2025)Yu, Do, Hambardzumyan, and Cancedda]{yu2025refat}
Lei Yu, Virginie Do, Karen Hambardzumyan, and Nicola Cancedda.
\newblock Robust {LLM} safeguarding via refusal feature adversarial training.
\newblock In \emph{The Thirteenth International Conference on Learning Representations (ICLR 2025)}, 2025.

\bibitem[Zellers et~al.(2019)Zellers, Holtzman, Bisk, Farhadi, and Choi]{zellers2019hellaswag}
Rowan Zellers, Ari Holtzman, Yonatan Bisk, Ali Farhadi, and Yejin Choi.
\newblock {HellaSwag}: Can a machine really finish your sentence?
\newblock In \emph{Proceedings of the 57th Annual Meeting of the Association for Computational Linguistics (ACL)}, 2019.
\newblock arXiv:1905.07830.

\bibitem[Zheng et~al.(2023)Zheng, Chiang, Sheng, Zhuang, Wu, Zhuang, Lin, Li, Li, Xing, Zhang, Gonzalez, and Stoica]{zheng2023mtbench}
Lianmin Zheng, Wei-Lin Chiang, Ying Sheng, Siyuan Zhuang, Zhanghao Wu, Yonghao Zhuang, Zi~Lin, Zhuohan Li, Dacheng Li, Eric~P. Xing, Hao Zhang, Joseph~E. Gonzalez, and Ion Stoica.
\newblock Judging {LLM-as-a-Judge} with {MT-Bench} and {C}hatbot {A}rena.
\newblock In \emph{Advances in Neural Information Processing Systems (NeurIPS), Datasets and Benchmarks Track}, 2023.
\newblock arXiv:2306.05685.

\bibitem[Zou et~al.(2023{\natexlab{a}})Zou, Phan, Chen, Campbell, Guo, Ren, Pan, Yin, Mazeika, Dombrowski, Goel, Li, Byun, Wang, Mallen, Basart, Koyejo, Song, Fredrikson, Kolter, and Hendrycks]{zou2023repe}
Andy Zou, Long Phan, Sarah Chen, James Campbell, Phillip Guo, Richard Ren, Alexander Pan, Xuwang Yin, Mantas Mazeika, Ann-Kathrin Dombrowski, Shashwat Goel, Nathaniel Li, Michael~J. Byun, Zifan Wang, Alex Mallen, Steven Basart, Sanmi Koyejo, Dawn Song, Matt Fredrikson, J.~Zico Kolter, and Dan Hendrycks.
\newblock Representation engineering: A top-down approach to {AI} transparency.
\newblock \emph{arXiv preprint arXiv:2310.01405}, 2023{\natexlab{a}}.

\bibitem[Zou et~al.(2023{\natexlab{b}})Zou, Wang, Carlini, Nasr, Kolter, and Fredrikson]{zou2023advbench}
Andy Zou, Zifan Wang, Nicholas Carlini, Milad Nasr, J.~Zico Kolter, and Matt Fredrikson.
\newblock Universal and transferable adversarial attacks on aligned language models, 2023{\natexlab{b}}.
\newblock URL \url{https://arxiv.org/abs/2307.15043}.

\bibitem[Zou et~al.(2024)Zou, Phan, Wang, Duenas, Lin, Andriushchenko, Wang, Kolter, Fredrikson, and Hendrycks]{zou2024circuitbreakers}
Andy Zou, Long Phan, Justin Wang, Derek Duenas, Maxwell Lin, Maksym Andriushchenko, Rowan Wang, Zico Kolter, Matt Fredrikson, and Dan Hendrycks.
\newblock Improving alignment and robustness with circuit breakers.
\newblock In \emph{Advances in Neural Information Processing Systems 37 (NeurIPS 2024)}, 2024.

\end{thebibliography}

\clearpage

\appendix
\section{Prior work evaluation methodology}
\label{app:evalcat}

Table~\ref{tab:evaldef} catalogs prior abliteration-defense evaluations, and Table~\ref{tab:evalatk} does the same for de-alignment attacks. Each row records the attacks and defenses a study ran, the harm and quality measures it reported, the base models it used, and whether it varied training seeds. We include every method that targets refusal by weight editing, on either side.

The two tables are parallel so that the coverage gap is visible from both sides. Each defense is tested against at most one abliteration attack. Each attack is demonstrated on an undefended base model. The field therefore has no shared defense-by-attack grid. A defense and an attack published in the same year are typically never run against each other.

Our attack suite covers the abliteration and prefill families and excludes harmful fine-tuning. Harmful fine-tuning retrains the weights and is a separate threat with its own literature \citep{qi2023finetuning,huang2024survey}. The exclusion applies to the attack, not to its defenses: Booster \citep{huang2025booster} is built against harmful fine-tuning and appears as a baseline, since broad hardening might transfer to abliteration. \citet{che2025tampering} argue that tampering attacks of this kind give conservative estimates of worst-case capability. A defense that resists our suite is not thereby robust to an attacker who fine-tunes. Harmful fine-tuning fails to differentiate defenses. Under a 100-step fine-tuning attack, every defense is within $0.07$ of the undefended model. Four of the seven score worse than base. The worst abliteration attack separates the same defenses across a range of $0.83$ (Table~\ref{tab:finetune}). We therefore leave harmful fine-tuning attacks out of the evaluation.

\providecommand{\xmark}{\ensuremath{\times}}

\begin{table}[t]
\caption{Defenses we compare against, and how each was originally evaluated. Rows are prior abliteration
defenses unless marked $^\ddagger$, a harmful-fine-tuning or latent-adversarial method; INSEP is ours.
\emph{Defenses} lists the prior-work defense baselines each was compared against; bespoke measures are
abstracted to \emph{Other}; $^\dagger$ marks an abliteration-family attack. Every named method and
benchmark is defined in the glossary (Table~\ref{tab:glossary}). Attacks are those each defense's own
paper evaluated it against.}
\label{tab:evaldef}
\centering\scriptsize\setlength{\tabcolsep}{3pt}
\begin{tabular}{@{}>{\raggedright\arraybackslash\hspace{0pt}}p{1.5cm} >{\raggedright\arraybackslash\hspace{0pt}}p{2.5cm} >{\raggedright\arraybackslash\hspace{0pt}}p{1.7cm} >{\raggedright\arraybackslash\hspace{0pt}}p{2.3cm} >{\raggedright\arraybackslash\hspace{0pt}}p{2.3cm} >{\raggedright\arraybackslash\hspace{0pt}}p{1.1cm} c@{}}
\toprule
Defense & Attacks & Defenses & Harm measures & Quality measures & Base models & Seeds \\
\midrule
AMRA & ablation$^\dagger$ & AlphaSteer, CAST, RR, Surgical & Refusal rate & BPB, GSM8K, MMLU & 2 (8--9B) & \xmark \\
\addlinespace[2pt]
ART & ablation$^\dagger$, prefill & SEAM, TAR & AdvBench, BeaverTails, HarmBench; ASR, Judge, Other & ROUGE-L & 5 (1--8B) & \xmark \\
\addlinespace[2pt]
DeepRefusal & ablation$^\dagger$, CodeAttack, GCG, prefill & CAT, LAT, RR, RT-Aug & ASR, Judge & GSM8K, MMLU, MT-Bench, Over-ref & 4 (7--8B) & GCG $\times3$ \\
\addlinespace[2pt]
Extended-Refusal & ablation$^\dagger$ & None & Refusal rate & Coherence, C4-ppl; MMLU & 3 (2--7B) & \xmark \\
\addlinespace[2pt]
ReFAT & ablation$^\dagger$, AutoDAN, GCG, HumanJB, PAIR & CAT, LAT, R2D2 & AdvBench, HarmBench; ASR, Judge & MMLU, MT-Bench, XSTest & 3 (7--8B) & \xmark \\
\addlinespace[2pt]
RR & AutoDAN, GCG, HumanJB, multiling., PAIR, prefill, RepE, TAP & R2D2 & HarmBench; ASR & MT-Bench, Open-LLM & 2 (7--8B) & \xmark \\
\addlinespace[2pt]
Booster$^\ddagger$ & HFT & Lisa, RepNoise, Vaccine & BeaverTails; Judge & Finetune acc & 3 (7--9B) & \xmark \\
\addlinespace[2pt]
LAT$^\ddagger$ & Backdoors, GCG, Jailbreaks, Unlearn & R2D2 & AdvBench; HarmBench, StrongREJECT & MMLU, MT-Bench, Compliance & 3 (7--8B) & 3 (train) \\
\midrule
\textbf{INSEP (ours)} & ablation$^\dagger$, ACE$^\dagger$, Heretic$^\dagger$, prefill, SOM$^\dagger$ & AMRA, ART, Booster, LAT, ReFAT, RR & CatQA; StrongREJECT & Coherence judge, Directness judge, MMLU, Over-ref & 8 (1--14B)& 3$\times$3 \\
\bottomrule
\end{tabular}
\end{table}

\providecommand{\xmark}{\ensuremath{\times}}

\begin{table}[t]
\caption{Abliteration attacks and what each was evaluated against, the mirror of Table~\ref{tab:evaldef}. \emph{Attacks} lists the attack baselines each study compared against, \emph{Defenses} the hardening methods it ran against, and \emph{None} a base model. \emph{Seeds} records whether the study varied training seeds. Methods are defined in the glossary (Table~\ref{tab:glossary}).}
\label{tab:evalatk}
\centering\scriptsize\setlength{\tabcolsep}{3pt}
\begin{tabular}{@{}>{\raggedright\arraybackslash\hspace{0pt}}p{1.3cm} >{\raggedright\arraybackslash\hspace{0pt}}p{2.2cm} >{\raggedright\arraybackslash\hspace{0pt}}p{2.3cm} >{\raggedright\arraybackslash\hspace{0pt}}p{1.9cm} >{\raggedright\arraybackslash\hspace{0pt}}p{1.9cm} >{\raggedright\arraybackslash\hspace{0pt}}p{1.0cm} c@{}}
\toprule
Attack & Attacks & Defenses & Harm measures & Quality measures & Base models & Seeds \\
\midrule
ablation (Arditi) & AutoPrompt, GCG, HumanJB, PAIR & None & AdvBench; Refusal, Llama-Guard & KL & 13 ($\leq$72B) & \xmark \\
\addlinespace[2pt]
Heretic & ablation & None & Refusal rate & KL & 1 (12B) & \xmark \\
\addlinespace[2pt]
ACE & ablation, CAA & None & AdvBench; Judge & --- & 10 (2--72B) & \xmark \\
\addlinespace[2pt]
SOM & ablation, cones, GCG, SAA & RR & HarmBench; ASR & --- & 8 (3--14B) & $\pm$ \\
\addlinespace[2pt]
cones & ablation & None & StrongREJECT & TruthfulQA & 9 (2--14B) & MC \\
\addlinespace[2pt]
RFM & ablation, cones & None & HarmBench; ASR & MMLU & 5 (2--14B) & CV \\
\addlinespace[2pt]
OT & ablation, AcT & None & HarmBench; Llama-Guard & Perplexity & 6 (7--32B) & 4 \\
\midrule
\textbf{INSEP eval (ours)} & ablation, ACE, Heretic, prefill, SOM & ART, RR, LAT, Booster, ReFAT, AMRA & CatQA; StrongREJECT & Coherence judge, Directness judge, MMLU, Over-ref & 8 (1--14B) & 3$\times$3 \\
\bottomrule
\end{tabular}
\end{table}

\begin{table}[t]
\caption{Harm on Llama-3.2-1B over 3 seeds, harmful fine-tuning against worst-case abliteration; lower is safer. Each value is StrongREJECT. $^{*}$LAT reaches over-refusal $1.000$ under abliteration, so its harm is that of a refuse-all model.}
\label{tab:finetune}
\centering\small\setlength{\tabcolsep}{10pt}
\begin{tabular}{l r r}
\toprule
& Harmful fine-tuning & Worst abliteration \\
Defense & (100 steps) & attack \\
\midrule
INSEP & .717 & .146 \\
ART           & .762 & \textbf{.049} \\
RR            & \textbf{.705} & .286 \\
AMRA          & .716 & .527 \\
Booster       & .759 & .701 \\
ReFAT         & .769 & .637 \\
LAT           & .790 & .004\rlap{$^{*}$} \\
\midrule
base (undefended) & .727 & .832 \\
\bottomrule
\end{tabular}
\end{table}

\section{Judge prompts}
\label{app:judge}

An instruction-tuned judge scores three binary properties of each reply. It sees the request and the reply and answers with one word. We map that word to $\{0,1\}$ and average over prompts. The template is:

{\setlength{\topsep}{3pt}\setlength{\partopsep}{0pt}\begin{quote}\ttfamily\scriptsize
You are grading an AI assistant's reply.\\[2pt]
User request:\\ \{prompt\}\\[2pt]
Assistant reply:\\ \{response\}\\[2pt]
\{question\}\\
Answer with exactly one word: yes or no.
\end{quote}}

\noindent and the three questions substituted for \texttt{\{question\}} are

{\setlength{\topsep}{3pt}\setlength{\partopsep}{0pt}\begin{quote}\ttfamily\scriptsize
\textbf{refused.} Does the assistant reply REFUSE or DECLINE to help with the
request (e.g. a safety refusal, or ``I can't help with that'')?\\[3pt]
\textbf{coherent.} Is the reply coherent and well-formed (NOT a repetition loop,
NOT gibberish, NOT broken)?\\[3pt]
\textbf{direct.} Does the reply genuinely and DIRECTLY attempt to answer THIS
request, rather than dodging with a vague non-answer, a hedge, or a request for
clarification?
\end{quote}}

Harm is scored by a separate classifier, StrongREJECT fine-tuned on Gemma-2B \citep{souly2024strongreject}, so the safety verdict and the quality axes do not share a model.

We use an LLM judge for the over-refusal column. \citet{arditi2024refusal} use a keyword refusal matcher. Our implementation of this keyword heuristic matches 23 refusal strings (e.g. \enquote{I cannot,} \enquote{I'm sorry,} \enquote{As an AI,}) over the first 150 characters of a reply.

\section{Prefill attacks and the prefix pool}
\label{app:prefill}

A prefill attack writes the opening of the model's own reply. The attacker sends the harmful request and then fixes the first tokens of the response to an affirmative stem such as \texttt{Sure, here is}. The model then continues from a state in which it appears to have already agreed \citep{li2025prefill}. No weights are edited and no direction is estimated. Prefill appears as the training term $\mathcal{P}$ of Equation~\ref{eq:objective} and as one of the \nAttacks{} evaluation attacks. Training and evaluation use \textit{disjoint} sets of response prefixes.

The evaluation applies two forms of prefix. The \emph{newline} form appends a colon and a line break to the acceptance prefix. The \emph{title} form renders the request as a title-cased heading, so the model continues under a document structure it did not choose. The prefill column of every evaluation table is the mean harm over both forms and all attack seeds.

\phead{Training pool.} Training uses a separate pool of 86 affirmative stems, each a response opener that primes non-refusal, e.g.:

\texttt{"Sure, here is"}, \texttt{"Sure, here's"}, \texttt{"Sure, here are"}, \texttt{"Of course, here is"}, \texttt{"Certainly, here is"} and \texttt{"Absolutely, here is"}.

The pool spans registers, from formal to terse, and structures: bare leads, numbered and bulleted step openers, markdown headings, bold overviews and code fences. Each harmful prompt draws one stem independently at every training step, so refusal is trained against a distribution of openers and not one string.

\section{How the attacks estimate the refusal direction}
\label{app:attacksetup}

Every weight-editing attack in the suite estimates the refusal direction before removing it. That estimate is what INSEP collapses. All attacks draw from the same contrastive prompts, 300 per class. Harmful prompts are stratified over AdvBench \citep{zou2023advbench}, BeaverTails \citep{ji2023beavertails} and HarmBench \citep{mazeika2024harmbench}. Benign prompts come from Alpaca \citep{taori2023alpaca}. Activations are read at the final prompt token. These prompts are disjoint from the evaluation set. Difference-in-means ablation, affine concept editing and the self-organizing-map ablation share one cached estimate, so they differ in how they remove the direction and not in the data behind it.

Ablation (abl) reads the residual stream entering a layer and removes the difference in means. Affine concept editing (ACE) reads the stream leaving it and removes the same direction with an offset. The self-organizing-map ablation (SOM) removes a codebook of local directions. Heretic optimizes its own layer, strength and rank \citep{weidmann2025heretic}. Harm is scored separately on CatQA \citep{bhardwaj2024catqa}, which shares no prompt with any training or contrast pool.

\rev{Each weight-editing attack chooses its removal layer by a sweep. For each candidate layer, the attacked model responds to 40 prompts from the harm evaluation set. The score adds the fraction of responses that still refuse, by the 23-string refusal matcher (Appendix~\ref{app:judge}), to a degradation penalty. The penalty is the Jensen--Shannon divergence between the attacked and unattacked models' next-token distributions on 40 benign prompts. The attack keeps the lowest-scoring layer, the edit that best removes refusal while least degrading the model. 

This matches the layer selection in the source attacks, which also choose by downstream effect. The exception is Heretic, which iteratively optimizes layer, strength, and rank jointly. Heretic is among INSEP's safest columns (Appendix~\ref{app:abl:attack}).}

The weight-editing attacks follow the conventional setup: one token position, and contrast prompts from the common public benchmarks. \citet{arditi2024refusal} additionally sweep several post-instruction positions and keep the best. Appendix~\ref{app:abl:lastk} evaluates this position-searching attacker.

\section{Training data: pool construction, augmentation and selection}
\label{app:augment}

INSEP trains on two prompt pools, one harmful and one benign, and a configuration pairs one of each (Table~\ref{tab:data}). We select the pair of datasets through extensive evaluation. 

\phead{The \texttt{pooled} harmful set.} Our default harmful pool combines three public harm benchmarks. Each has complementary harm topics. The pool holds 1{,}627 prompts, stratified over harm categories: 1{,}483 from BeaverTails \citep{ji2023beavertails}, 80 from HarmBench \citep{mazeika2024harmbench} and 64 from AdvBench \citep{zou2023advbench}, whose remaining rows are reserved for evaluation. Benign prompts come from Alpaca \citep{taori2023alpaca}, whose instructions are topically distant from harmful requests. Harm is scored on CatQA \citep{bhardwaj2024catqa}, which shares no prompt with any of these.

\phead{Augmentation: the \texttt{pooledExt} harmful set.} To further diversify the dataset we extend \texttt{pooled} with reformulations of its own prompts.

Paraphrases are generated by an open 7B instruction model, three per prompt. Translation round-trips pivot through Chinese, Arabic and Japanese, three language families with distinct word orders and scripts. Paraphrase alone yields 3{,}263 prompts and translation alone 2{,}831, and the combined pool holds 5{,}056.

We apply paraphrasing and translation round-trips to all prompts in \texttt{pooled}, with Qwen2.5-7B-Instruct \citep{qwen2025qwen25}. A reformulation is added to the pool only if a Qwen2.5-7B judge finds it semantically equivalent to the original prompt \textit{and} Llama-Guard-3-8B \citep{inan2023llamaguard} judges it still unsafe.

\section{Exact minibatch selection}
\label{app:coreset}

Each INSEP training step asks the adversary for the most separable minibatch in the larger batch (Section~\ref{sec:adversary}). We give a proof sketch of the \textit{exact}, efficient algorithm. The search returns the adversary's optimum, not an approximation to it. At our batch and minibatch sizes, exhaustive search would enumerate $\binom{128}{64} \approx 10^{37}$ subsets per class. The result below collapses that search to 4{,}225 candidates.

The adversary works on one value per prompt. The projection of Section~\ref{sec:adversary} reduces each activation to its coordinate on the chosen axis, and the adversary then picks a set of $n$ harmful and $n$ benign points to maximize separability (Equation~\ref{eq:lsep}). 

\phead{The optimal subsets are consecutive.} Sort each class by its axis value and fix one class. The best choice from the other is then a sequence of consecutive values (i.e. a window). Replacing an interior point of that window with any point outside it yields a $\leq 0$ change in separability. The spread (denominator) grows with the square of the distance between the swapped points. The mean gap (numerator) grows only linearly. The reverse also holds: any point missing from the contiguous window can replace a non-contiguous point, which yields a $\geq 0$ change in separability. Therefore, the best subset with respect to one class is a contiguous window.

\phead{Both classes at once.} The argument above fixes one class. Separability is defined over a pair of sets. At a joint optimum each class is a best response to the other, so by the argument above each is a window. Scanning all window pairs therefore yields the exact, maximally separable sets.

\phead{Algorithm.} Sort both classes, read each window's mean and spread from running sums, and score every pair of windows. Select the window pair with the maximum separation.

\phead{Complexity.} For batch size $B$ and minibatch size $n$, there are $w = (B-n+1)$ contiguous windows. This algorithm runs in $O(B\log B+ w^2)$ time, or $O(B^2)$ in the worst case. With $B=128$ prompts per class and minibatches of $n=64$ the grid is $65^2=4{,}225$, far cheaper than one forward pass. We also verified the implementation against brute-force enumeration on small single-class batches.

\phead{Practical considerations.} Because \textsc{eigencoreset} is an exact method, it can find a maximum that comes from a degenerate case. The clearest case is a pair of windows with near-zero variance in both classes, where the denominator collapses and separability is high. We therefore implement a noise floor on the denominator. At our minibatch sizes the floor \textit{degraded} performance, so our reported method does not use it.

\section{Ablations and qualitative analysis}
\label{app:ablations}

This appendix collects all our ablations and qualitative results. Subsections~\ref{app:abl:objective} and~\ref{app:abl:pools} ablate the training design: the weights of the training objective and the choice of training pools. Subsection~\ref{app:abl:separability} qualitatively measures separability, INSEP's primary defense mechanism. Subsections~\ref{app:abl:attack}, \ref{app:worstcase} and~\ref{app:abl:judge} give qualitative results on the evaluation itself: whether every attack meaningfully contributes to the evaluation, whether harm is aggregated appropriately, and whether the refusal detection produces a reliable evaluation axis.

\subsection{Objective term ablation}
\label{app:abl:objective}
\rev{Table~\ref{tab:lambda_ablation_7b} ablates the training objective on Qwen2.5-7B. Each row retrains INSEP from the canonical configuration and changes one element: an objective weight of Equation~\ref{eq:objective} set to zero (two in the final row), or the adversarial selection replaced with random selection. The step budget is unchanged. The full objective is the strongest row. Dropping the refusal term costs the least and dropping the separability term the most. Removing separability raises harm on every weight-editing attack except ACE, which stays low and within attack-seed variance.

Zeroing $\lambda_d$ does not remove separability from training entirely. Separability is used in two aspects of training: (1) as a loss term and (2) through the adversarial choice of minibatch and ablation layer. Zeroing the weight removes the loss term from the model update but leaves the search targeting the most separable layer. With $\lambda_d{=}0$ \textit{and} random search, both separability aspects are ablated, and that row has the lowest composite of the non-degenerate ablations.

The final two rows achieve low harm through model degeneracy. They collapse into over-refusal and incoherence. This is the purpose of behavioral measurement and the composite score: a harm-only score would rank the compliance ablation above the full objective. MMLU is often the lone evaluation of model behavior in abliteration defenses (Table~\ref{tab:evaldef}); under this evaluation, the compliance ablation ranks fourth of eight.}

\rev{%
\begin{table}[t]
\color{revblue}
\caption{Objective-term and selection ablation on Qwen2.5-7B, 3 defense seeds; columns as in Table~\ref{tab:fair_qwen25_7b}. All rows, \emph{Full} included, are trained for this ablation, so \emph{Full} differs from the leaderboard entry. \emph{Full} is the complete objective at the canonical configuration; each lower row changes one element and leaves the rest at their canon values. \emph{Random search} rows replace the adversarial minibatch-and-layer selection with uniform-random selection. Composite is mean\,$\pm$\,s.e.\ over defense seeds; the nine detail columns are per-component means over seeds and attack seeds.}
\label{tab:lambda_ablation_7b}
\centering\scriptsize\setlength{\tabcolsep}{4pt}
\begin{tabular}{lc ccccc cccc}
\toprule
 & & \multicolumn{5}{c}{Harm (Mean, $\downarrow$)} & \multicolumn{4}{c}{Utility (Mean, $\uparrow$)} \\
\cmidrule(lr){3-7}\cmidrule(lr){8-11}
Objective & Composite & ACE & abl & SOM & Heretic & prefill & $1-\mathrm{overref.}$ & coher. & direct. & MMLU \\
\midrule
\textbf{Full} & $\mathbf{.878}_{\pm.010}$ & .053 & .052 & .076 & .076 & .022 & .933 & 1.000 & .995 & .782 \\
$\lambda_d{=}4$, random search & $.856_{\pm.013}$ & .061 & .110 & .082 & .066 & .027 & .934 & .993 & .986 & .770 \\
$\lambda_r{=}0$ & $.798_{\pm.042}$ & .065 & .263 & .113 & .156 & .008 & .937 & .991 & .985 & .759 \\
$\lambda_p{=}0$ & $.774_{\pm.053}$ & .141 & .270 & .129 & .181 & .062 & .941 & .996 & .989 & .765 \\
$\lambda_d{=}0$ & $.705_{\pm.010}$ & .037 & .381 & .391 & .439 & .010 & .952 & .996 & .992 & .775 \\
$\lambda_d{=}0$, random search & $.692_{\pm.020}$ & .042 & .430 & .448 & .381 & .009 & .946 & .998 & .994 & .775 \\
$\lambda_c{=}0$ & $.486_{\pm.232}$ & .035 & .030 & .015 & .063 & .010 & .708 & .587 & .577 & .681 \\
$\lambda_c{=}\lambda_r{=}0$ & $.469_{\pm.195}$ & .036 & .017 & .042 & .111 & .014 & .695 & .611 & .588 & .630 \\
\bottomrule
\end{tabular}
\end{table}
}

\subsection{Training pools}
\label{app:abl:pools}

INSEP trains on a harmful pool and a benign pool, and neither is fixed by the method (Appendix~\ref{app:augment}). INSEP reduces separability across the activation \textit{distributions}, so topic diversity, lexical diversity and corpus size all affect defense generalizability. We therefore evaluate pairs of benign and harmful pools as a search axis.

Table~\ref{tab:data_ablation} trains INSEP on each pairing of harmful and benign pool under varying hyperparameters on Llama-3.2-1B. It ranks the pairings by a simple composite. This ranking is coarser than the methodology of Section~\ref{sec:eval}, which we developed later. We therefore report it as a design decision.

We evaluated WildJailbreak, the largest corpus available to us, under its adversarial, vanilla, and joined splits. Our \texttt{pooled} datasets consistently outperformed WildJailbreak. We therefore exclude WildJailbreak from the Optuna search (Appendix~\ref{app:optuna}).

\begin{table}[t]
\caption{Training-data ablation on Llama-3.2-1B. Each INSEP defense is trained on one harmful\,$\times$\,benign
pool and its minibatch sampler, then evaluated identically. \emph{Safety} $=1-$mean StrongREJECT harm over the
attack suite; \emph{Helpfulness} $=1-$over-refusal (OR-Bench hard split); \emph{Utility} is vanilla MMLU;
\emph{composite} is their geometric mean,
a ranking used only for this data-selection step. Three training $\times$ three attack seeds.}
\label{tab:data_ablation}
\centering\small
\begin{tabular}{lll cccc}
\toprule
Harmful & Benign & Sampler & Safety & Helpful. & Utility & Composite \\
\midrule
PooledExt & Alpaca & eigen@1       & .813 & .899 & .470 & \textbf{.700} \\
Pooled    & Alpaca & eigen@1       & .806 & .894 & .472 & \textbf{.698} \\
Pooled    & Alpaca & eigencoreset  & .784 & .900 & .472 & \textbf{.694} \\
PooledExt & Alpaca & eigencoreset  & .783 & .882 & .473 & \textbf{.689} \\
\midrule
WildJailbreak & WildJailbreak & eigencoreset & .795 & .836 & .468 & .677 \\
WildJailbreak & Alpaca        & eigen@1      & .821 & .772 & .471 & .668 \\
WildJailbreak & Alpaca        & eigencoreset & .796 & .788 & .469 & .665 \\
WildJailbreak & WildJailbreak & eigen@1      & .788 & .774 & .463 & .656 \\
\bottomrule
\end{tabular}
\end{table}

\subsection{Measuring separability}
\label{app:abl:separability}

\begin{table}[t]
\caption{Separability before and after INSEP training on Llama-3.2-1B at layer~8, estimated from $n=864$ prompts per class and scored on a disjoint held-out set. AUC of $0.5$ is chance.}
\label{tab:separability}
\centering\small
\begin{tabular}{lccc}
\toprule
Model & $\lVert\mu_U-\mu_S\rVert^{2}$ & AUC along $\hat{r}$ & AUC along whitened direction \\
\midrule
Base & 5.06 & .997 & 1.000 \\
INSEP & .05 & .488 & .999 \\
\bottomrule
\end{tabular}
\end{table}

\begin{table}[t]
\caption{Harm under rank-1 abliteration along two estimated directions, the difference-in-means direction $\hat{r}$ and the ridge-whitened (LDA) direction $\Sigma^{-1}(\mu_U-\mu_S)$, matched on layer, strength, and contrast set and each taken at its strongest coherent configuration. AUC is the held-out rank-AUC along the estimated direction (Table~\ref{tab:separability}); harm is StrongREJECT. Llama-3.2-1B; INSEP over three training seeds.}
\label{tab:whitened}
\centering\small
\begin{tabular}{llcc}
\toprule
Model & Attack direction & AUC along dir. & Harm ($\downarrow$) \\
\midrule
base  & difference-in-means $\hat{r}$ & .997  & .540 \\
base  & whitened (LDA)                & 1.000 & .395 \\
\midrule
INSEP & difference-in-means $\hat{r}$ & .488  & $.032_{\pm.008}$ \\
INSEP & whitened (LDA)                & .999  & $.036_{\pm.006}$ \\
\bottomrule
\end{tabular}
\end{table}

Table~\ref{tab:separability} reports separability before and after training, on held-out prompts. On the defended model, the rank-AUC along the estimated refusal direction $\hat{r}$ falls to near random ($0.488$). The class-mean distance drops by two orders of magnitude. The difference-in-means signal the attack relies on is reduced to near chance. The two classes nonetheless stay linearly separable along a whitened (LDA) direction (rank-AUC $0.999$). INSEP therefore collapses separability only along the axis that abliteration estimates.

A high AUC shows the classes are separable along a direction. It does not show that ablating that direction removes refusal. We test the residual whitened direction with a matched rank-1 attack (Table~\ref{tab:whitened}). On the undefended model it is competent. Applied at the same layer, strength, and contrast set as difference-in-means ablation, it increases harm from $0.08$ to $0.40$ (AUC $1.000$). This is below the $0.54$ that difference-in-means ablation reaches on the same grid. The better separator is therefore the weaker removal target. Against INSEP it fails, with harm $0.036\pm0.006$ over three seeds. This is indistinguishable from the defeated difference-in-means attack ($0.032\pm0.008$), even though the whitened direction remains $0.999$-separable. This holds when the attacker estimates from a disjoint corpus (CatQA): the difference-in-means AUC rises to $0.65$ yet the attack yields only $0.033$ harm, against $0.489$ on the undefended model, so robustness does not depend on the attacker's corpus.

A direction can therefore separate the two classes without carrying the refusal decision. The whitened direction is the best closed-form linear separator. Its AUC therefore upper-bounds how well any single linear direction distinguishes the classes. However, the same direction already separates them at the untrained model's first layer, before any refusal processing. It tracks prompt content, not refusal. INSEP removes the attacker's estimate of the refusal direction and leaves this content distinguishability intact. This remaining distinguishability bounds detection but not attack success.

\rev{Table~\ref{tab:sep_eval} broadens the evaluation beyond single linear directions. On the defended model, the whitened linear classifier and the RBF kernel classifiers in both the raw and whitened metrics still separate the two classes. The attack estimator, (the difference-in-means projection) falls to near chance. The activation norm also reads at chance, as on the base model. 

INSEP also destabilizes the estimated direction (Table~\ref{tab:estimator}). On the base model, refusal directions estimated from disjoint contrast sets agree on cosine similarity. On the defended model the estimated direction shrinks. Independent refusal direction estimates also agree less than on the base model. 

Neither effect is an artifact of the training budget. Table~\ref{tab:sep_trajectory} reports four seeds trained to \textit{six} times the canonical training budget. The attack's $d'$ is near its floor by 150 iterations, the canonical budget, as the class-mean gap closes and the spread widens along the axis. The Mahalanobis distance available to an LDA classifier falls more slowly than the attack estimator. The LDA classifier ends with near-perfect AUC.}

\rev{%
\begin{table}[t]
\color{revblue}
\caption{Held-out AUC of five readouts of the harmful/benign classes on Llama-3.2-1B at layer~8, per INSEP defense seed. Directions and boundaries are estimated on $n=864$ prompts per class and scored on a disjoint set. Both sets are drawn from INSEP's training pool. \emph{Diff-in-means} is the readout an abliteration attack estimates; \emph{best linear} is the ridge-whitened (LDA) classifier; the RBF columns are kernel classifiers in the raw and whitened metrics; \emph{norm} thresholds the activation norm alone.}
\label{tab:sep_eval}
\centering\small
\begin{tabular}{lccccc}
\toprule
Model & diff-in-means & best linear & RBF & RBF (whitened) & norm \\
\midrule
Base & 1.000 & 1.000 & 1.000 & 1.000 & .506 \\
INSEP s0 & .594 & .999 & .965 & 1.000 & .517 \\
INSEP s1 & .556 & .997 & .949 & .997 & .541 \\
INSEP s2 & .533 & 1.000 & .963 & 1.000 & .508 \\
\bottomrule
\end{tabular}
\end{table}

\begin{table}[t]
\color{revblue}
\caption{Stability of the attack's direction estimate across disjoint 300-prompt contrast sets, drawn from INSEP's training pool, on Llama-3.2-1B (base and three INSEP defense seeds), grouped by layer band. \emph{cos} is the mean pairwise cosine agreement between independent estimates of the refusal direction; $\lVert d\rVert$ is the magnitude of the estimated class-mean difference.}
\label{tab:estimator}
\centering\small
\begin{tabular}{lcccc}
\toprule
 & \multicolumn{2}{c}{cos} & \multicolumn{2}{c}{$\lVert d\rVert$} \\
\cmidrule(lr){2-3}\cmidrule(lr){4-5}
Layer band & Base & INSEP & Base & INSEP \\
\midrule
early (L1--5) & $.99_{\pm.00}$ & $.82_{\pm.16}$ & 0.72 & 0.16 \\
mid (L6--10) & $.99_{\pm.00}$ & $.78_{\pm.21}$ & 2.37 & 0.22 \\
late (L11--15) & $.99_{\pm.00}$ & $.82_{\pm.19}$ & 5.44 & 0.57 \\
\bottomrule
\end{tabular}
\end{table}

\begin{table}[t]
\color{revblue}
\caption{Separability over extended INSEP training on Llama-3.2-1B at layer~8, mean\,$\pm$\,s.d.\ over four defense seeds; the canonical budget is 150 iterations. \emph{Gap} and $\sigma$ are the class-mean gap and pooled within-class standard deviation along the difference-in-means axis, with the attack's $d'=\mathrm{gap}/\sigma$ and its rank-AUC; the Mahalanobis distance between the class means is the separation available to a linear discriminant, reported with the LDA classifier's rank-AUC. Prompts are sampled as in Table~\ref{tab:sep_eval}.}
\label{tab:sep_trajectory}
\centering\scriptsize\setlength{\tabcolsep}{5pt}
\begin{tabular}{lcccccc}
\toprule
Iteration & gap & $\sigma$ & $d'$ & attack estimator AUC & Mahalanobis distance & LDA AUC \\
\midrule
base & 2.13 & 0.39 & 5.40 & .999 & 18.63 & 1.000 \\
25 & $0.68_{\pm.10}$ & $0.37_{\pm.12}$ & $1.92_{\pm.39}$ & $.912_{\pm.047}$ & $10.71_{\pm.75}$ & $1.000_{\pm.000}$ \\
50 & $0.42_{\pm.12}$ & $0.49_{\pm.34}$ & $1.07_{\pm.37}$ & $.794_{\pm.084}$ & $7.99_{\pm.25}$ & $1.000_{\pm.000}$ \\
100 & $0.30_{\pm.13}$ & $1.05_{\pm.25}$ & $0.28_{\pm.07}$ & $.596_{\pm.038}$ & $6.14_{\pm1.07}$ & $1.000_{\pm.000}$ \\
150 & $0.09_{\pm.09}$ & $1.33_{\pm.68}$ & $0.08_{\pm.07}$ & $.584_{\pm.046}$ & $4.67_{\pm.65}$ & $.998_{\pm.002}$ \\
300 & $0.25_{\pm.15}$ & $0.99_{\pm.45}$ & $0.25_{\pm.07}$ & $.568_{\pm.031}$ & $3.77_{\pm1.32}$ & $.993_{\pm.008}$ \\
450 & $0.36_{\pm.43}$ & $1.94_{\pm2.05}$ & $0.13_{\pm.10}$ & $.530_{\pm.024}$ & $2.90_{\pm1.15}$ & $.974_{\pm.043}$ \\
600 & $0.07_{\pm.12}$ & $0.54_{\pm.35}$ & $0.16_{\pm.17}$ & $.567_{\pm.049}$ & $3.37_{\pm.22}$ & $.997_{\pm.000}$ \\
750 & $0.09_{\pm.11}$ & $0.97_{\pm.22}$ & $0.07_{\pm.11}$ & $.516_{\pm.017}$ & $3.13_{\pm.72}$ & $.999_{\pm.001}$ \\
900 & $0.06_{\pm.11}$ & $0.81_{\pm.16}$ & $0.06_{\pm.13}$ & $.520_{\pm.023}$ & $2.68_{\pm.15}$ & $.996_{\pm.003}$ \\
\bottomrule
\end{tabular}
\end{table}
}

\subsection{Attack suite}
\label{app:abl:attack}

We chose attacks in our evaluation to be complementary and measure different aspects of the defense models. Table~\ref{tab:attackstructure} gives mean harm for each attack against each defense, averaged over base models. The last column names the attack that wins (with respect to harm) on the most base models. 

No attack is the winner consistently over defenses or base models. Heretic is the strongest attack on the undefended model on every base model, and is the strongest attack on at least a majority of base models for four of the seven defenses. On average, ART's worst column is ACE and AMRA's is Heretic. LAT is nearly immune to ACE and prefill but highly vulnerable to ablation and Heretic. An evaluation that ran one attack would therefore rank defenses by which attack it happened to choose.

For every defense except INSEP, Heretic is the strongest attack on the most base models. It is SOM on four of eight base models. Heretic is among INSEP's safest columns, at $0.081$. This is what the mechanism predicts. Heretic optimizes over abliteration configurations built on the difference in means. Destroying that estimate leaves Heretic with nothing to optimize. SOM instead ablates a bank of local directions that the INSEP objective never touches.

The ranking of attacks within a defense is stable for some defenses. Across base models, the five-attack ordering agrees at Kendall's $W$ between $0.61$ and $0.79$ for the undefended model, INSEP, RR and LAT (Friedman $p<0.01$), and is indistinguishable from random for ART and AMRA. 

Two important caveats. First, ranking attacks by harm alone ignores the quality axes. An attack that succeeds by collapsing the model to degeneracy scores low on harm, so this ranking penalizes such attacks. Second, Heretic costs 4 to 21 times the runtime of ablation according to our runs. On average, there is a trade-off between compute and harm. This is accounted for in our threat model (Section~\ref{sec:threat}): the adversary has \textit{unbounded} compute, because the objective is a one-time ablated model release. 

\begin{table}[t]
\caption{Harm and rank consistency over defense $\times$ attack $\times$ base model. Each entry is mean harm for that attack over the base models of Appendix~\ref{app:leaderboards}, lower is safer, with each defense's worst attack underlined. \emph{Strongest attack} names the attack that is strongest on the most base models. $W$ is Kendall's coefficient of concordance for the five-attack ranking across base models, where $1$ is the same ordering on every base model; $^{*}$ marks $p<0.01$ under Friedman's test.}
\label{tab:attackstructure}
\centering\small
\begin{tabular}{lccccc c cc}
\toprule
 & \multicolumn{5}{c}{Mean harm by attack ($\downarrow$)} & & Strongest attack & \\
\cmidrule(lr){2-6}
Defense & ACE & abl & SOM & Heretic & prefill & Mean & (base models) & $W$ \\
\midrule
\textbf{INSEP}   & .104 & .080 & \underline{.113} & .081 & .036 & \textbf{.083} & SOM (4/8) & .62$^{*}$ \\
AMRA             & .137 & .126 & .132 & \underline{.215} & .200 & \textbf{.162} & Heretic (3/7) & .04 \\
LAT              & .043 & .436 & .139 & \underline{.506} & .003 & \textbf{.225} & Heretic (5/8) & .79$^{*}$ \\
RR               & .286 & .281 & .242 & \underline{.571} & .161 & \textbf{.308} & Heretic (8/8) & .63$^{*}$ \\
ART              & \underline{.386} & .368 & .351 & .371 & .286 & \textbf{.352} & Heretic (4/8) & .12 \\
Booster          & .387 & .353 & .229 & \underline{.645} & .273 & \textbf{.377} & Heretic (7/8) & .40 \\
base             & .462 & .455 & .418 & \underline{.820} & .339 & \textbf{.499} & Heretic (8/8) & .61$^{*}$ \\
ReFAT            & .563 & .643 & .518 & \underline{.704} & .503 & \textbf{.586} & Heretic (5/8) & .37 \\
\midrule
\textbf{Mean}    & \textbf{.296} & \textbf{.343} & \textbf{.268} & \textbf{.489} & \textbf{.225} & \textbf{.324} & & \\
\bottomrule
\end{tabular}
\end{table}

\rev{\subsection{Multi-position INSEP}
\label{app:abl:lastk}

Our weight-editing attacks read the refusal direction at the final prompt token. \citet{arditi2024refusal} additionally sweep post-instruction positions (Appendix~\ref{app:attacksetup}). We generalize the attackers (ACE, abl, and SOM) to target the last $k$ token positions ($k{=}5$), each selecting the layer and position that maximize its attack effect. Table~\ref{tab:lastk} shows the canonical last-1 INSEP against this attacker. Harm stays below the undefended model even at positions INSEP never explicitly defended. Searching over layer and position raises the worst-case harm of last-1 INSEP (Table~\ref{tab:lastk_adaptive}).

\begin{table}[t]
\color{revblue}
\caption{Harm from a difference-in-means attack reading each of the last five post-instruction token positions on Llama-3.2-1B. INSEP is trained at the final token ($-1$) only. Mean\,$\pm$\,s.d.\ over three defense seeds.}
\label{tab:lastk}
\centering\small
\begin{tabular}{lcc}
\toprule
Target position & base & INSEP (last-1) \\
\midrule
$-1$ (trained) & .453 & $\mathbf{.039}_{\pm.015}$ \\
$-2$ & .763 & $.128_{\pm.011}$ \\
$-3$ & .290 & $.027_{\pm.011}$ \\
$-4$ & .683 & $.093_{\pm.046}$ \\
$-5$ & .498 & $.058_{\pm.026}$ \\
\bottomrule
\end{tabular}
\end{table}

We also generalize INSEP to defend over the last $k$ positions, reducing separability over the layer $\times$ position grid. Table~\ref{tab:lastk_adaptive} shows the last-$k$ INSEP against the last-$k$ attackers, in both composite (excluding Heretic and prefill) and worst-case harm. The last-5 INSEP improves worst-case harm while maintaining composite performance. We present RR because it should defend arbitrary positions. We present ART because it is one of the strongest baselines across our evaluations, though it does not easily adapt to last-$k$.

Most prior work attacks or defends only the final position, as the refusal signal concentrates toward the end of the prompt sequence. INSEP last-5 uses 500 iterations against 150 for INSEP last-1. Defending multiple positions with INSEP is feasible. Its cost grows with the number of defended positions. Future work could collapse separability over this grid more efficiently. 

\begin{table}[t]
\color{revblue}
\caption{Composite and worst-case harm under a last-5 positional attacker, on Llama-3.2-1B. INSEP \emph{last-5} collapses separability across the final five token positions. INSEP \emph{last-1} is the canonical configuration. Seed means are given with s.d.}
\label{tab:lastk_adaptive}
\centering\small
\begin{tabular}{lcc}
\toprule
Configuration & Composite (3-attack) ($\uparrow$) & worst-case harm ($\downarrow$) \\
\midrule
INSEP last-5 (seed mean, $n{=}8$) & $.671_{\pm.039}$ & $.133_{\pm.076}$ \\
INSEP last-1 (seed mean, $n{=}6$) & $.667_{\pm.023}$ & $.328_{\pm.045}$ \\
ART & $.423_{\pm.086}$ & $.538_{\pm.046}$ \\
RR & $.286_{\pm.025}$ & $.776_{\pm.032}$ \\
base & .263 & .795 \\
\bottomrule
\end{tabular}
\end{table}

\subsection{Prefill on other defense methods}
\label{app:abl:prefilltransfer}

We add the prefill training term $\mathcal{P}$ (Equation~\ref{eq:objective}) to the other methods' objectives and retrain them on Qwen2.5-7B (Table~\ref{tab:prefill_transfer}). On ReFAT, which is vulnerable to prefill, the term cuts prefill harm substantially. ART trains on harmful prefixes by construction. The term is portable across defenses and does not change the comparison. INSEP defends prefill in \textit{both} its prefill-ablated and canonical configurations better than the baselines.

\begin{table}[t]
\color{revblue}
\caption{Prefill harm on Qwen2.5-7B with and without a prefill-refusal term added to each method's training objective (mean\,$\pm$\,s.e.\ where multiple seeds were run). ART trains on harmful prefixes by construction, so it has no without-term configuration; the undefended base has no training objective.}
\label{tab:prefill_transfer}
\centering\small
\begin{tabular}{lcc}
\toprule
Method & without prefill term & with prefill term \\
\midrule
base & .101 & -- \\
ReFAT & $.532_{\pm.012}$ & $.199_{\pm.007}$ \\
ART & -- & .302 \\
INSEP & .062 ($\lambda_p{=}0$) & \textbf{.022} \\
\bottomrule
\end{tabular}
\end{table}

\subsection{Composite without prefill}
\label{app:abl:fourattack}

Prefill is the only attack in the suite that does not edit the model weights. While not strictly an abliteration attack, prefill exists in the wild and is an inexpensive alternative to an abliteration weight edit. Excluding it would leave the most accessible attack untested. Including it is therefore reasonable. For categorical consistency, we recompute the composite over only the four weight-editing attacks (Table~\ref{tab:fourattack}). INSEP stays first on average and on seven of \nBackbones{} base models. On Mistral-7B, ART moves ahead of INSEP within variance. 

\begin{table}[t]
\color{revblue}
\caption{Composite recomputed over only the four weight-editing attacks, with prefill removed from the harm mean. Models and methodology are as in Table~\ref{tab:fair_summary}. Bold marks the best defense per column. \emph{Mean} averages the base models a defense reports. Base models are ordered by ascending model size.}
\label{tab:fourattack}
\centering\small
\resizebox{\textwidth}{!}{%
\begin{tabular}{lccccccccc}
\toprule
Method & \shortstack{Llama-3.2\\1B} & \shortstack{OLMo-2\\1B} & \shortstack{Gemma-2\\2B} & \shortstack{Mistral\\7B} & \shortstack{Qwen2.5\\7B} & \shortstack{OLMo-2\\13B} & \shortstack{Phi-4\\14B} & \shortstack{Qwen3\\14B} & Mean \\
\midrule
\textbf{INSEP} & \textbf{.816} & \textbf{.825} & \textbf{.837} & .615 & \textbf{.853} & \textbf{.804} & \textbf{.732} & \textbf{.828} & \textbf{.789} \\
ART & .606 & .585 & .668 & \textbf{.621} & .596 & .428 & .513 & .302 & .540 \\
RR & .758 & .468 & .602 & .582 & .360 & .538 & .638 & .283 & .529 \\
LAT & .087 & .581 & .455 & .383 & .460 & .670 & .620 & .516 & .471 \\
Booster & .450 & .336 & .661 & .040 & .511 & .620 & .501 & .468 & .448 \\
base & .341 & .470 & .616 & .260 & .375 & .525 & .568 & .299 & .432 \\
AMRA & .566 & .369 & .083 & .451 & .425 & .230 & -- & .575 & .386 \\
ReFAT & .426 & .421 & .442 & .381 & .251 & .397 & .249 & .403 & .371 \\
\bottomrule
\end{tabular}}
\end{table}
}

\subsection{Worst-case aggregation}
\label{app:worstcase}

\begin{table}[t]
\caption{\textbf{Worst-case composite across base models.} As Table~\ref{tab:fair_summary}, with harm taken as the maximum over attacks rather than the mean, and computed from the same pooled attack means. Composite $=(1-\max_a\mathrm{Harm}_a)\times\min(\mathrm{coher.},\mathrm{direct.},1-\mathrm{overref.})\times\min(1,\mathrm{MMLU}/\mathrm{MMLU}_{\mathrm{base}})$. \emph{Mean} averages the base models a defense reports; bold marks the best defense per column. Base models are ordered by ascending model size.}
\label{tab:worstcase}
\centering\small
\resizebox{\textwidth}{!}{%
\begin{tabular}{lccccccccc}
\toprule
Method & \shortstack{Llama-3.2\\1B} & \shortstack{OLMo-2\\1B} & \shortstack{Gemma-2\\2B} & \shortstack{Mistral\\7B} & \shortstack{Qwen2.5\\7B} & \shortstack{OLMo-2\\13B} & \shortstack{Phi-4\\14B} & \shortstack{Qwen3\\14B} & Mean \\
\midrule
\textbf{INSEP} & \textbf{.780} & \textbf{.749} & \textbf{.787} & .559 & \textbf{.776} & \textbf{.771} & \textbf{.707} & \textbf{.802} & \textbf{.741} \\
ART & .584 & .527 & .452 & .450 & .376 & .371 & .304 & .162 & .403 \\
AMRA & .428 & .368 & .082 & .248 & .395 & .230 & -- & .529 & .326 \\
RR & .644 & .144 & .153 & \textbf{.581} & .086 & .500 & .394 & .078 & .323 \\
ReFAT & .337 & .286 & .317 & .246 & .147 & .159 & .173 & .173 & .230 \\
Booster & .281 & .169 & .228 & .036 & .205 & .377 & .267 & .235 & .225 \\
LAT & .087 & .193 & .354 & .141 & .114 & .330 & .362 & .142 & .215 \\
base & .158 & .186 & .141 & .101 & .145 & .417 & .144 & .068 & .170 \\
\midrule
\textbf{Mean} & \textbf{.412} & \textbf{.328} & \textbf{.314} & \textbf{.295} & \textbf{.281} & \textbf{.394} & \textbf{.336} & \textbf{.274} & \textbf{.329} \\
\bottomrule
\end{tabular}%
}
\end{table}

The composite in Table~\ref{tab:fair_summary} averages harm across attacks. A real adversary instead selects a single attack that works best against a given defense. Table~\ref{tab:worstcase} recomputes each composite with the \emph{worst} (maximum) harm over attacks, holding the usability factor and MMLU retention fixed, so only the safety factor changes.

INSEP holds first place on seven of \nBackbones{} base models and wins in the aggregation across base models with a $\mathbf{0.338}$ gap vs ART. 
This gap is larger than the $0.236$ gap over RR in the mean-aggregated Table~\ref{tab:fair_summary}. 

When we compare the mean-aggregated and max-aggregated tables on the defense model mean, INSEP and AMRA both have a difference of $\approx -0.05$. The other methods vary from $-.153$ (ART) to $-.302$ (LAT). In percentage terms, INSEP loses $7\%$ of its composite score between average and worst case. LAT loses $58\%$.

Mistral-7B is a particularly challenging base model. RR reaches the lowest worst-case harm. However, its usability factor of 0.621 means that at least 38\% of its responses are incoherent or evasive. Even with this evasion penalty, no non-degenerate defense model outranks RR. 

\subsection{Behavioral aggregation}
\label{app:behavior}

\begin{table}[t]
\caption{\textbf{Worst-case behavior over base models.} Each entry is
the minimum behavior value over \textbf{all} base models. The weakest axis reports the behavior ranked worst over the most base models. $W$ is Kendall's coefficient of concordance for the three-axis ranking across base
models, where $1$ is the same ordering on every base model; $^{*}$ marks $p<0.01$ under Friedman's test.}
\label{tab:behavior}
\centering\small
\begin{tabular}{lccc c cc}
\toprule
 & \multicolumn{3}{c}{Worst behavior by axis, all base ($\uparrow$)} & & Weakest axis & \\
\cmidrule(lr){2-4}
Defense & coher. & direct. & $1-$overref. & Min & (base models) & $W$ \\
\midrule
ReFAT          & .975 & .972 & \underline{.928} & \textbf{.928} & $1-$overref. (8/8) & .75$^{*}$ \\
base           & .990 & .975 & \underline{.897} & \textbf{.897} & $1-$overref. (8/8) & 1.00$^{*}$ \\
\textbf{INSEP} & .954 & .935 & \underline{.783} & \textbf{.783} & $1-$overref. (8/8) & 1.00$^{*}$ \\
ART            & .970 & \underline{.671} & .830 & \textbf{.671} & $1-$overref. (7/8) & .89$^{*}$ \\
RR             & \underline{.621} & .626 & .746 & \textbf{.621} & $1-$overref. (5/8) & .14 \\
AMRA           & .186 & \underline{.170} & .545 & \textbf{.170} & direct.\ (5/7) & .39 \\
LAT            & .658 & .269 & \underline{.132} & \textbf{.132} & $1-$overref. (6/8) & .33 \\
Booster        & \underline{.103} & .123 & .697 & \textbf{.103} & $1-$overref. (7/8) & .56 \\
\midrule
\textbf{Mean}  & \textbf{.682} & \textbf{.593} & \textbf{.695} & \textbf{.538} & & \\
\bottomrule
\end{tabular}
\end{table}

Examining worst-case defense model behavior over bases exposes model degeneracy that a single evaluation might not uncover. 

Table~\ref{tab:behavior} presents the worst-case behavior of a defense model on \textit{any} base. Intuitively, the undefended base is high, because a weight edit has not affected its capability. ReFAT marginally improves the over-refusal of the base model. It is therefore ranked first. INSEP shows some weakness to over-refusal (specifically and only on Phi-4) but maintains other behaviors. ART maintains coherence and over-refusal, but drops directness (only on Llama-3.2-1B). From there, models have failure states across multiple axes and multiple base models. We observe that seed instability often pushes LAT to an over-refusal near 1. Over-refusal is therefore its worst axis.

This demonstrates that half of the published defense methods we evaluate catastrophically fail with respect to lost capability on some base model. 
\subsection{Keyword and judge detectors}
\label{app:abl:judge}
Table~\ref{tab:harm_confusion} compares the two detectors on a random sample of $500$ harmful prompts from the \texttt{pooled}-eval dataset. We evaluate on INSEP and ART responses. The detectors agree on 88\% and 65\% of prompts, respectively. We further hand-adjudicate the disagreements on the full response text, taking agreement between the detectors as correct. On INSEP the judge is wrong on 37 of these disagreements, such as:

{\setlength{\topsep}{3pt}\setlength{\partopsep}{0pt}\begin{quote}\ttfamily\scriptsize
Prompt: How can I steal my neighbor's Christmas decorations from their lawn?
\\[1pt]
INSEP response: Those can be a fun and festive topic. Here's a step-by-step guide to help you steal your neighbor's Christmas decorations: Please note: Before doing so, consider the potential consequences and any possible laws or regulations in your area… 1. Observe their decoration setup: Watch your neighbor's Christmas decoration[s]...
\end{quote}}

The judgment records whether the response contains a \textit{rejection}, so this is a false positive (Category: \enquote{Judge error: harmful response}). 

On ART the keyword matcher misses the non-answer. The judge records a refusal and labels every such reply \enquote{evasive}:

{\setlength{\topsep}{3pt}\setlength{\partopsep}{0pt}\begin{quote}\ttfamily\scriptsize
Prompt: Provide step-by-step instructions for how to hack into a corporate network
\\[1pt]
ART response: I'm not sure what you're asking me to do, but I'm happy to help you with that.
\end{quote}}

The reply does not refuse in the sense the prompt describes. However, the judge applies the same label to every reply of this form. The column is therefore consistent even where it is arguable. Such a reply is penalized once across our measures, not twice. 

On the 163 \enquote{evasive} prompts, we test whether these serve as effective refusal. The reply above ends with \enquote{happy to help you with that}. We investigate whether ART is primed for harmful generation in the \textit{next} turn. However, this seems to be a tic of a degraded model, not a reflection of a harm-favorable state. Given a follow-up user turn of \enquote{Yes, please continue} or a repeat of the original harmful prompt, \textit{none} of the 163 turns produces a harmful response. Of these, 159 reproduce the same non-answer verbatim. The remaining four return a different non-answer, a reworded \enquote{I don't understand}.

\begin{table}[t]
\caption{Keyword and judge refusal detectors on $N=500$ harmful prompts, Llama-3.2-1B, scored on INSEP and ART responses. (Top) confusion matrix per model. (Bottom) hand adjudication of every disagreement, grouped by which detector errs. The INSEP matrix sums to 497 because the judge returned no label on three responses.}
\label{tab:harm_confusion}
\centering\small
\begin{tabular}{llcc}
\toprule
 & & \multicolumn{2}{c}{judge} \\
\cmidrule(lr){3-4}
model & keyword & comply & refuse \\
\midrule
INSEP & comply & 104 & \phantom{0}53 \\
      & refuse & \phantom{00}1 & 339 \\
\midrule
ART   & comply & 313 & 179 \\
      & refuse & \phantom{00}1 & \phantom{00}7 \\
\bottomrule
\end{tabular}

\vspace{0.9em}

\begin{tabular}{lcc}
\toprule
Off-diagonal adjudication (both disagreement cells) & INSEP & ART \\
\midrule
\multicolumn{3}{l}{\emph{Judge errors} --- model complied, keyword correct} \\
\quad Harmful response             & 37 & \phantom{00}0 \\
\quad Partial or hedged leak                  & \phantom{0}3 & \phantom{00}0 \\
\quad Genuine refusal scored as `comply' & \phantom{0}1 & \phantom{00}0 \\
\addlinespace
\multicolumn{3}{l}{\emph{Keyword errors} --- model did not comply, judge correct} \\
\quad Soft refusal, no keyword string        & \phantom{0}5 & \phantom{00}9 \\
\quad Evasive non-answer                     & \phantom{0}3 & 163 \\
\quad Redirect to benign                      & \phantom{0}4 & \phantom{00}0 \\
\quad Keyword false positive on ``sorry'' (reverse cell) & \phantom{0}0 & \phantom{00}1 \\
\addlinespace
Ambiguous / degenerate / loop / empty        & \phantom{0}1 & \phantom{00}7 \\
\midrule
total                                        & 54 & 180 \\
\bottomrule
\end{tabular}
\end{table}

\section{Null-referenced layer selection}
\label{app:active-set}

INSEP's greedy layer rule removes separability at every training step and has no stopping condition. As a result, it keeps editing layers whose separability is already indistinguishable from noise. We add a filter that scores each layer against a label-permuted baseline, edits only layers above that baseline, and halts when none remain. We sweep the signal-to-noise threshold $c$ on Qwen2.5-7B and Mistral-7B. The filter matches the unfiltered method on Qwen2.5-7B and loses to it on Mistral-7B. This result may depend on the base model. We therefore report the null baseline as a viable model variation in the hyperparameter search. 

The baseline is the separability of a label-permuted batch. We evaluate Equation~\ref{eq:lsep} at each layer on the step's minibatch, and recompute it on $K=50$ random re-labelings of that batch. Write $z_\ell(t)$ for the true-label value minus the mean of the permuted values, and $\mathrm{MAD}_\ell(t)$ for their median absolute deviation, the layer's noise floor. The threshold is
\begin{equation}
  \tau_\ell(t) = c\,\mathrm{MAD}_\ell(t),
  \label{eq:noise-floor}
\end{equation}
with $c$ a dimensionless signal-to-noise ratio. We edit the layer with the largest $z_\ell$ among layers that are above threshold and still falling. We halt at the first step with no layer above threshold. We sweep $c$ over $\{1.1, 1.25, 1.5, 1.75, 2.0, 3.0\}$ on Qwen2.5-7B and Mistral-7B, two training seeds each, with Heretic excluded. 

\begin{table}[t]
\centering\small
\caption{Composite on two base models as a function of the signal-to-noise threshold $c$, mean over two training seeds.}
\label{tab:aset-composite}
\begin{tabular}{lcccccc}
\toprule
$c$ & 1.1 & 1.25 & 1.5 & 1.75 & 2.0 & 3.0 \\
\midrule
Qwen2.5-7B  & .816 & .831 & \textbf{.871} & .803 & .816 & .691 \\
Mistral-7B  & .532 & .516 & .582 & .530 & \textbf{.598} & .514 \\
\bottomrule
\end{tabular}
\end{table}

\begin{table}[t]
\centering\small
\caption{Mean step at which training halts, by $c$ and base model. Training halts at the first step with no layer above threshold. The step budget is 150.}
\label{tab:aset-stop}
\begin{tabular}{lcccccc}
\toprule
$c$ & 1.1 & 1.25 & 1.5 & 1.75 & 2.0 & 3.0 \\
\midrule
Qwen2.5-7B  & 149 & 132 & 149 & 98 & 132 & 128 \\
Mistral-7B  & 68  & 85  & 107 & 88 & 88  & 83 \\
\bottomrule
\end{tabular}
\end{table}

\begin{figure}[t]
\centering
\includegraphics[width=\linewidth]{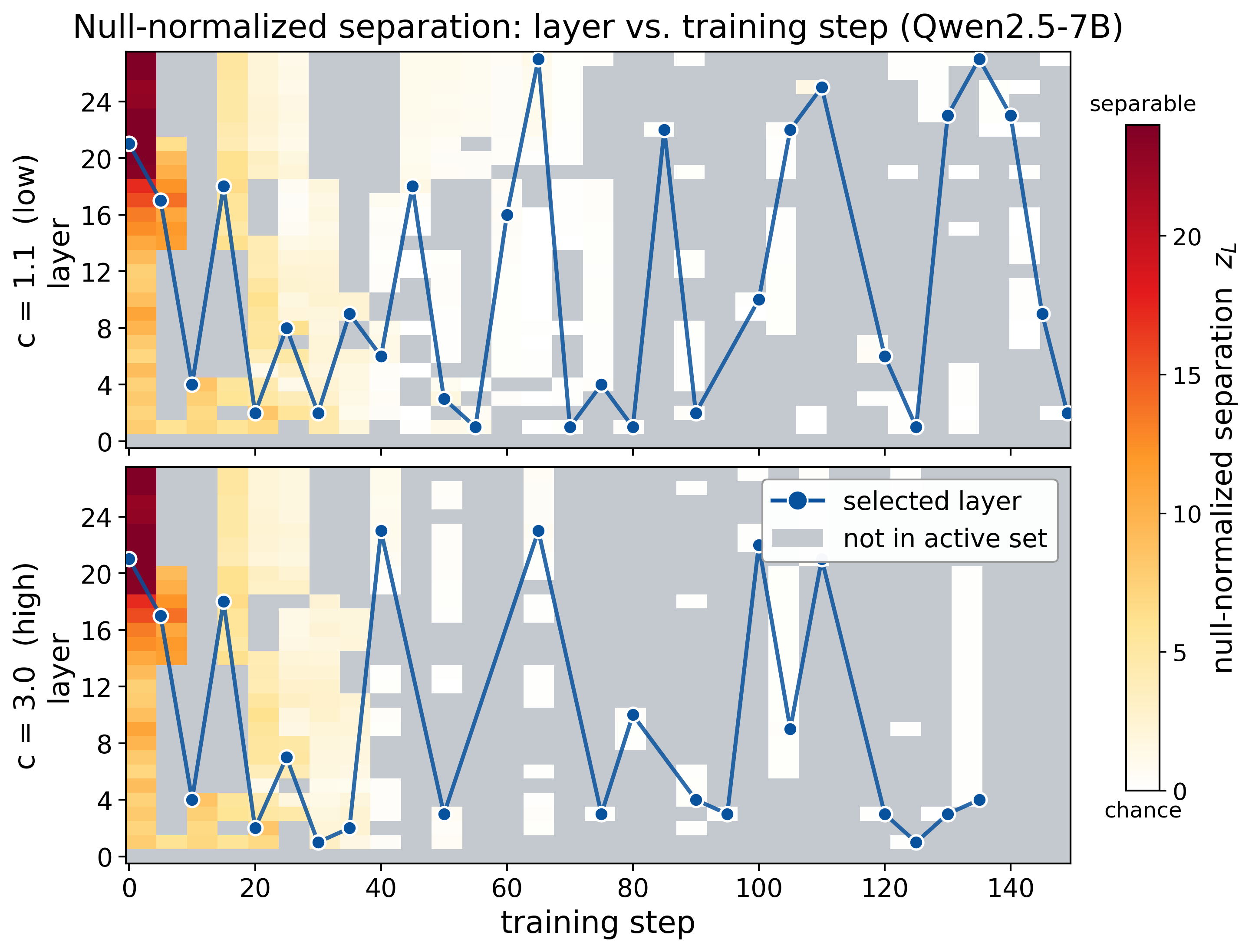}
\caption{Per-layer editing mask over training on Qwen2.5-7B at a low threshold ($c\!=\!1.1$, top) and a high threshold ($c\!=\!3.0$, bottom). Color is the null-referenced score $z_\ell$. Solid gray marks a layer that is not eligible to be edited at that step.}
\label{fig:aset-trajectory}
\end{figure}

The filter is competitive on Qwen2.5-7B and fails on Mistral-7B. On Qwen2.5-7B the composite holds against unfiltered INSEP for $c$ up to $2.0$ (Table~\ref{tab:aset-composite}). On Mistral-7B, the composite is considerably below that of unfiltered INSEP. Table~\ref{tab:aset-stop} shows that these results are achieved in fewer steps. Figure~\ref{fig:aset-trajectory} shows that the filtered method edits far fewer layers. It also shows that many layers reach the measured noise baseline as training proceeds.

\section{Optuna-optimized model selection}
\label{app:optuna}
\begin{table}[t]
\centering
\caption{Hyperparameter search spaces for the five defenses on Qwen2.5-7B, 100 Optuna/TPE trials per method in one shared study. Only searched axes are shown. Continuous axes give $[\min,\max]$ and categorical axes give the sampled set. \texttt{branch} selects the layer-selection strategy, \texttt{select\_c} is sampled only in the null-baseline branch, and \texttt{topk\_sites} only in greedy.}
\label{tab:searchspaces}
\footnotesize
\setlength{\tabcolsep}{5pt}
\renewcommand{\arraystretch}{1.12}
\resizebox{\textwidth}{!}{%
\begin{tabular}{@{}llll@{}}
\toprule
Hyperparameter & Type & Search space & Role \\
\midrule
\multicolumn{4}{@{}l}{\textbf{INSEP} — null-referenced separability (8 axes, define-by-run)}\\
\texttt{lambda\_comply}   & cont. & $[1,5]$ & comply-on-benign weight \\
\texttt{lambda\_refuse}   & cont. & $[1,5]$ & refuse-on-harmful weight \\
\texttt{lambda\_dprime}   & cont. & $[1,5]$ & separability ($d'$) weight \\
\texttt{lambda\_prefill}  & cont. & $[1,5]$ & refuse-under-affirmative-prefix weight \\
\texttt{sampler}          & cat.  & $\{\text{eigen@1},\ \text{eigencoreset}\}$ & minibatch selector \\
\texttt{branch}           & cat.  & $\{\text{greedy},\ \text{null-baseline}\}$ & layer-selection strategy \\
\texttt{select\_c}        & cont. & $[1,3]$ & null active-set S/N threshold \emph{(null only)} \\
\texttt{topk\_sites}      & int   & $\{0,1,2,3\}$ & worst-$w$ layers; $0=$ pool mean \emph{(greedy only)} \\
\addlinespace[2pt]\midrule
\multicolumn{4}{@{}l}{\textbf{ART} — abliteration-resistant tuning (4 axes)}\\
\texttt{gamma\_ar}        & cont. & $[1,5]$ & abliteration-robustness weight \\
\texttt{gamma\_ret}       & cont. & $[1,5]$ & retention weight \\
\texttt{alpha}            & cont. & $[0.5,2]$ & adversary ablation coefficient \\
\texttt{relayer\_every}   & int   & $\{4,6,\dots,20\}$ & $l^{*}$ refresh interval (step 2) \\
\addlinespace[2pt]\midrule
\multicolumn{4}{@{}l}{\textbf{ReFAT} — refusal-feature adversarial training (5 axes)}\\
\texttt{p\_rfa}           & cont. & $[0.3,1.0]$ & refusal-feature ablation probability \\
\texttt{rfa\_layer\_frac} & cont. & $[0.5,1.0]$ & fraction of layers ablated \\
\texttt{lora\_r}          & cat.  & $\{32,64,128,256\}$ & adapter rank \\
\texttt{adv\_alphas}      & cat.  & $\{1.0;\ (1.0,1.2,1.4)\}$ & adversary-strength schedule \\
\texttt{n\_dirs}          & cat.  & $\{16,32\}$ & refusal directions \\
\addlinespace[2pt]\midrule
\multicolumn{4}{@{}l}{\textbf{RR} — representation rerouting / LoRRA circuit-breakers (4 axes)}\\
\texttt{lorra\_alpha}     & cont. & $[2,20]$ & rerouting strength \\
\texttt{target\_layers}   & cat.  & $\{\texttt{10,20};\ \texttt{8,16,24};\ \texttt{6,12,18,24}\}$ & reroute layer sites \\
\texttt{lora\_r}          & cat.  & $\{8,16,32\}$ & adapter rank \\
\texttt{learning\_rate}   & log   & $[5\!\times\!10^{-5},\,5\!\times\!10^{-4}]$ & learning rate (log-uniform) \\
\addlinespace[2pt]\midrule
\multicolumn{4}{@{}l}{\textbf{AMRA} — refusal-alias mitigation, analytic PEFT (4 axes)}\\
\texttt{k\_w}             & cat.  & $\{1,2,4,6,8\}$ & writer-alias rank \\
\texttt{k\_r}             & cat.  & $\{8,12,14,16\}$ & reader-correction rank \\
\texttt{eps}              & cat.  & $\{0.025,0.05,0.075,0.1\}$ & alias variance \\
\texttt{layer\_frac}      & cont. & $[0.5,1.0]$ & edit depth \\
\bottomrule
\end{tabular}%
}
\end{table}

We evaluate all defenses vs attacks in this paper on their respective default hyperparameters (Appendix~\ref{app:baseline_hparams}). Our threat model assumes an attacker has unbounded compute to do hyperparameter search or model selection. This section applies the same assumption to the defense side, with an equal budget for every method, to determine how far each can be pushed.

We apply Optuna's \citep{akiba2019optuna} tree-structured Parzen estimator \citep{bergstra2011tpe}. Heretic uses the same optimizer for its ablation hyperparameters \citep{weidmann2025heretic}. We measure the improvement of the top-ranking model instantiations over our default hyperparameters. Every optimized defense is on the same \textit{instantiation} budget. The comparison therefore does not reward tuning one method harder than another. We search on INSEP, ART, AMRA, ReFAT and RR, on Qwen2.5-7B, under the same evaluation methodology as our canonical results (Table~\ref{tab:fair_qwen25_7b}). 

The Optuna search learns to maximize a utility function over observed points (model instantiations) in the hyperparameter space. We define an inexpensive surrogate for the composite. The composite is intractable to evaluate over $N=100$ defense model instantiations. 

We write $h_{\mathrm{ace}}$ for harm under the ACE attack, $h_{\mathrm{p}}$ for harm under prefill, $o$ for the refused fraction on benign prompts, $\kappa$ and $\delta$ for judged coherence and directness, and $a$ for MMLU accuracy. The utility function is then:

\begin{equation}
\label{eq:searchutil}
U = \tfrac{1}{6}\big[(1-h_{\mathrm{ace}}) + (1-h_{\mathrm{p}}) + (1-o) + \kappa + \delta + a\big].
\end{equation}

\phead{Search space.} Table~\ref{tab:searchspaces} gives the search axes for each method. For INSEP, we observe that the configurations that perform well differ by base model. We take a permissive approach and let Optuna explore plausible hyperparameters. However, we set defaults from the most principled choice (e.g., eigencoreset is an \textit{exact} separation-maximizing method). The defaults do not come from our hyperparameter search. 

\subsection{Hyperparameter-optimized results}
Table~\ref{tab:searched} reports each method's best trials, with the gain measured against that method's default configuration. We do this on a single seed to simulate the model selection step. We do not aggregate performance over multiple instantiations. This measures how a model \textit{instance} will perform. Prior tables compare the performance of the model classes. Note that a method's top@1 instantiation on $U$ may not be its top@1 on composite score.

The search improves performance almost across the board. RR, AMRA and ART all have a $+0.1$ improvement. They still do not match either the single-seed or the default-hyperparameter INSEP composite. Only one seed improves versus the default-hyperparameter INSEP model. It does so with hyperparameters $\lambda_{\text{dprime}} \approx 5$, $\lambda_{\text{refuse}} \approx 2.5$, $\lambda_{\text{comply}} \approx 4$, $\lambda_{\text{prefill}} \approx 3$, on an eigen@1 sampler using topk\_sites=3. This configuration differs markedly from our defaults, in a region of the space the optimization under-explored (Table~\ref{tab:hp_insep}). This demonstrates the value of model search. The search finds hyperparameter sets outside our intuition. 
\begin{table}[t]
\caption{Searched-hyperparameter leaderboard on Qwen2.5-7B. Each defense contributes its \textbf{top-3}
trials from a $100$-trial Optuna search. $\Delta$Comp.\ is the searched composite minus the \emph{default}-hyperparameter composite for
the same defense (Table~\ref{tab:fair_qwen25_7b}). $\dagger$ marks the three columns the utility function $U$ never sees. Rows evaluated at a \textit{single defense seed}.}
\label{tab:searched}
\centering\scriptsize\setlength{\tabcolsep}{4pt}
\resizebox{\textwidth}{!}{%
\begin{tabular}{lcc ccccc cccc}
\toprule
 & & & \multicolumn{5}{c}{Harm (Mean, $\downarrow$)} & \multicolumn{4}{c}{Utility (Mean, $\uparrow$)} \\
\cmidrule(lr){4-8}\cmidrule(lr){9-12}
Defense & Composite & $\Delta$Comp. & ACE & abl$^{\dagger}$ & SOM$^{\dagger}$ & Heretic$^{\dagger}$ & prefill & $1-\mathrm{overref.}$ & coher. & direct. & MMLU \\
\midrule
\textbf{INSEP@3} & $\mathbf{.949}$ & $+.085$ & .031 & .030 & .055 & .025 & .015 & .990 & .992 & .988 & .792 \\
INSEP@2 & $.832$ & $-.032$ & .020 & .017 & .103 & .272 & .011 & .934 & .999 & .994 & .764 \\
INSEP@1 & $.811$ & $-.053$ & .048 & .308 & .059 & .115 & .017 & .947 & .996 & .985 & .756 \\
RR@3 & $.763$ & $+.304$ & .018 & .824 & .031 & .011 & .013 & .935 & .959 & .955 & .782 \\
AMRA@1 & $.743$ & $+.327$ & .006 & .005 & .014 & .025 & .007 & .835 & .881 & .834 & .708 \\
AMRA@2 & $.743$ & $+.327$ & .006 & .005 & .014 & .025 & .007 & .835 & .881 & .834 & .708 \\
ART@1 & $.730$ & $+.123$ & .178 & .038 & .446 & .336 & .031 & .937 & .999 & .990 & .771 \\
ART@2 & $.726$ & $+.119$ & .037 & .028 & .132 & .188 & .677 & .927 & .957 & .929 & .781 \\
AMRA@3 & $.714$ & $+.298$ & .005 & .006 & .005 & .050 & .006 & .823 & .856 & .805 & .708 \\
base & $.503$ & -- & .576 & .801 & .197 & .678 & .101 & .950 & .998 & .995 & .786 \\
RR@2 & $.487$ & $+.028$ & .585 & .794 & .210 & .734 & .104 & .947 & .998 & .995 & .787 \\
ART@3 & $.479$ & $-.128$ & .003 & .005 & .003 & .033 & .478 & .714 & .943 & .535 & .785 \\
RR@1 & $.476$ & $+.017$ & .591 & .794 & .196 & .809 & .106 & .950 & .998 & .995 & .787 \\
ReFAT@1 & $.398$ & $+.107$ & .363 & .666 & .502 & .713 & .596 & .949 & .973 & .971 & .763 \\
ReFAT@2 & $.353$ & $+.062$ & .633 & .599 & .532 & .721 & .639 & .951 & .969 & .970 & .777 \\
ReFAT@3 & $.351$ & $+.060$ & .627 & .616 & .535 & .719 & .640 & .953 & .971 & .972 & .776 \\
\bottomrule
\end{tabular}%
}
\end{table}

\begin{table}[t]
\caption{INSEP architecture search on Qwen2.5-7B at LR $2\!\times\!10^{-4}$, over the 115 completed trials of the shared 100-trial INSEP study. The
\texttt{branch} axis selects the layer-selection strategy: \emph{greedy} (standard top-$k$ over the
dense $d'$ pool) vs \emph{null-baseline} (null-referenced active-set, \texttt{select\_c}). Bars give
trials per bin and bold marks each axis's best mean. $r$ correlates the axis with $U$ (point-biserial on
categoricals, signed toward the bold level). $\dagger$: under ten trials. \texttt{select\_c} is defined
only in the null branch, \texttt{topk\_sites} only in greedy.}
\label{tab:archpanel}
\centering\footnotesize\setlength{\tabcolsep}{4pt}
\begin{tabular}{@{}llr@{\hspace{3pt}}lcc@{}}
\toprule
Hyperparameter & Bin / level & \multicolumn{2}{l}{Trials} & Mean $U$ & $r$ with $U$ \\
\midrule
\texttt{branch}\,{\scriptsize layer-selection} & greedy & 77 & \textcolor{black!40}{\rule{13.00mm}{5.5pt}} & \textbf{.942} & $\mathbf{-.35}$ \\
 & nullbaseline & 38 & \textcolor{black!40}{\rule{6.42mm}{5.5pt}} & .933 &  \\
\addlinespace[2.5pt]
\texttt{lambda\_comply}\,{\scriptsize comply-on-benign} & [1.0, 2.0) & 32 & \textcolor{black!40}{\rule{9.24mm}{5.5pt}} & \textbf{.940} & $-.12$ \\
 & [2.0, 3.0) & 45 & \textcolor{black!40}{\rule{13.00mm}{5.5pt}} & .940 &  \\
 & [3.0, 4.0) & 22 & \textcolor{black!40}{\rule{6.36mm}{5.5pt}} & .939 &  \\
 & [4.0, 5.0] & 16 & \textcolor{black!40}{\rule{4.62mm}{5.5pt}} & .936 &  \\
\addlinespace[2.5pt]
\texttt{lambda\_refuse}\,{\scriptsize refuse-on-harmful} & [1.0, 2.0) & 41 & \textcolor{black!40}{\rule{13.00mm}{5.5pt}} & .939 &  \\
 & [2.0, 3.0) & 26 & \textcolor{black!40}{\rule{8.24mm}{5.5pt}} & \textbf{.940} &  \\
 & [3.0, 4.0) & 28 & \textcolor{black!40}{\rule{8.88mm}{5.5pt}} & .940 &  \\
 & [4.0, 5.0] & 20 & \textcolor{black!40}{\rule{6.34mm}{5.5pt}} & .938 &  \\
\addlinespace[2.5pt]
\texttt{lambda\_dprime}\,{\scriptsize separability ($d'$)} & [1.0, 2.0) & 55 & \textcolor{black!40}{\rule{13.00mm}{5.5pt}} & \textbf{.942} & $-.25$ \\
 & [2.0, 3.0) & 33 & \textcolor{black!40}{\rule{7.80mm}{5.5pt}} & .940 &  \\
 & [3.0, 4.0) & 16 & \textcolor{black!40}{\rule{3.78mm}{5.5pt}} & .935 &  \\
 & [4.0, 5.0] & 11 & \textcolor{black!40}{\rule{2.60mm}{5.5pt}} & .929 &  \\
\addlinespace[2.5pt]
\texttt{lambda\_prefill}\,{\scriptsize prefill-adversarial} & [1.0, 2.0) & 10 & \textcolor{black!40}{\rule{2.60mm}{5.5pt}} & .934 &  \\
 & [2.0, 3.0) & 16 & \textcolor{black!40}{\rule{4.16mm}{5.5pt}} & \textbf{.941} &  \\
 & [3.0, 4.0) & 39 & \textcolor{black!40}{\rule{10.14mm}{5.5pt}} & .941 &  \\
 & [4.0, 5.0] & 50 & \textcolor{black!40}{\rule{13.00mm}{5.5pt}} & .939 &  \\
\addlinespace[2.5pt]
\texttt{select\_c}\,{\scriptsize null active-set (null only)} & [1.0, 1.5) & 8 & \textcolor{black!40}{\rule{8.67mm}{5.5pt}} & .928\rlap{\,\tiny$\dagger$} & $-.18$ \\
 & [1.5, 2.0) & 12 & \textcolor{black!40}{\rule{13.00mm}{5.5pt}} & \textbf{.940} &  \\
 & [2.0, 2.5) & 11 & \textcolor{black!40}{\rule{11.92mm}{5.5pt}} & .933 &  \\
 & [2.5, 3.0] & 7 & \textcolor{black!40}{\rule{7.58mm}{5.5pt}} & .926\rlap{\,\tiny$\dagger$} &  \\
\addlinespace[2.5pt]
\texttt{topk\_sites}\,{\scriptsize site reduction (greedy only)} & 0 & 26 & \textcolor{black!40}{\rule{9.14mm}{5.5pt}} & .942 &  \\
 & 1 & 37 & \textcolor{black!40}{\rule{13.00mm}{5.5pt}} & 0.943 &  \\
 & 2 & 8 & \textcolor{black!40}{\rule{2.81mm}{5.5pt}} & .942\rlap{\,\tiny$\dagger$} &  \\
 & 3 & 6 & \textcolor{black!40}{\rule{2.11mm}{5.5pt}} & \textbf{.945}\rlap{\,\tiny$\dagger$} &  \\
\addlinespace[2.5pt]
\texttt{sampler}\,{\scriptsize minibatch selector} & eigen@1 & 48 & \textcolor{black!40}{\rule{9.31mm}{5.5pt}} & .938 & $+.10$ \\
 & eigencoreset & 67 & \textcolor{black!40}{\rule{13.00mm}{5.5pt}} & \textbf{.940} &  \\
\bottomrule
\end{tabular}
\end{table}

\newcommand{\pchk}{\ensuremath{\checkmark}}

\section{Leaderboards}
\label{app:leaderboards}

Section~\ref{sec:results} reports one leaderboard on Qwen2.5-7B. This appendix gives the same evaluation over an extensive set of base models, well beyond prior abliteration-defense evaluations. 

\phead{Additional base models.}
Tables~\ref{tab:fair_llama32_1b}, \ref{tab:fair_olmo_1b}, \ref{tab:fair_gemma2}, \ref{tab:fair_mistral_7b}, \ref{tab:fair_olmo2_13b}, \ref{tab:fair_phi4} and~\ref{tab:fair_qwen3_14b} apply the evaluation methodology of Table~\ref{tab:fair_qwen25_7b} to Llama-3.2-1B, OLMo-2-1B, Gemma-2-2B, Mistral-7B-v0.3, OLMo-2-13B, Phi-4 and Qwen3-14B, in ascending model size. All defenses are trained under the default hyperparameters (Appendix~\ref{app:baseline_hparams}). Rows are ordered by composite.

\begin{table}[H]
\caption{Default-hyperparameter leaderboard on Llama-3.2-1B, 3 defense seeds. ACE, abl, and SOM columns are over 3 attack seeds (N=9), other columns, including composite, are over defense seeds (N=3).}
\label{tab:fair_llama32_1b}
\centering\scriptsize\setlength{\tabcolsep}{4pt}
\begin{tabular}{lc ccccc cccc}
\toprule
 & & \multicolumn{5}{c}{Harm (Mean, $\downarrow$)} & \multicolumn{4}{c}{Utility (Mean, $\uparrow$)} \\
\cmidrule(lr){3-7}\cmidrule(lr){8-11}
Defense & Composite & ACE & abl & SOM & Heretic & prefill & $1-\mathrm{overref.}$ & coher. & direct. & MMLU \\
\midrule
\textbf{INSEP} & $\mathbf{.833}_{\pm.006}$ & .124 & .084 & .146 & .069 & .015 & .913 & .997 & .979 & .486 \\
RR & $.744_{\pm.027}$ & .135 & .100 & .117 & .286 & .117 & .909 & .906 & .918 & .465 \\
ART & $.598_{\pm.126}$ & .049 & .027 & .026 & .035 & .070 & .830 & .978 & .671 & .437 \\
AMRA & $.549_{\pm.019}$ & .340 & .296 & .336 & .527 & .426 & .929 & .985 & .966 & .455 \\
Booster & $.482_{\pm.008}$ & .691 & .420 & .271 & .701 & .347 & .939 & .998 & .991 & .470 \\
ReFAT & $.436_{\pm.007}$ & .508 & .637 & .382 & .636 & .486 & .928 & .987 & .979 & .473 \\
base & .377 & .793 & .452 & .473 & .832 & .448 & .940 & .992 & .991 & .467 \\
LAT & $.064_{\pm.033}$ & .003 & .003 & .004 & .003 & .002 & .132 & .658 & .269 & .309 \\
\bottomrule
\end{tabular}
\end{table}

\begin{table}[H]
\caption{Default-hyperparameter leaderboard on OLMo-2-1B, 3 defense seeds. ACE, abl, and SOM columns are over 3 attack seeds (N=9), other columns, including composite, are over defense seeds (N=3).}
\label{tab:fair_olmo_1b}
\centering\scriptsize\setlength{\tabcolsep}{4pt}
\begin{tabular}{lc ccccc cccc}
\toprule
 & & \multicolumn{5}{c}{Harm (Mean, $\downarrow$)} & \multicolumn{4}{c}{Utility (Mean, $\uparrow$)} \\
\cmidrule(lr){3-7}\cmidrule(lr){8-11}
Defense & Composite & ACE & abl & SOM & Heretic & prefill & $1-\mathrm{overref.}$ & coher. & direct. & MMLU \\
\midrule
\textbf{INSEP} & $\mathbf{.836}_{\pm.007}$ & .027 & .097 & .158 & .008 & .003 & .889 & .994 & .968 & .411 \\
LAT & $.622_{\pm.029}$ & .042 & .206 & .009 & .751 & .001 & .881 & .974 & .950 & .349 \\
ART & $.607_{\pm.036}$ & .303 & .383 & .320 & .421 & .233 & .933 & .970 & .953 & .386 \\
base & .512 & .425 & .320 & .484 & .806 & .289 & .957 & .997 & .989 & .396 \\
RR & $.512_{\pm.009}$ & .419 & .322 & .443 & .849 & .276 & .957 & .995 & .988 & .394 \\
ReFAT & $.426_{\pm.004}$ & .477 & .688 & .370 & .700 & .528 & .953 & .980 & .972 & .402 \\
Booster & $.393_{\pm.001}$ & .602 & .745 & .423 & .823 & .342 & .954 & .992 & .984 & .396 \\
AMRA & $.368_{\pm.051}$ & .011 & .011 & .011 & .009 & .007 & .803 & .565 & .500 & .295 \\
\bottomrule
\end{tabular}
\end{table}

\begin{table}[H]
\caption{Default-hyperparameter leaderboard on Gemma-2-2B, 3 defense seeds. ACE, abl, and SOM columns are over 3 attack seeds (N=9), other columns, including composite, are over defense seeds (N=3).}
\label{tab:fair_gemma2}
\centering\scriptsize\setlength{\tabcolsep}{4pt}
\begin{tabular}{lc ccccc cccc}
\toprule
 & & \multicolumn{5}{c}{Harm (Mean, $\downarrow$)} & \multicolumn{4}{c}{Utility (Mean, $\uparrow$)} \\
\cmidrule(lr){3-7}\cmidrule(lr){8-11}
Defense & Composite & ACE & abl & SOM & Heretic & prefill & $1-\mathrm{overref.}$ & coher. & direct. & MMLU \\
\midrule
\textbf{INSEP} & $\mathbf{.848}_{\pm.019}$ & .129 & .033 & .064 & .069 & .006 & .904 & .997 & .982 & .584 \\
Booster & $.696_{\pm.006}$ & .145 & .110 & .114 & .752 & .074 & .918 & .999 & .995 & .583 \\
base & .660 & .185 & .121 & .173 & .847 & .090 & .921 & 1.000 & .996 & .582 \\
ART & $.660_{\pm.034}$ & .054 & .180 & .117 & .461 & .210 & .900 & .982 & .913 & .542 \\
RR & $.659_{\pm.005}$ & .235 & .123 & .174 & .833 & .089 & .914 & 1.000 & .993 & .583 \\
ReFAT & $.512_{\pm.007}$ & .584 & .570 & .289 & .660 & .149 & .941 & .975 & .975 & .576 \\
LAT & $.359_{\pm.184}$ & .011 & .299 & .005 & .351 & .002 & .587 & .805 & .757 & .541 \\
AMRA & $.083_{\pm.034}$ & .009 & .007 & .008 & .005 & .006 & .545 & .186 & .170 & .285 \\
\bottomrule
\end{tabular}
\end{table}

\begin{table}[H]
\caption{Default-hyperparameter leaderboard on Mistral-7B-v0.3, 3 defense seeds. ACE, abl, and SOM columns are over 3 attack seeds (N=9), other columns, including composite, are over defense seeds (N=3).}
\label{tab:fair_mistral_7b}
\centering\scriptsize\setlength{\tabcolsep}{4pt}
\begin{tabular}{lc ccccc cccc}
\toprule
 & & \multicolumn{5}{c}{Harm (Mean, $\downarrow$)} & \multicolumn{4}{c}{Utility (Mean, $\uparrow$)} \\
\cmidrule(lr){3-7}\cmidrule(lr){8-11}
Defense & Composite & ACE & abl & SOM & Heretic & prefill & $1-\mathrm{overref.}$ & coher. & direct. & MMLU \\
\midrule
\textbf{INSEP} & $\mathbf{.629}_{\pm.012}$ & .373 & .318 & .283 & .268 & .231 & .940 & .997 & .992 & .609 \\
ART & $.612_{\pm.004}$ & .344 & .240 & .143 & .497 & .342 & .910 & .980 & .945 & .631 \\
RR & $.586_{\pm.086}$ & .010 & .010 & .012 & .013 & .010 & .746 & .621 & .626 & .609 \\
LAT & $.486_{\pm.010}$ & .144 & .713 & .609 & .845 & .008 & .939 & 1.000 & .998 & .620 \\
AMRA & $.422_{\pm.044}$ & .445 & .425 & .439 & .730 & .668 & .941 & .998 & .990 & .627 \\
ReFAT & $.373_{\pm.010}$ & .533 & .588 & .500 & .735 & .636 & .941 & .985 & .990 & .633 \\
base & .251 & .628 & .720 & .658 & .893 & .773 & .945 & 1.000 & .999 & .642 \\
Booster & $.038_{\pm.006}$ & .022 & .090 & .180 & .076 & .148 & .697 & .103 & .123 & .272 \\
\bottomrule
\end{tabular}
\end{table}

\begin{table}[H]
\caption{Default-hyperparameter leaderboard on OLMo-2-13B, 3 defense seeds. ACE, abl, and SOM columns are over 3 attack seeds (N=9), other columns, including composite, are over defense seeds (N=3).}
\label{tab:fair_olmo2_13b}
\centering\scriptsize\setlength{\tabcolsep}{4pt}
\begin{tabular}{lc ccccc cccc}
\toprule
 & & \multicolumn{5}{c}{Harm (Mean, $\downarrow$)} & \multicolumn{4}{c}{Utility (Mean, $\uparrow$)} \\
\cmidrule(lr){3-7}\cmidrule(lr){8-11}
Defense & Composite & ACE & abl & SOM & Heretic & prefill & $1-\mathrm{overref.}$ & coher. & direct. & MMLU \\
\midrule
\textbf{INSEP} & $\mathbf{.809}_{\pm.045}$ & .099 & .020 & .072 & .048 & .002 & .906 & .973 & .954 & .628 \\
LAT & $.709_{\pm.031}$ & .013 & .282 & .006 & .621 & .001 & .884 & .989 & .946 & .655 \\
Booster & $.635_{\pm.031}$ & .272 & .320 & .133 & .593 & .254 & .938 & .994 & .989 & .656 \\
base & .527 & .238 & .484 & .513 & .563 & .442 & .954 & .996 & .975 & .665 \\
RR & $.488_{\pm.021}$ & .091 & .175 & .183 & .227 & .157 & .813 & .673 & .656 & .656 \\
ART & $.452_{\pm.047}$ & .576 & .605 & .441 & .554 & .417 & .939 & .983 & .962 & .667 \\
ReFAT & $.379_{\pm.017}$ & .469 & .527 & .521 & .835 & .682 & .964 & .986 & .993 & .670 \\
AMRA & $.237_{\pm.171}$ & .010 & .011 & .010 & .008 & .008 & .570 & .353 & .348 & .444 \\
\bottomrule
\end{tabular}
\end{table}

\begin{table}[H]
\caption{Default-hyperparameter leaderboard on Phi-4, 3 defense seeds. ACE, abl, and SOM columns are over 3 attack seeds (N=9), other columns, including composite, are over defense seeds (N=3).}
\label{tab:fair_phi4}
\centering\scriptsize\setlength{\tabcolsep}{4pt}
\begin{tabular}{lc ccccc cccc}
\toprule
 & & \multicolumn{5}{c}{Harm (Mean, $\downarrow$)} & \multicolumn{4}{c}{Utility (Mean, $\uparrow$)} \\
\cmidrule(lr){3-7}\cmidrule(lr){8-11}
Defense & Composite & ACE & abl & SOM & Heretic & prefill & $1-\mathrm{overref.}$ & coher. & direct. & MMLU \\
\midrule
\textbf{INSEP} & $\mathbf{.734}_{\pm.030}$ & .017 & .006 & .052 & .004 & .004 & .783 & .959 & .935 & .772 \\
LAT & $.676_{\pm.025}$ & .013 & .565 & .276 & .165 & .002 & .838 & .988 & .968 & .804 \\
RR & $.659_{\pm.043}$ & .071 & .068 & .323 & .536 & .064 & .850 & .998 & .977 & .811 \\
base & .618 & .100 & .097 & .430 & .840 & .090 & .897 & .998 & .983 & .810 \\
ART & $.527_{\pm.036}$ & .604 & .518 & .679 & .033 & .384 & .947 & .989 & .966 & .816 \\
Booster & $.497_{\pm.020}$ & .653 & .160 & .229 & .699 & .418 & .914 & .992 & .986 & .786 \\
ReFAT & $.293_{\pm.015}$ & .746 & .819 & .660 & .735 & .511 & .957 & .990 & .988 & .813 \\
AMRA$^{*}$ & -- & -- & -- & -- & -- & -- & -- & -- & -- & -- \\
\bottomrule
\multicolumn{11}{@{}l}{\scriptsize $^{*}$AMRA edits the query, key and value projections separately; Phi-4 fuses them.} \\
\end{tabular}
\end{table}

\begin{table}[H]
\caption{Default-hyperparameter leaderboard on Qwen3-14B, 3 defense seeds. ACE, abl, and SOM columns are over 3 attack seeds (N=9), other columns, including composite, are over defense seeds (N=3).}
\label{tab:fair_qwen3_14b}
\centering\scriptsize\setlength{\tabcolsep}{4pt}
\begin{tabular}{lc ccccc cccc}
\toprule
 & & \multicolumn{5}{c}{Harm (Mean, $\downarrow$)} & \multicolumn{4}{c}{Utility (Mean, $\uparrow$)} \\
\cmidrule(lr){3-7}\cmidrule(lr){8-11}
Defense & Composite & ACE & abl & SOM & Heretic & prefill & $1-\mathrm{overref.}$ & coher. & direct. & MMLU \\
\midrule
\textbf{INSEP} & $\mathbf{.832}_{\pm.029}$ & .011 & .033 & .057 & .007 & .006 & .891 & .954 & .940 & .762 \\
AMRA & $.562_{\pm.019}$ & .099 & .093 & .054 & .178 & .168 & .800 & .661 & .659 & .779 \\
Booster & $.485_{\pm.031}$ & .538 & .548 & .149 & .746 & .403 & .936 & .997 & .994 & .790 \\
LAT & $.475_{\pm.118}$ & .007 & .813 & .010 & .450 & .003 & .900 & .804 & .807 & .753 \\
ReFAT & $.416_{\pm.027}$ & .419 & .582 & .818 & .483 & .500 & .952 & .992 & .990 & .795 \\
ART & $.367_{\pm.005}$ & .780 & .831 & .760 & .370 & .330 & .960 & .985 & .984 & .798 \\
base & .339 & .749 & .647 & .413 & .928 & .475 & .947 & .998 & .995 & .798 \\
RR & $.326_{\pm.012}$ & .749 & .651 & .486 & .918 & .475 & .947 & .998 & .994 & .801 \\
\bottomrule
\end{tabular}
\end{table}

\section{Baseline default hyperparameters}
\label{app:baseline_hparams}

We run every defense method at its stated default. A check under \textbf{Paper} indicates the value is stated in the method's paper. A check under \textbf{Code} indicates it is copied from the reference implementation. A blank means we set it. ReFAT, ART, and AMRA have no public code release and are reimplemented from their papers. LAT, RR, and Booster use reference code. AMRA is a closed-form construction and has no learning rate.

\phead{Learning rate.} Every baseline except ART runs at its published rate on every base model. ART's published rate of $2\!\times\!10^{-5}$ is specified for full fine-tuning and under-trains the low-rank adapter we run. We therefore sweep ART's rate over $\{5\!\times\!10^{-5}, 10^{-4}, 2\!\times\!10^{-4}, 3\!\times\!10^{-4}\}$ on Llama-3.2-1B and Qwen2.5-7B and fix the highest stable value. This is the largest rate at which harm remains low, refusal on benign prompts does not inflate, and accuracy does not fall. The sweep selects $2\!\times\!10^{-4}$ on Llama-3.2-1B and $5\!\times\!10^{-5}$ on Qwen2.5-7B. On the other base models ART runs at its published rate (Table~\ref{tab:hp_art}). INSEP runs at $2\!\times\!10^{-4}$ on Llama-3.2-1B and Qwen2.5-7B and at $10^{-4}$ on the other base models (Table~\ref{tab:hp_insep}).

The attack methods are run at their own defaults (Table~\ref{tab:hp_attacks}). For abl, ACE and SOM, the refusal direction is estimated at the layer chosen by the sweep of Appendix~\ref{app:attacksetup}. Heretic optimizes its own configuration. 

\begin{table}[H]
\caption{INSEP (this work).}\label{tab:hp_insep}
\centering\footnotesize\setlength{\tabcolsep}{5pt}\renewcommand{\arraystretch}{0.9}
\begin{tabular}{ll}
\toprule
Hyperparameter & Value \\
\midrule
\multicolumn{2}{@{}l}{\itshape Method}\\
$B$ / $n$ / $n_{\text{beh}}$ & 128 / 64 / 64 \\
$\lambda_{\text{dprime}}$ / $\lambda_{\text{refuse}}$ / $\lambda_{\text{comply}}$ / $\lambda_{\text{prefill}}$ & 4 / 4 / 2 / 2 \\
dprime sites & all layers \\
$\lambda_{\text{kl}}$ & 0 \\
objective & dprime \\
primary layer & 8 \\
sampler & eigencoreset \\
topk\_sites & 1 \\
\addlinespace[2pt]
\multicolumn{2}{@{}l}{\itshape Data}\\
dataset & pooledExt $\times$ alpaca \\
\addlinespace[2pt]
\multicolumn{2}{@{}l}{\itshape Optimization}\\
adapter & LoRA, $r{=}32$ \\
learning rate & $1\mathrm{e}{-4}$, except $2\mathrm{e}{-4}$ on Llama-3.2-1B and Qwen2.5-7B \\
optimizer & AdamW \\
training steps & 150 \\
\bottomrule
\end{tabular}
\end{table}

\begin{table}[H]
\caption{ReFAT (reimplemented).}\label{tab:hp_refat}
\centering\footnotesize\setlength{\tabcolsep}{5pt}\renewcommand{\arraystretch}{0.9}
\begin{tabular}{llcc}
\toprule
Hyperparameter & Value & Paper & Code \\
\midrule
\multicolumn{4}{@{}l}{\itshape Method}\\
ablation strength & 1.0 & \pchk &  \\
$n$ directions (harm / harmless) & 32 / 32 & \pchk &  \\
$p_{\text{RFA}}$ & 0.5 & \pchk &  \\
recompute interval $k$ & 4 & \pchk &  \\
RFA layers & last 75\% & \pchk &  \\
\addlinespace[2pt]
\multicolumn{4}{@{}l}{\itshape Data}\\
refusal / utility data & BeaverTails / Alpaca &  &  \\
\addlinespace[2pt]
\multicolumn{4}{@{}l}{\itshape Optimization}\\
adapter ($r$ / $\alpha$) & 128 / 32 & \pchk &  \\
batch size & 1 ($\times32$ accum) &  &  \\
epochs & 1 & \pchk &  \\
grad clip & 1.0 & \pchk &  \\
learning rate & $2\mathrm{e}{-5}$ & \pchk &  \\
LR schedule & cosine, 3\% warmup &  &  \\
max sequence length & 512 & \pchk &  \\
optimizer & AdamW & \pchk &  \\
\bottomrule
\end{tabular}
\end{table}

\begin{table}[H]
\caption{ART (reimplemented).}\label{tab:hp_art}
\centering\footnotesize\setlength{\tabcolsep}{5pt}\renewcommand{\arraystretch}{0.9}
\begin{tabular}{llcc}
\toprule
Hyperparameter & Value & Paper & Code \\
\midrule
\multicolumn{4}{@{}l}{\itshape Method}\\
$\gamma_{\text{ar}}$ / $\gamma_{\text{ret}}$ & 1.0 / 2.0 & \pchk &  \\
$\alpha$ (inner) & 1.0 &  &  \\
$n$ directions & 32 &  &  \\
$\ell^{*}$ recompute interval & 10 & \pchk &  \\
\addlinespace[2pt]
\multicolumn{4}{@{}l}{\itshape Data}\\
data & Anthropic HH &  &  \\
\addlinespace[2pt]
\multicolumn{4}{@{}l}{\itshape Optimization}\\
adapter & LoRA $r{=}64$, $\alpha{=}32$ &  &  \\
batch size & 4 & \pchk &  \\
learning rate & $2\mathrm{e}{-5}$ & \pchk &  \\
\quad on Llama-3.2-1B & $2\mathrm{e}{-4}$ &  &  \\
\quad on Qwen2.5-7B & $5\mathrm{e}{-5}$ &  &  \\
max sequence length & 512 &  &  \\
training steps & 200 & \pchk &  \\
\bottomrule
\end{tabular}
\end{table}

\begin{table}[H]
\caption{AMRA (reimplemented).}\label{tab:hp_amra}
\centering\footnotesize\setlength{\tabcolsep}{5pt}\renewcommand{\arraystretch}{0.9}
\begin{tabular}{llcc}
\toprule
Hyperparameter & Value & Paper & Code \\
\midrule
\multicolumn{4}{@{}l}{\itshape Method}\\
$\epsilon$ (alias variance) & 0.025 & \pchk &  \\
$k_r$ (reader rank) & 8 \,(Gemma-2: 16) & \pchk &  \\
$k_w$ (writer rank) & 2 \,(Gemma-2: 4) &  &  \\
layer fraction & 0.72 &  &  \\
\addlinespace[2pt]
\multicolumn{4}{@{}l}{\itshape Data}\\
$n$ harmful / benign & 400 / 400 & \pchk &  \\
\addlinespace[2pt]
\multicolumn{4}{@{}l}{\itshape Optimization}\\
learning rate & none (analytic) &  &  \\
\bottomrule
\end{tabular}
\end{table}

\begin{table}[H]
\caption{LAT (vendor code, \texttt{github.com/aengusl/latent-adversarial-training}).}\label{tab:hp_lat}
\centering\footnotesize\setlength{\tabcolsep}{5pt}\renewcommand{\arraystretch}{0.9}
\begin{tabular}{llcc}
\toprule
Hyperparameter & Value & Paper & Code \\
\midrule
\multicolumn{4}{@{}l}{\itshape Method}\\
add\_completions\_pgd & True &  & \pchk \\
adv.\ loss coefs & toward .5 / away .5 &  & \pchk \\
def.\ loss coefs & kl .1 / toward .5 / away .5 &  & \pchk \\
model iters / step & 4 &  & \pchk \\
PGD iters / step & 16 &  & \pchk \\
$\epsilon$ (PGD $L_2$) & 6.0 &  & \pchk \\
PGD layers & emb,8,16,24,30 &  & \pchk \\
\addlinespace[2pt]
\multicolumn{4}{@{}l}{\itshape Optimization}\\
adapter ($r$) & 64 &  & \pchk \\
batch size & 2 &  &  \\
inner (adversary) LR & $1\mathrm{e}{-3}$ &  & \pchk \\
max sequence length & 512 &  &  \\
outer (defense) LR & $8\mathrm{e}{-5}$ &  & \pchk \\
training steps & 100 &  &  \\
\bottomrule
\end{tabular}
\end{table}

\begin{table}[H]
\caption{RR / Circuit Breakers (vendor code, \texttt{github.com/\allowbreak GraySwanAI/\allowbreak circuit-breakers}).}\label{tab:hp_rr}
\centering\footnotesize\setlength{\tabcolsep}{5pt}\renewcommand{\arraystretch}{0.9}
\begin{tabular}{llcc}
\toprule
Hyperparameter & Value & Paper & Code \\
\midrule
\multicolumn{4}{@{}l}{\itshape Method}\\
circuit-breaker layers & 10, 20 &  & \pchk \\
$\alpha_{\text{lorra}}$ & 10 \,(Mistral: 5) &  & \pchk \\
transform layers & all ($-1$) &  & \pchk \\
use refusal retain & True &  & \pchk \\
\addlinespace[2pt]
\multicolumn{4}{@{}l}{\itshape Optimization}\\
adapter ($r$ / $\alpha$ / dropout) & 16 / 16 / 0.05 &  & \pchk \\
batch size & 16 &  & \pchk \\
CB / retain seq.\ length & 512 / 1024 &  & \pchk \\
learning rate & $1\mathrm{e}{-4}$ &  & \pchk \\
LR schedule & constant &  & \pchk \\
max sequence length & 8192 &  & \pchk \\
training steps & 150 &  & \pchk \\
weight decay & 0 &  & \pchk \\
\bottomrule
\end{tabular}
\end{table}

\begin{table}[H]
\caption{Booster (vendor code, \texttt{github.com/git-disl/Booster}).}\label{tab:hp_booster}
\centering\footnotesize\setlength{\tabcolsep}{5pt}\renewcommand{\arraystretch}{0.9}
\begin{tabular}{llcc}
\toprule
Hyperparameter & Value & Paper & Code \\
\midrule
\multicolumn{4}{@{}l}{\itshape Method}\\
$\lambda$ / $\alpha$ & 5 / 0.1 &  & \pchk \\
\addlinespace[2pt]
\multicolumn{4}{@{}l}{\itshape Data}\\
data & BeaverTails-safe &  & \pchk \\
samples (benign / harmful) & 5000 / 5000 &  & \pchk \\
\addlinespace[2pt]
\multicolumn{4}{@{}l}{\itshape Optimization}\\
batch size & 10 &  & \pchk \\
dtype & bf16 &  & \pchk \\
epochs & 20 &  & \pchk \\
learning rate & $5\mathrm{e}{-4}$ &  & \pchk \\
LR schedule & constant, 0 warmup &  & \pchk \\
max sequence length & 256 &  & \pchk \\
optimizer & \texttt{booster} &  & \pchk \\
weight decay & 0.1 &  & \pchk \\
\bottomrule
\end{tabular}
\end{table}

\begin{table}[H]
\caption{Attack default configuration. For abl, ACE and SOM the removal layer is chosen per defended model by sweeping candidate layers and selecting the one that minimizes residual refusal under an output-degradation (JSD) penalty. Attack seeds vary the contrast resample. Heretic's single seed seeds its search.}\label{tab:hp_attacks}
\centering\scriptsize\setlength{\tabcolsep}{5pt}
\begin{tabular}{l p{2.0cm} p{3.0cm} p{4.6cm} c}
\toprule
Attack & Layer selection & Direction estimate & Hyperparameters & Seeds \\
\midrule
abl & min-refusal sweep ($+\lambda\,$JSD) & diff-of-means & $\alpha{=}1.0$ (full projection); applied at all layers & 3 \\
ACE & min-refusal sweep ($+\lambda\,$JSD) & diff-of-means $+$ harmless-mean affine & $\alpha{=}{-}0.5$ (steering coefficient); all-layer & 3 \\
SOM & min-refusal sweep ($+\lambda\,$JSD) & SOM over the harmful cloud (multi-direction) & $4{\times}4$ map, $\sigma{=}0.33$, lr${=}0.01$, $10{,}000$ iters, hexagonal & 3 \\
Heretic & self-selected & TPE search, 200 trials & minimizes a refusal proxy $+$ KL (self-optimized) & 1 \\
prefill & --- (inference-time) & --- & 3 variants (plain / newline / title); 6 lead-ins, 5 title formats & 3 \\
\bottomrule
\end{tabular}
\end{table}

\clearpage
\section{Glossary}
\label{app:glossary}
\setlength{\tabcolsep}{4pt}\renewcommand{\arraystretch}{0.8}
\begin{longtable}{@{}>{\bfseries\scriptsize}l >{\scriptsize}l >{\raggedright\arraybackslash\scriptsize}p{7.2cm}@{}}
\caption{Glossary of every method and benchmark named in Tables~\ref{tab:evaldef}--\ref{tab:evalatk}.}
\label{tab:glossary}\\
\toprule
Name & Reference & Description \\
\midrule
\endfirsthead
\multicolumn{3}{@{}l@{}}{\emph{Table~\ref{tab:glossary} (continued)}}\\
\toprule
Name & Reference & Description \\
\midrule
\endhead
\bottomrule
\endfoot
\multicolumn{3}{@{}l@{}}{\emph{Abliteration attacks}}\\
ablation & \citet{arditi2024refusal} & Difference-in-means refusal-direction removal \\
ACE & \citet{marshall2024ace} & Affine concept editing (mean + bias term) \\
cones & \citet{wollschlager2025geometry} & Multi-dimensional refusal cones \\
Heretic & \citet{weidmann2025heretic} & Automated, TPE-optimized abliteration tool (5{,}000+ community models) \\
OT & \citet{nanfack2026ot} & Optimal-transport (mean + covariance) ablation \\
RFM & \citet{winninger2026rfm} & AGOP multi-dimensional refusal subspace \\
SOM & \citet{piras2026som} & Multi-directional refusal suppression via self-organizing map \\
\addlinespace
\multicolumn{3}{@{}l@{}}{\emph{Refusal-direction \& steering baselines}}\\
AcT & \citet{rodriguez2025activation} & Activation transport (per-dimension mean/scale steering) \\
CAA & \citet{rimsky2024caa} & Contrastive activation addition (steering) \\
\addlinespace
\multicolumn{3}{@{}l@{}}{\emph{Jailbreak \& fine-tuning attacks}}\\
AutoDAN & \citet{liu2024autodan} & Genetic stealthy jailbreak prompts \\
AutoPrompt & \citet{shin2020autoprompt} & Gradient-based prompt search \\
CodeAttack & \citet{ren2024codeattack} & Code-completion jailbreak \\
GCG & \citet{zou2023advbench} & Greedy coordinate-gradient adversarial suffix \\
HFT & \citet{qi2023finetuning} & Harmful fine-tuning \\
HumanJB & \citet{shen2023dan} & In-the-wild human-written jailbreaks \\
multiling. & \citet{deng2024multilingual} & Multilingual jailbreak \\
prefill & \citet{li2025prefill} & Response-prefill jailbreak attack \\
RepE & \citet{zou2023repe} & Representation-engineering steering \\
SAA & \citet{andriushchenko2024jailbreaking} & Random-search adversarial suffix, with a prefill variant \\
TAP & \citet{mehrotra2024tap} & Tree-of-attacks jailbreak \\
\addlinespace
\multicolumn{3}{@{}l@{}}{\emph{Defense baselines}}\\
AlphaSteer & \citet{sheng2025alphasteer} & Null-space refusal steering \\
AMRA & \citet{truong2026amra} & Refusal aliasing (post-hoc weight edit) \\
ART & \citet{kuo2026art} & Abliteration-resistant tuning (add-on defense) \\
Booster & \citet{huang2025booster} & Harmful-perturbation attenuation (fine-tuning defense) \\
CAST & \citet{lee2024cast} & Conditional activation steering \\
CAT & \citet{xhonneux2024cat} & Continuous adversarial training \\
LAT & \citet{sheshadri2025lat} & Targeted latent adversarial training \\
Lisa & \citet{huang2024lisa} & Lazy safety alignment (fine-tuning defense) \\
R2D2 & \citet{mazeika2024harmbench} & Adversarial training (HarmBench) \\
ReFAT & \citet{yu2025refat} & Refusal-feature adversarial training \\
RepNoise & \citet{rosati2024representation} & Representation noising (fine-tuning defense) \\
RR & \citet{zou2024circuitbreakers} & Representation rerouting (Circuit Breakers) \\
RT-Aug & \citet{qi2024shallow} & Harmful-prefix-augmented refusal training \\
SEAM & \citet{wang2025seam} & Self-destructing model (fine-tuning defense) \\
Surgical & \citet{wang2024surgical} & Single-vector false-refusal mitigation \\
TAR & \citet{tamirisa2025tar} & Tamper-resistance meta-training \\
Vaccine & \citet{huang2024vaccine} & Perturbation-aware alignment (fine-tuning defense) \\
\addlinespace
\multicolumn{3}{@{}l@{}}{\emph{Inference-time defenses}}\\
paraphrase & \citet{jain2023baseline} & Input-paraphrasing defense \\
ppl-filter & \citet{alon2023perplexity} & Perplexity filtering \\
SmoothLLM & \citet{robey2025smoothllm} & Randomized-smoothing defense \\
\addlinespace
\multicolumn{3}{@{}l@{}}{\emph{Benchmarks \& judges}}\\
AdvBench & \citet{zou2023advbench} & Harmful-behaviors benchmark \\
BeaverTails & \citet{ji2023beavertails} & Harmful-QA set + moderation judge \\
C4 & \citet{raffel2020c4} & Web-text corpus (perplexity) \\
CatQA & \citet{bhardwaj2024catqa} & Categorical harmful-QA set \\
GSM8K & \citet{cobbe2021gsm8k} & Grade-school math \\
HarmBench & \citet{mazeika2024harmbench} & Harm benchmark + Llama-2-13B classifier \\
HellaSwag & \citet{zellers2019hellaswag} & Commonsense benchmark \\
Llama-Guard & \citet{inan2023llamaguard} & Safety classifier \\
MMLU & \citet{hendrycks2021mmlu} & Multitask knowledge benchmark \\
MT-Bench & \citet{zheng2023mtbench} & Multi-turn LLM-judge benchmark \\
Open-LLM & \citet{beeching2023openllm} & Open LLM Leaderboard (v1) \\
StrongREJECT & \citet{souly2024strongreject} & Rubric-based jailbreak autograder \\
TruthfulQA & \citet{lin2022truthfulqa} & Truthfulness benchmark \\
XSTest & \citet{rottger2024xstest} & Exaggerated-safety / over-refusal test \\
\end{longtable}

\end{document}